\documentclass[sigconf]{sty/aamas}
\usepackage{balance} 
\usepackage{algorithm}
\usepackage[noEnd=true,indLines=true,commentColor=black]{sty/algpseudocodex}
\usepackage{siunitx}
\usepackage[export]{adjustbox}
\usepackage{cleveref}
\usepackage{sty/mystyle}
\title
[Engineering \lacamstar for MAPF]
{Engineering LaCAM$^\ast$: Towards Real-Time, Large-Scale,\\and Near-Optimal Multi-Agent Pathfinding}

\author{Keisuke Okumura$^{1,2}$}
\affiliation{
  \institution{
    $^1$University of Cambridge
  }
  \city{Cambridge}
  \country{United Kingdom}
}
\affiliation{
  \institution{$^2$National Institute of Advanced Industrial Science and Technology (AIST)}
  \city{Tokyo}
  \country{Japan}
}
\email{ko393@cl.cam.ac.uk}

\makeatletter
\setcopyright{none}
\renewcommand\footnotetextcopyrightpermission[1]{}
\makeatother

\begin{abstract}
  This paper addresses the challenges of real-time, large-scale, and near-optimal multi-agent pathfinding (MAPF) through enhancements to the recently proposed \lacamstar algorithm.
  \lacamstar is a scalable search-based algorithm that guarantees the eventual finding of optimal solutions for cumulative transition costs.
  While it has demonstrated remarkable planning success rates, surpassing various state-of-the-art MAPF methods, its initial solution quality is far from optimal, and its convergence speed to the optimum is slow.
  To overcome these limitations, this paper introduces several improvement techniques, partly drawing inspiration from other MAPF methods.
  We provide empirical evidence that the fusion of these techniques significantly improves the solution quality of \lacamstar, thus further pushing the boundaries of MAPF algorithms.
\end{abstract}


\begin{document}
\pagestyle{fancy}
\fancyhf{}
\maketitle
\fancyfoot[C]{\thepage~of~\pageref{TotPages}}
\section{Introduction}
The \emph{multi-agent pathfinding (MAPF)} problem~\cite{stern2019def} seeks to find a collection of collision-free paths for multiple agents on graphs, with appealing applications such as warehouse automation~\cite{wurman2008coordinating} and railway scheduling~\cite{li2021scalable}.
The core challenge for MAPF algorithms is to derive plausible solutions that minimize redundant agent motions, even with hundreds of agents or more, within a realistic computational timeframe (i.e., in real-time; e.g., \SI{10}{\second}).
We address this ultimate objective by enhancing the recently introduced \emph{\lacamstar (lazy constraint addition search for MAPF)} algorithm~\cite{okumura2023lacam,okumura2023lacam2}.

\lacamstar is a search-based algorithm akin to the \astar search~\cite{hart1968formal}.
It is an anytime algorithm that, after the initial solution discovery, gradually improves the solution quality and eventually converges to optimal ones, provided that the solution cost takes the form of cumulative transition costs.
While \lacamstar has showcased remarkable scalability for the number of agents, outperforming other MAPF methods, the initial solution quality is somewhat compromised~\cite{shen2023tracking}.
Furthermore, the solution refinement of vanilla \lacamstar is notably slow and impractical~\cite{okumura2023lacam2};
achieving near-optimal solutions for large-scale MAPF in real-time remains extremely challenging.

{
  \setlength{\tabcolsep}{1pt}
  \newcommand{\figcol}[3]{
    \begin{minipage}{0.057\linewidth}
      \centering
      \begin{tikzpicture}
      \node[anchor=south east]() at (0, 2.95) {\tiny $#2$};
      \node[anchor=south east]() at (0, 2.78) {\tiny $#3$};
      \node[anchor=south east]() at (0, 2.15) {\includegraphics[width=0.57\linewidth,right]{fig/raw/maps/#1.pdf}};
      \node[anchor=south east]() at (0, 0) {\includegraphics[width=1.0\linewidth,height=2.0\linewidth]{fig/raw/top_#1.pdf}};
      \node[anchor=south west, rotate=90, text=black]() at (-0.05, 0.15) {\scriptsize \textcolor{black}{\mapname{#1}}};
      \end{tikzpicture}
    \end{minipage}
  }
  \newcommand{\ytitle}{
  \begin{minipage}{0.01\linewidth}
  \rotatebox[origin=c]{90}{\footnotesize $\leftarrow$~~sum-of-loss / LB\hspace{1.2cm}}
  \end{minipage}
  }
  \begin{figure*}[t!]
    \centering
    \begin{tabular}{crrrrrrrrrrrrrrrrrrr}
      \ytitle&
      \figcol{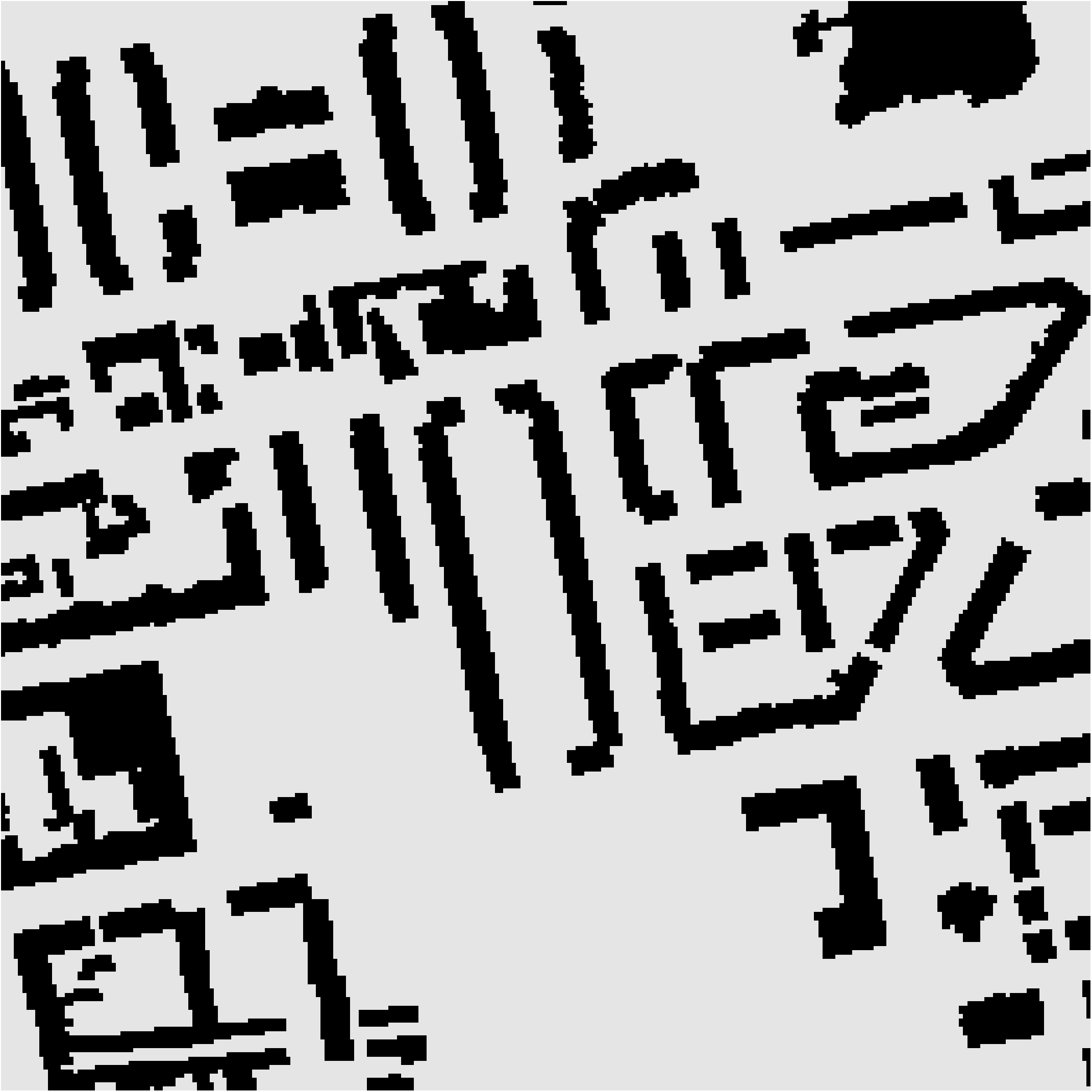}{|A|=1,000}{|V|=47,540}&
      \figcol{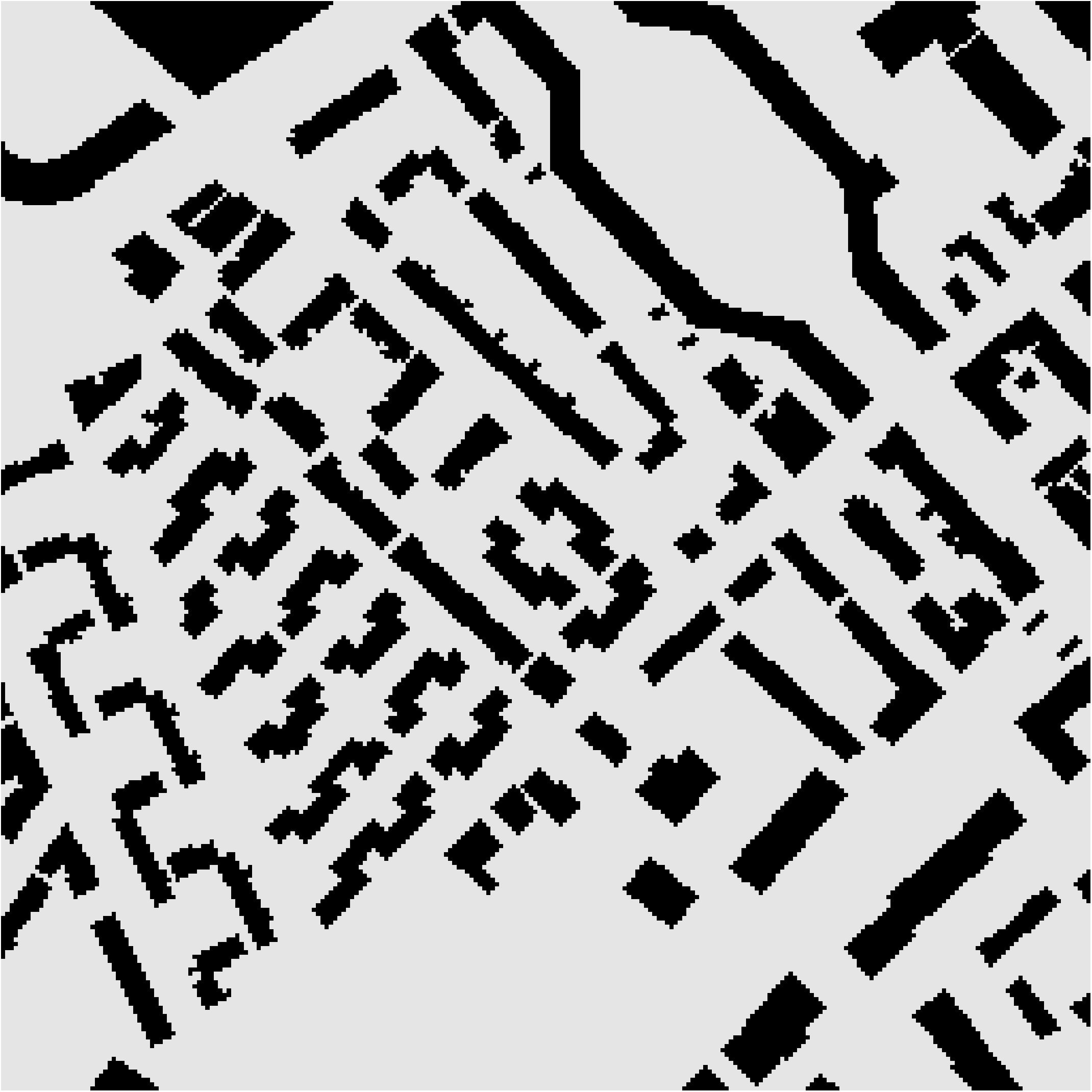}{1,000}{47,768}&
      \figcol{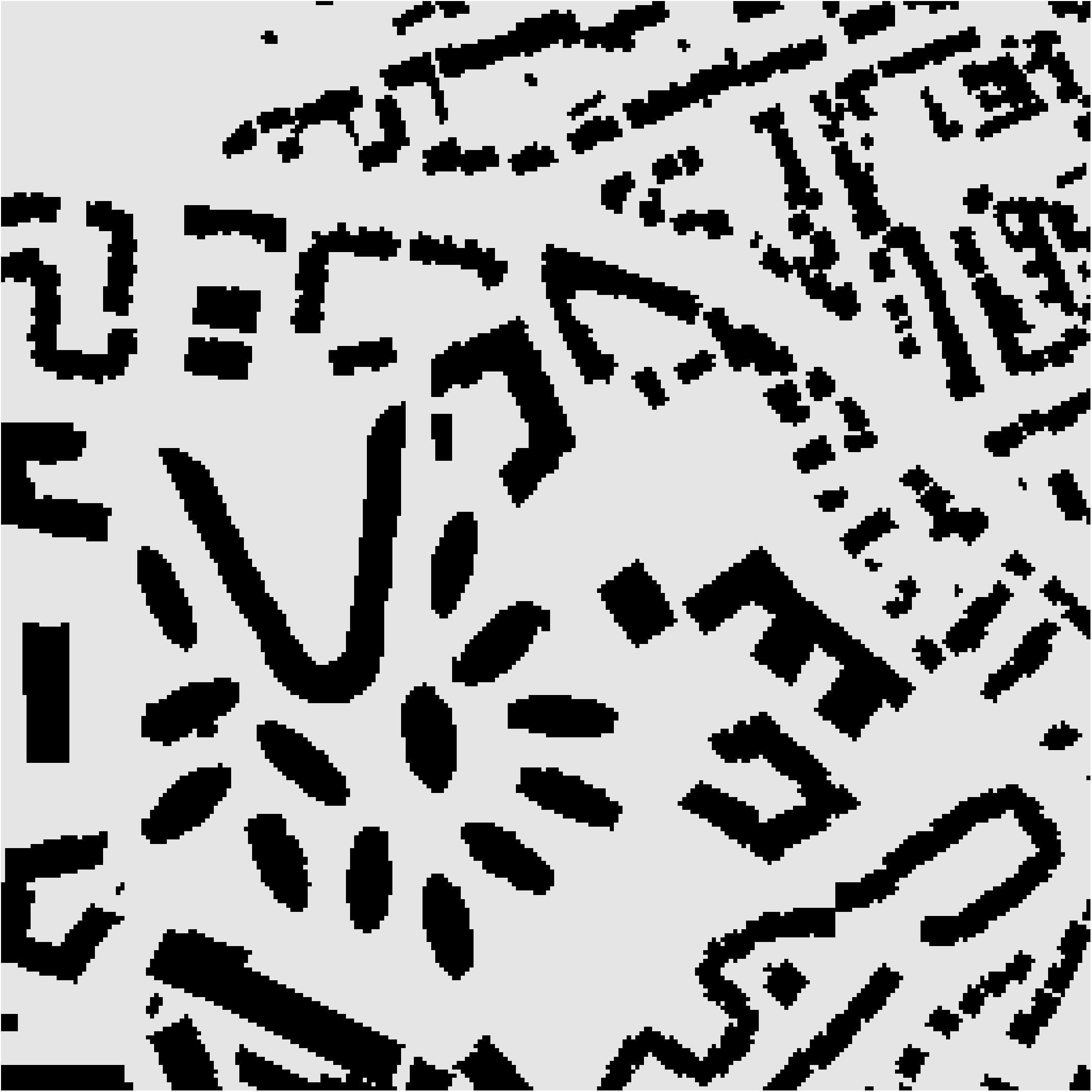}{1,000}{47,240}&
      \figcol{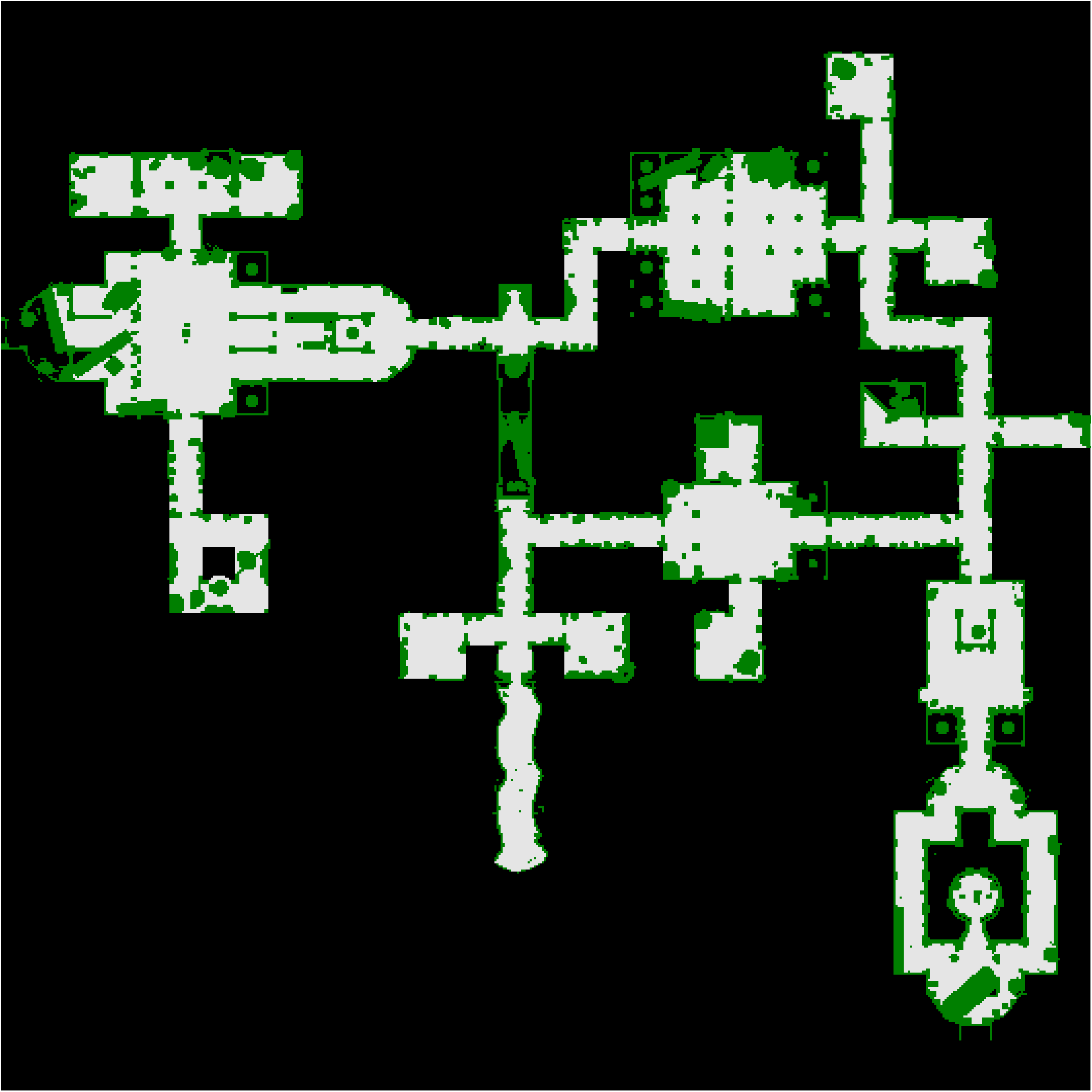}{1,000}{43,151}&
      \figcol{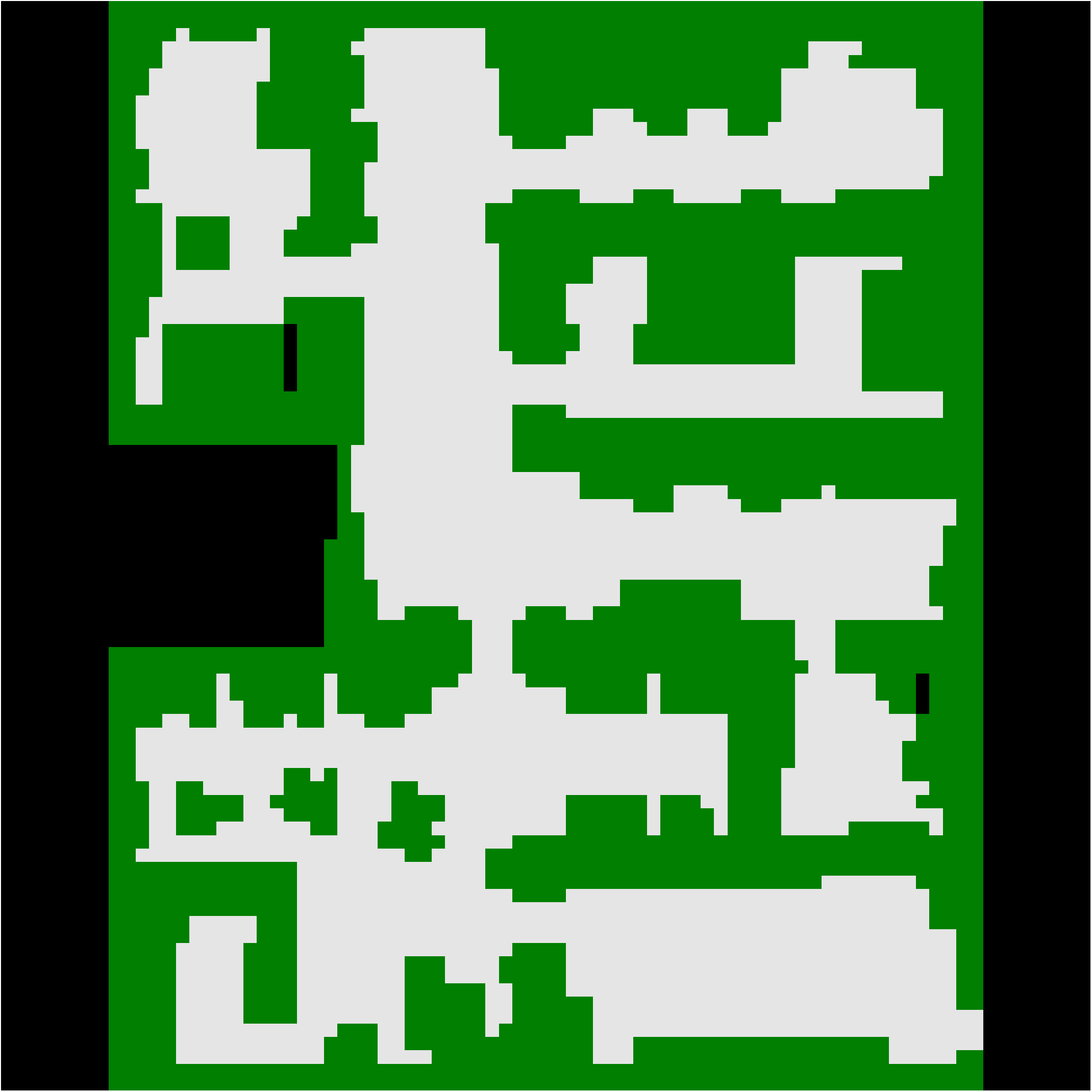}{1,000}{2,445}&
      \figcol{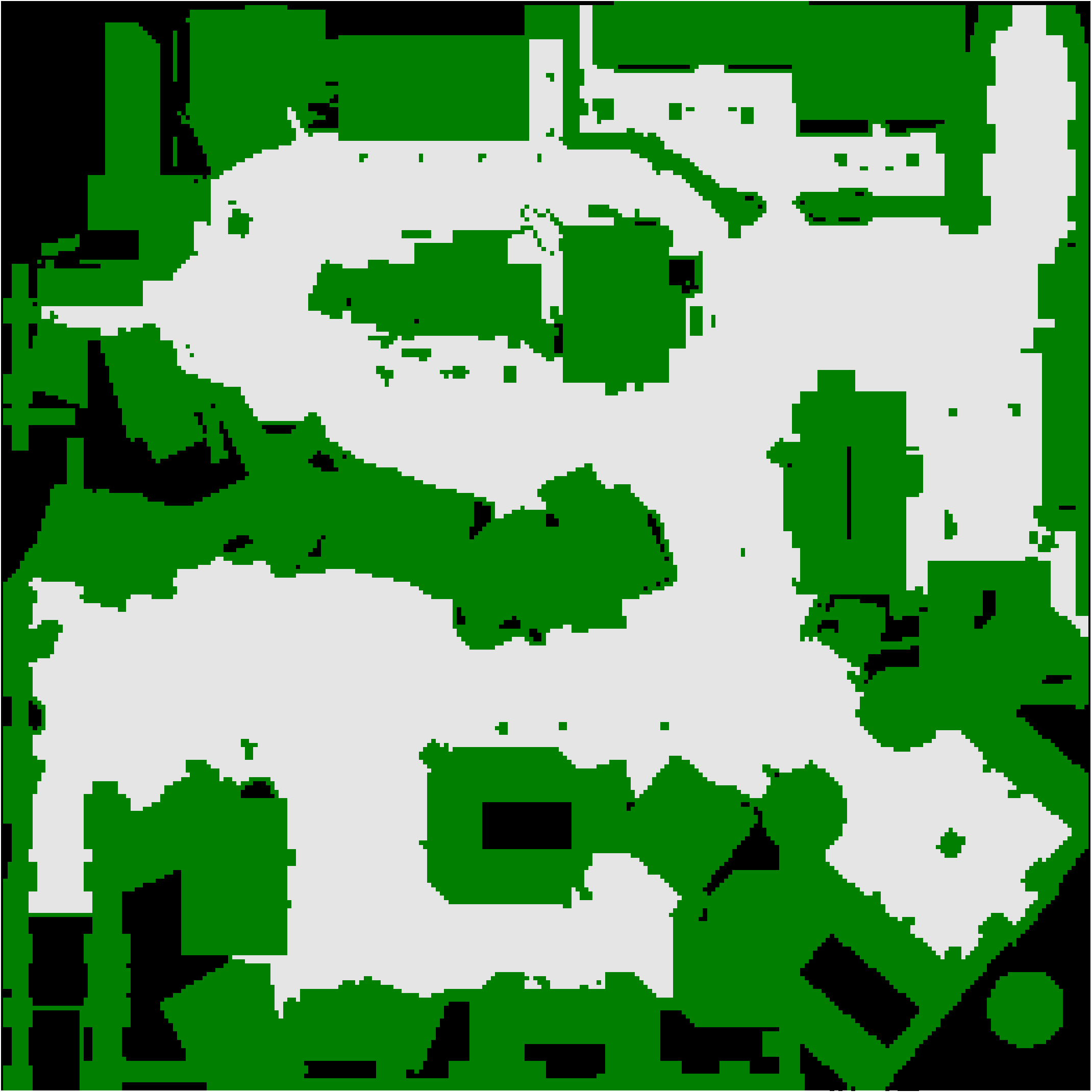}{1,000}{28,178}&
      \figcol{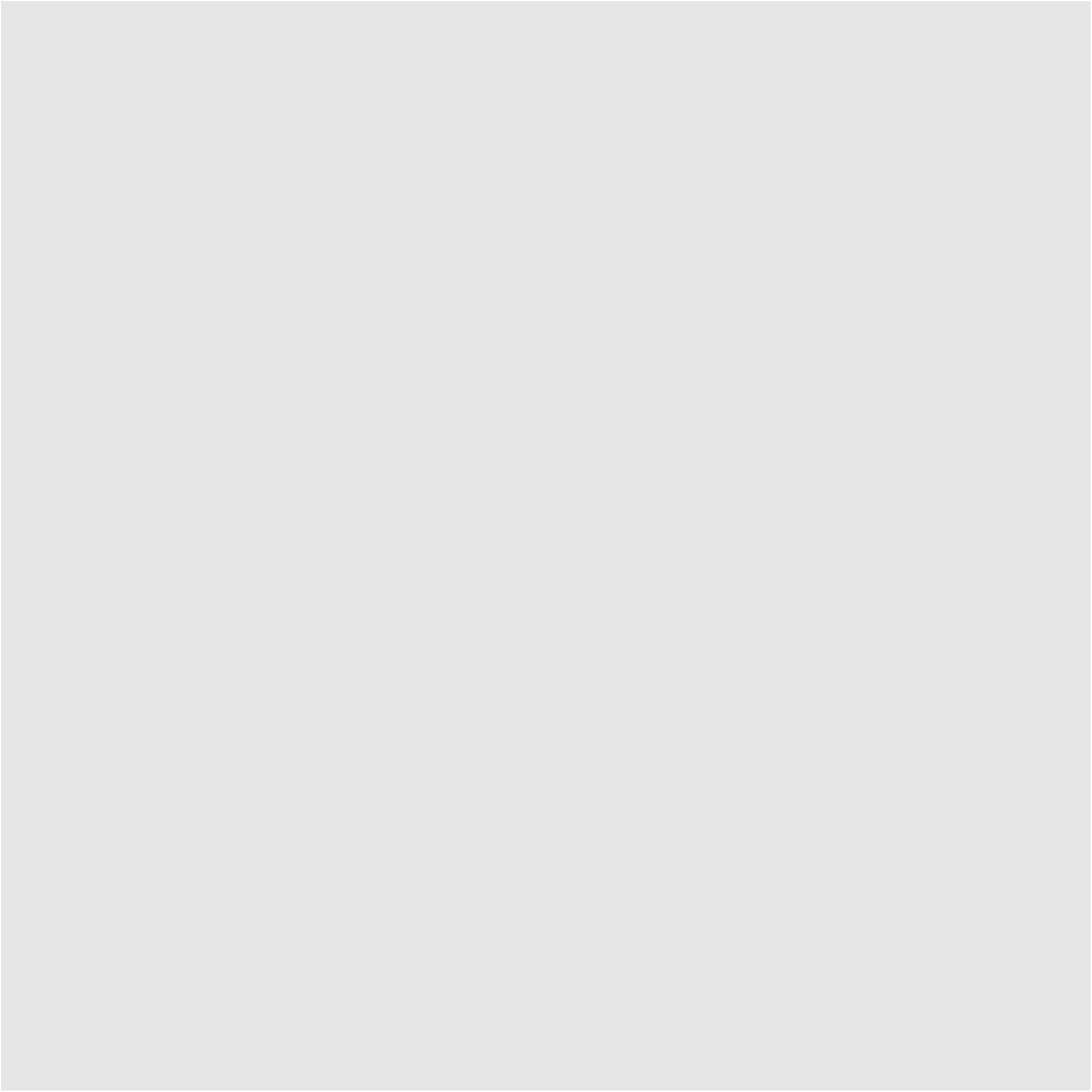}{128}{256}&
      \figcol{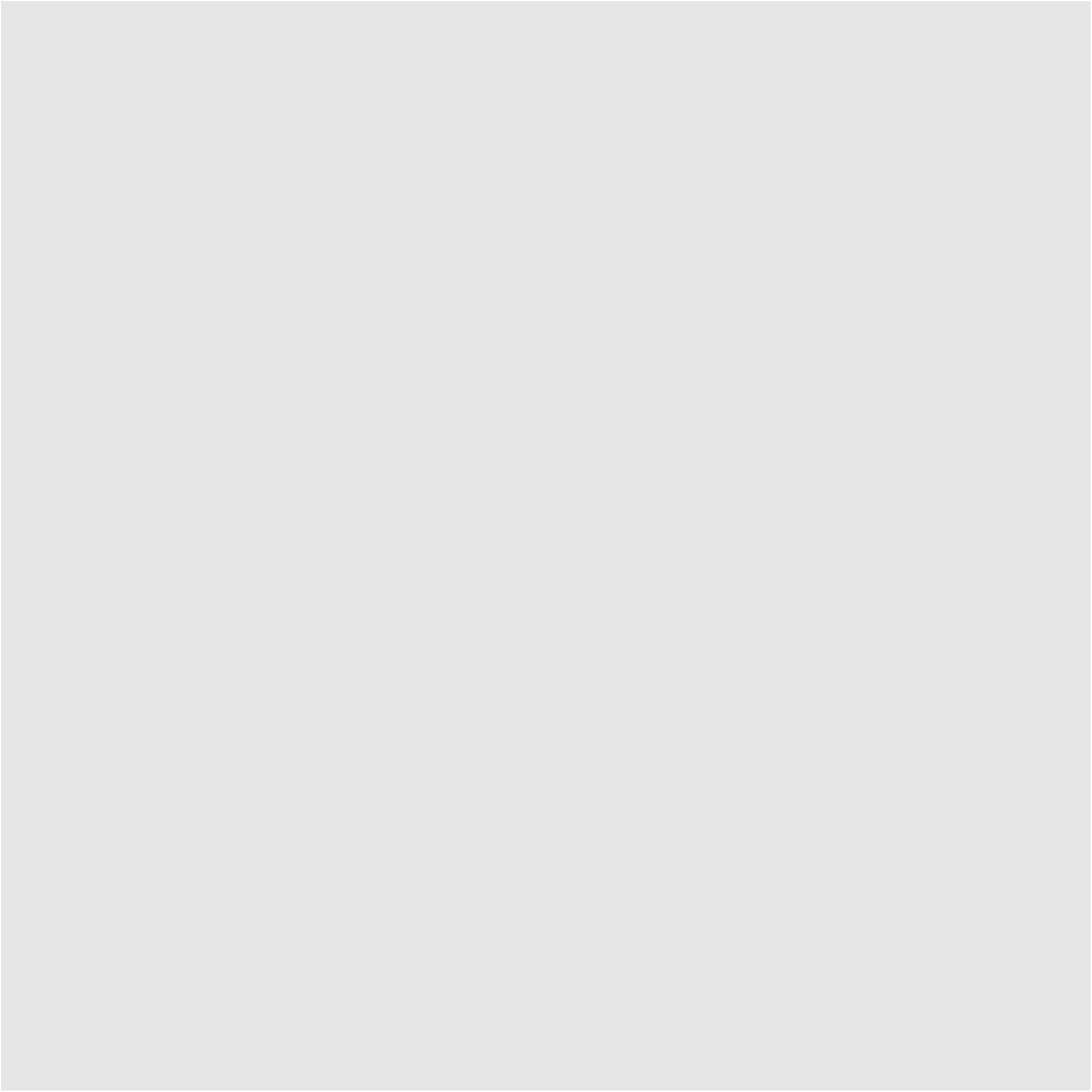}{512}{1,024}&
      \figcol{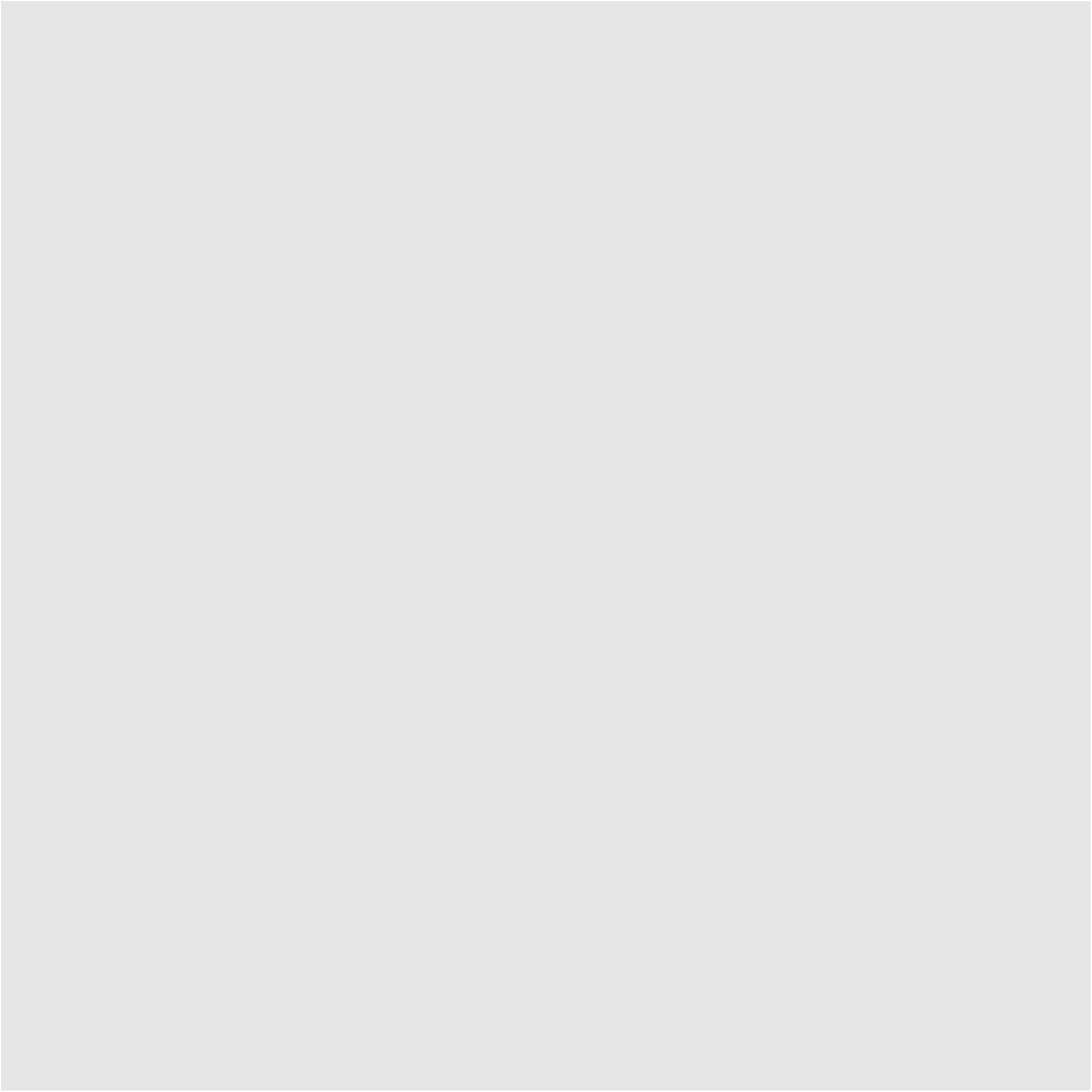}{1000}{2,304}&
      \figcol{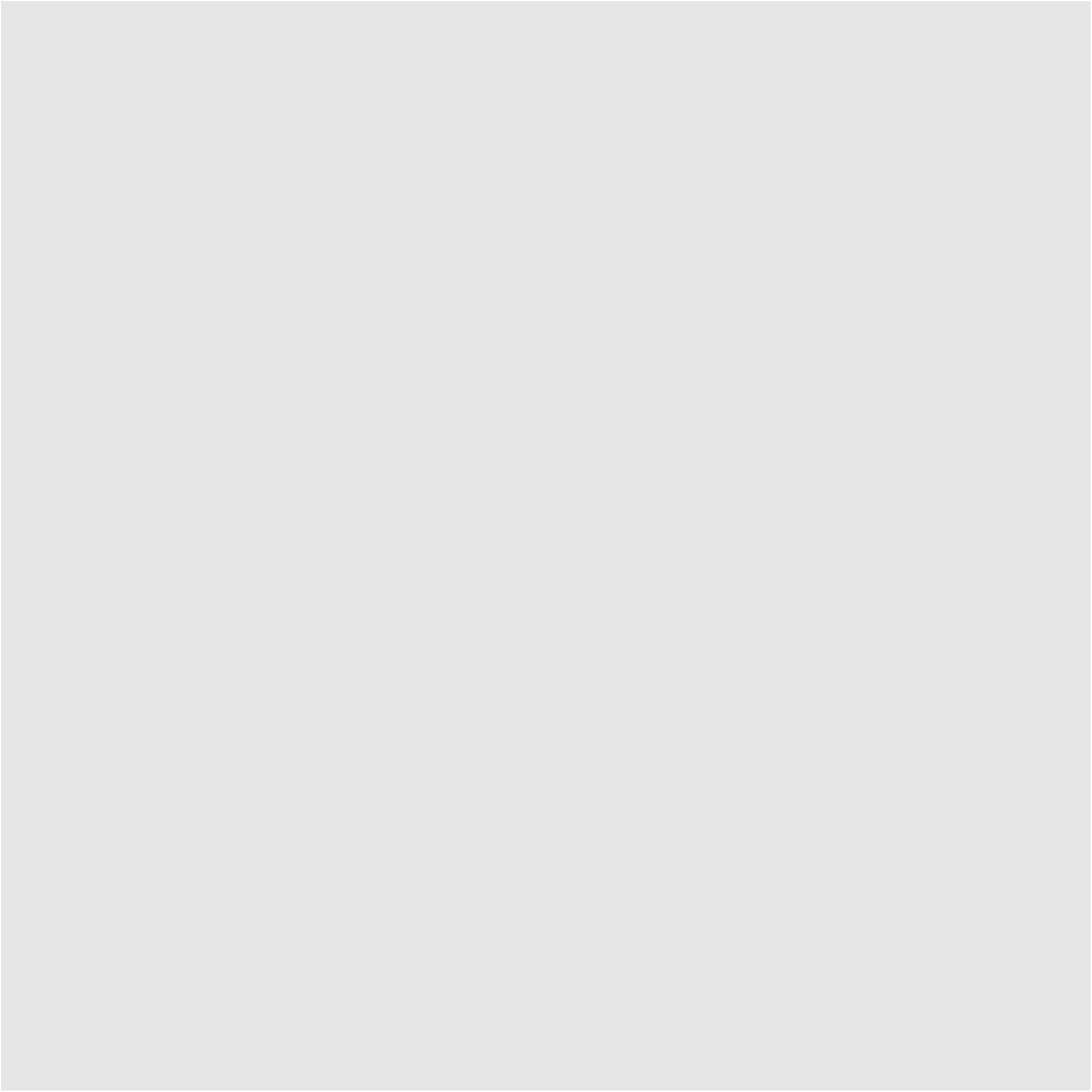}{32}{64}&
      \figcol{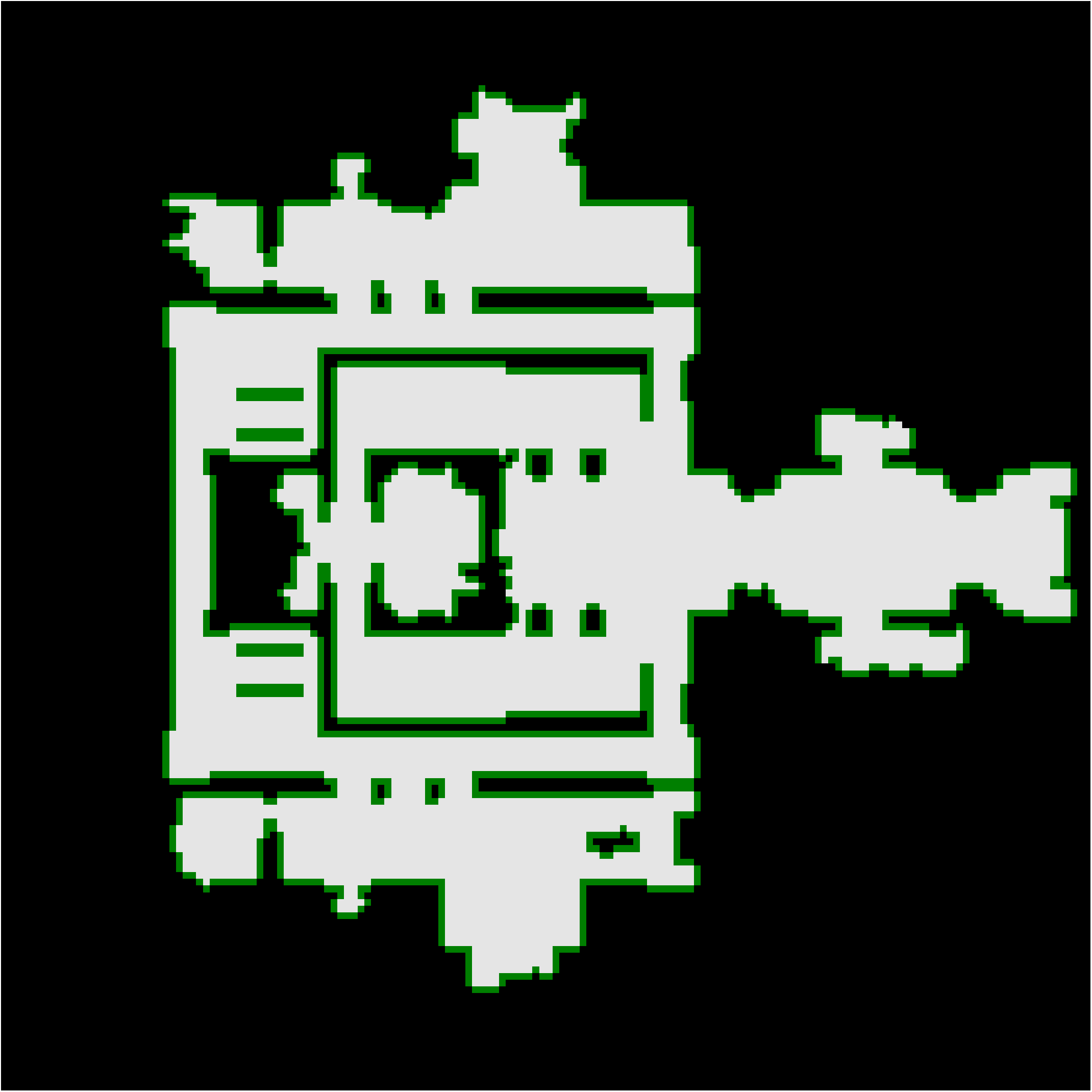}{1,000}{7,461}&
      \figcol{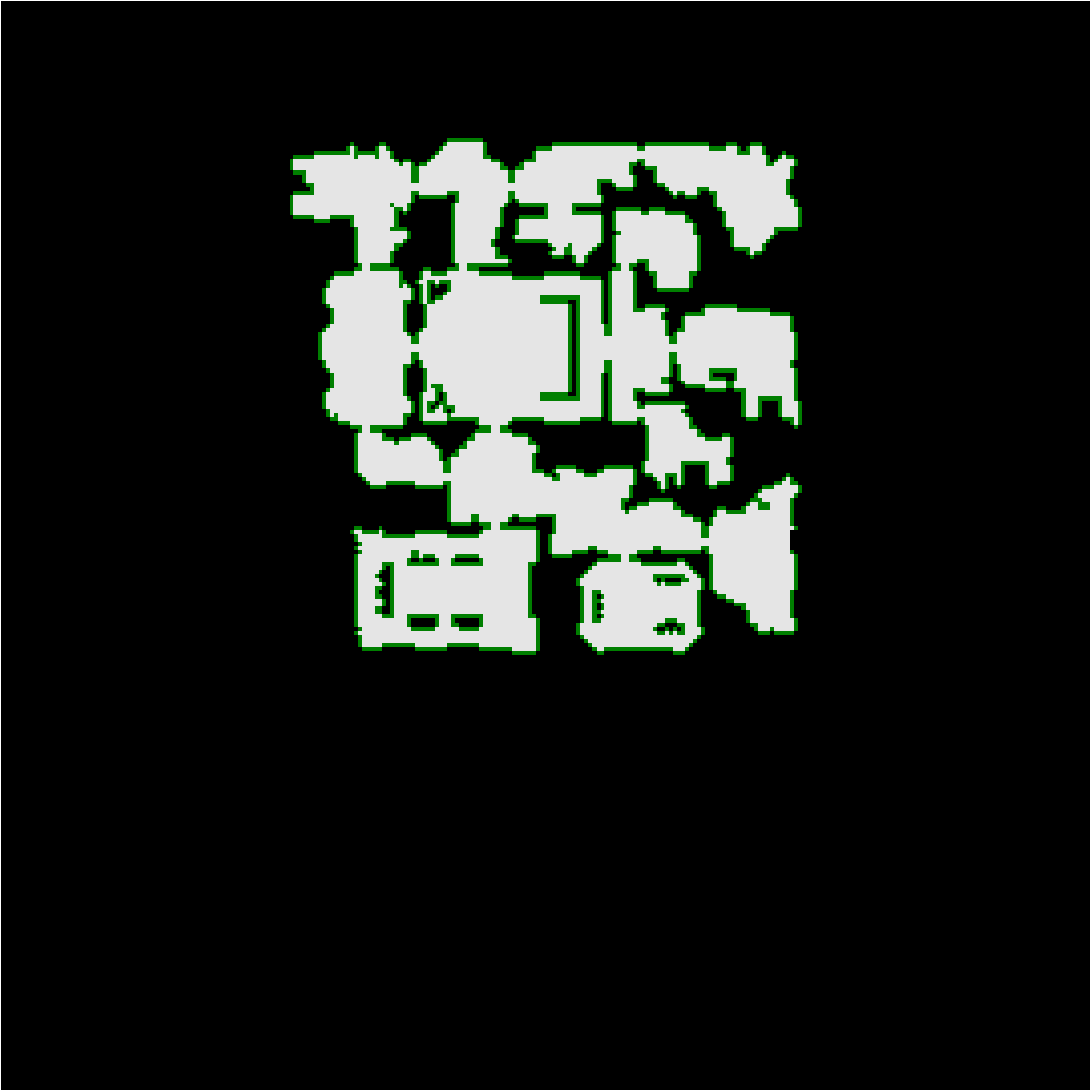}{1,000}{8,959}&
      \figcol{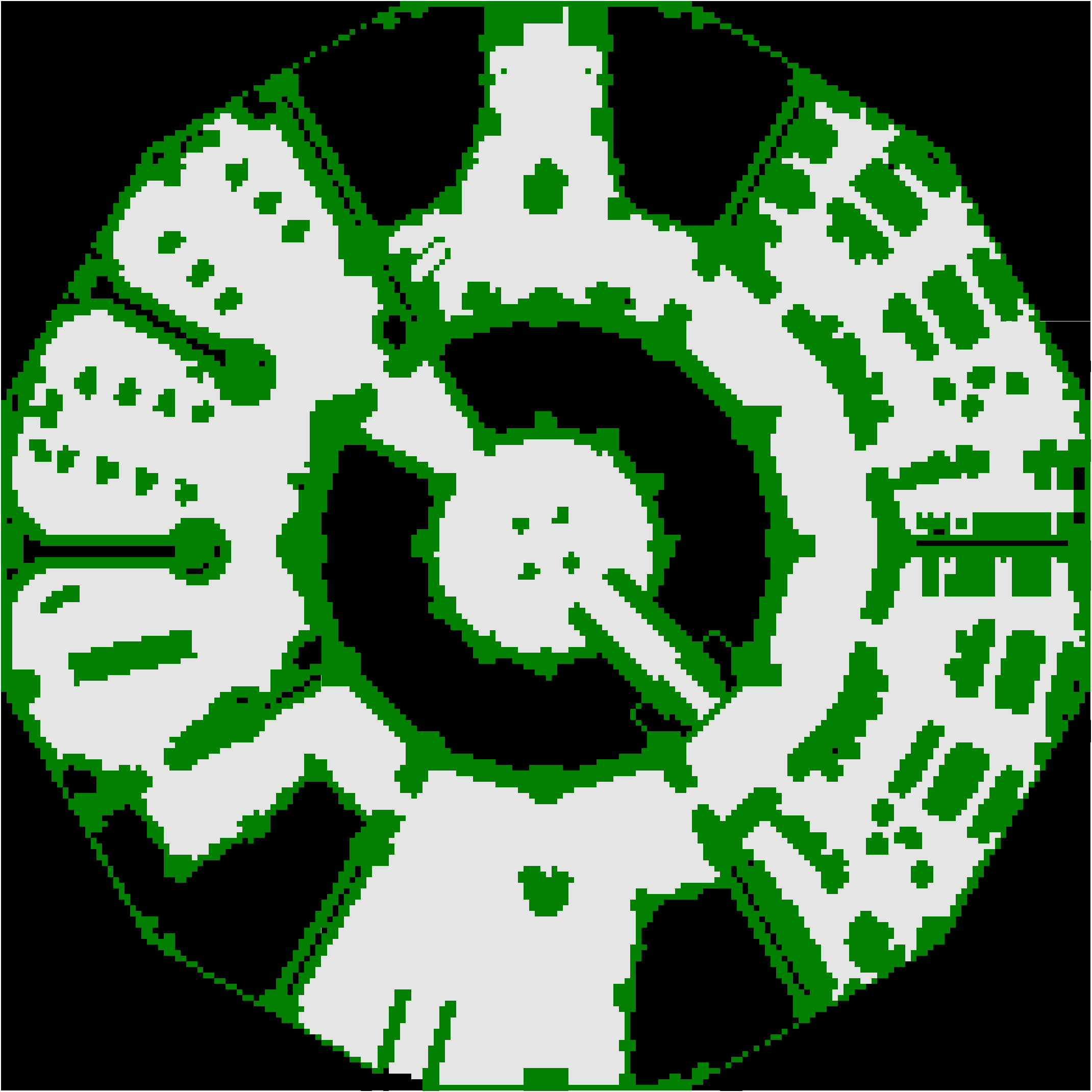}{1,000}{14,784}&
      \figcol{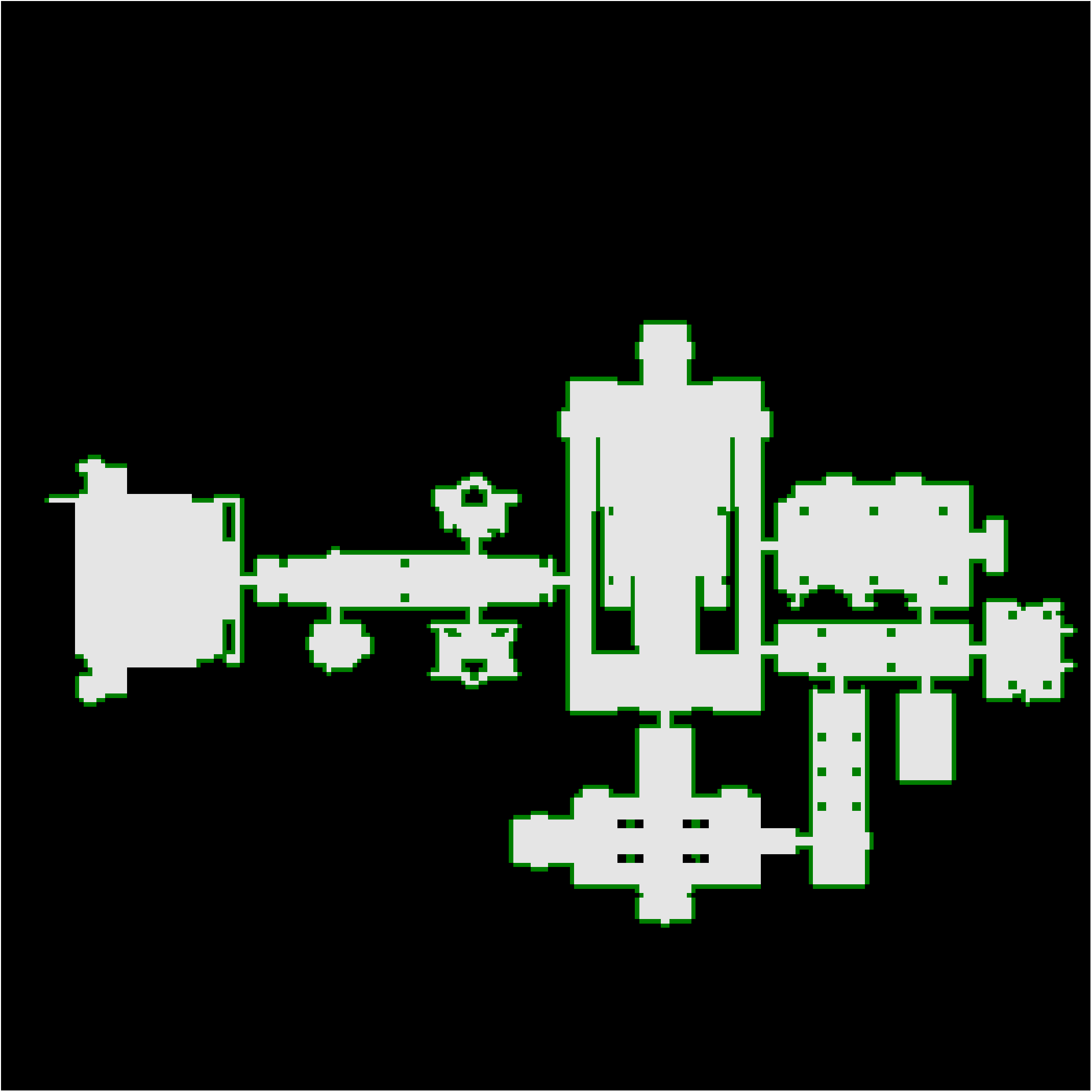}{1,000}{10,021}&
      \figcol{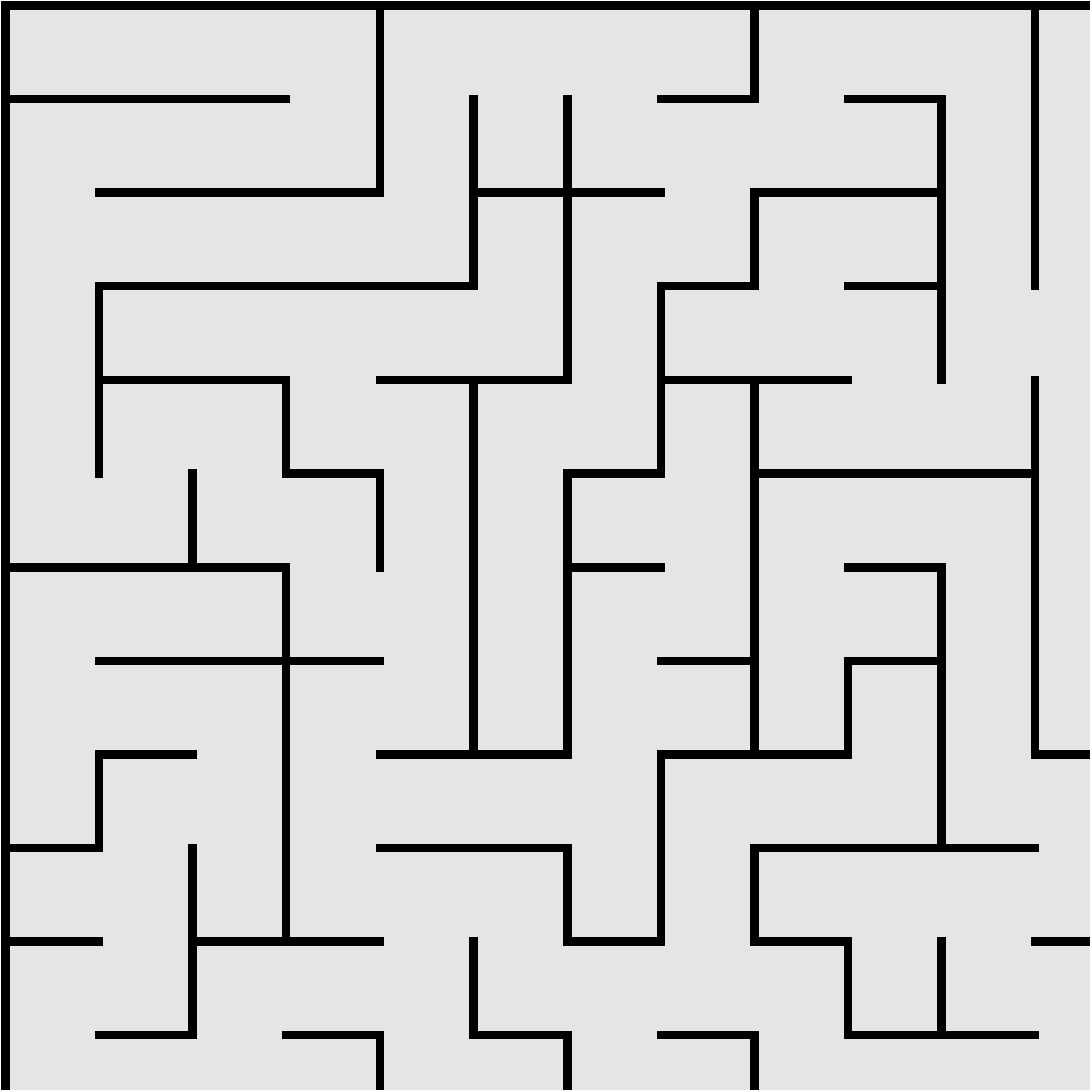}{1,000}{14,818}&
      \figcol{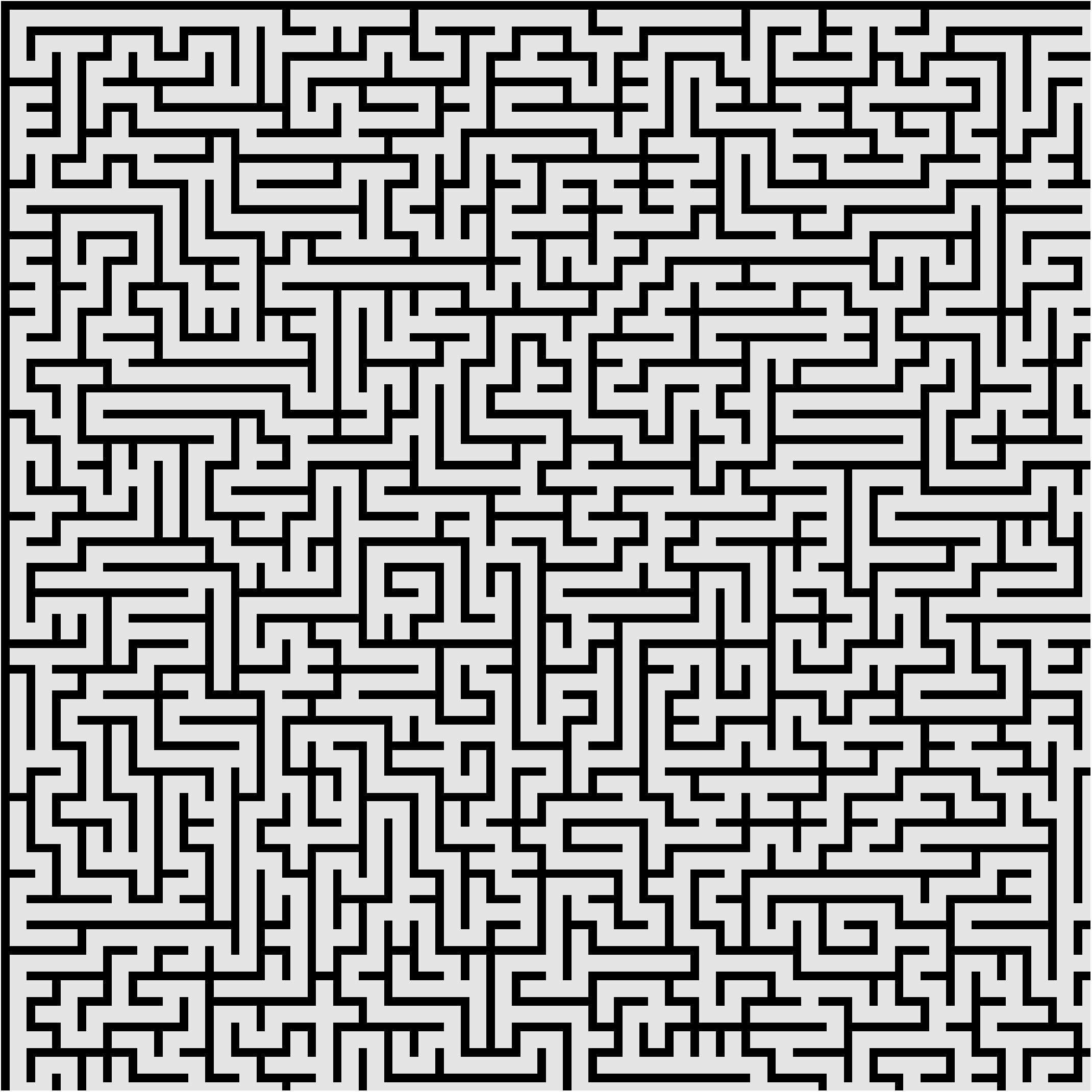}{1,000}{10,858}&
      \\
      \ytitle&
      \figcol{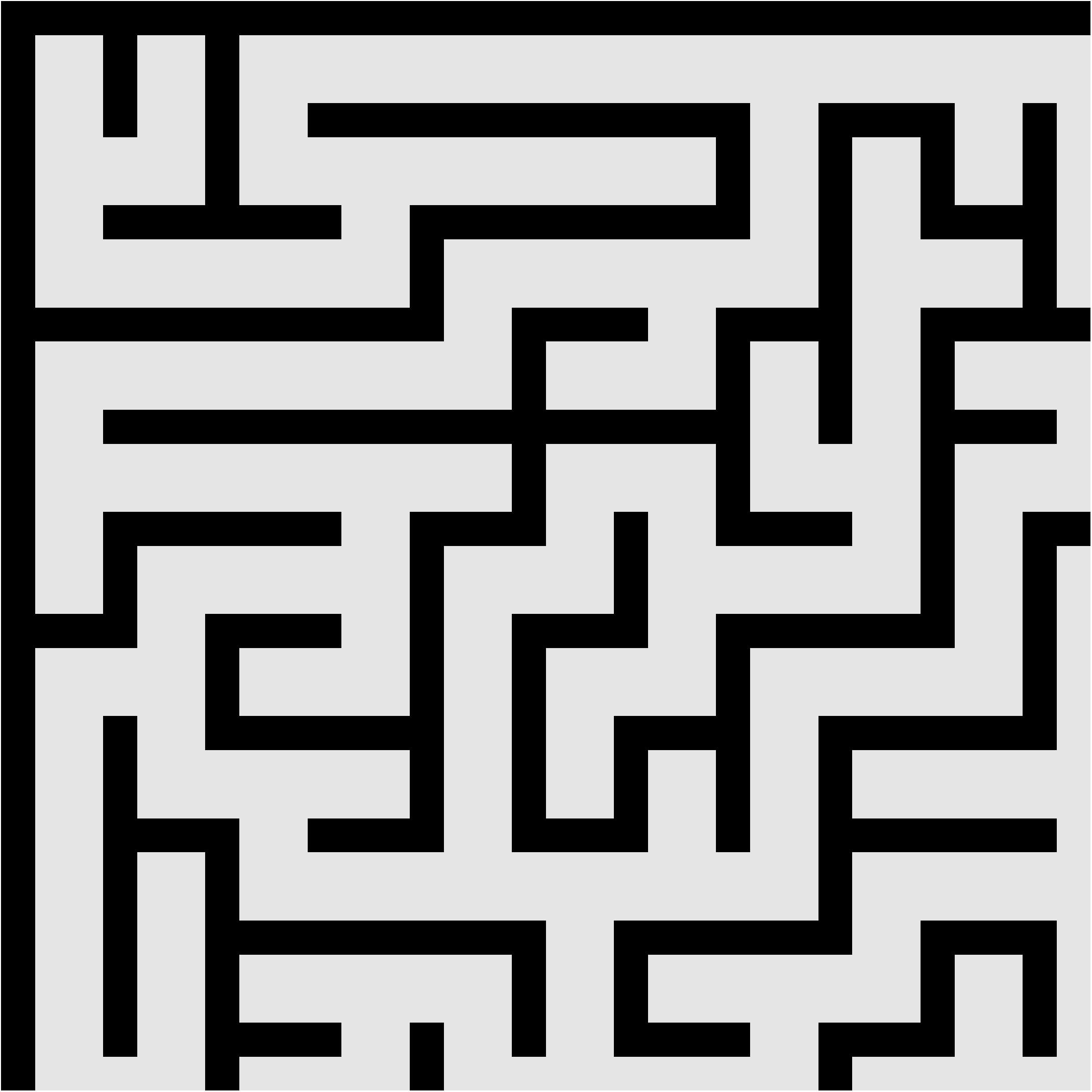}{|A|=333}{|V|=666}&
      \figcol{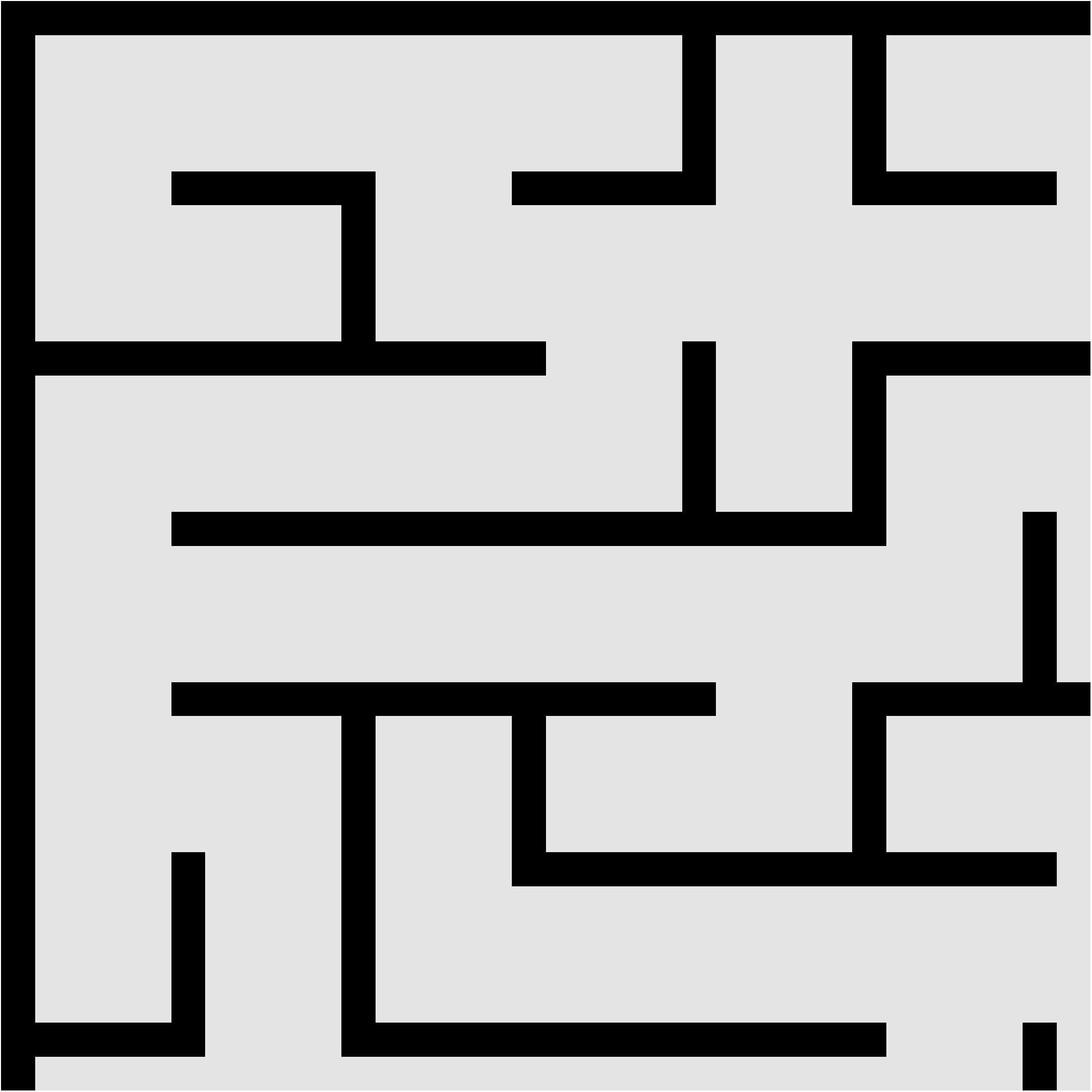}{395}{790}&
      \figcol{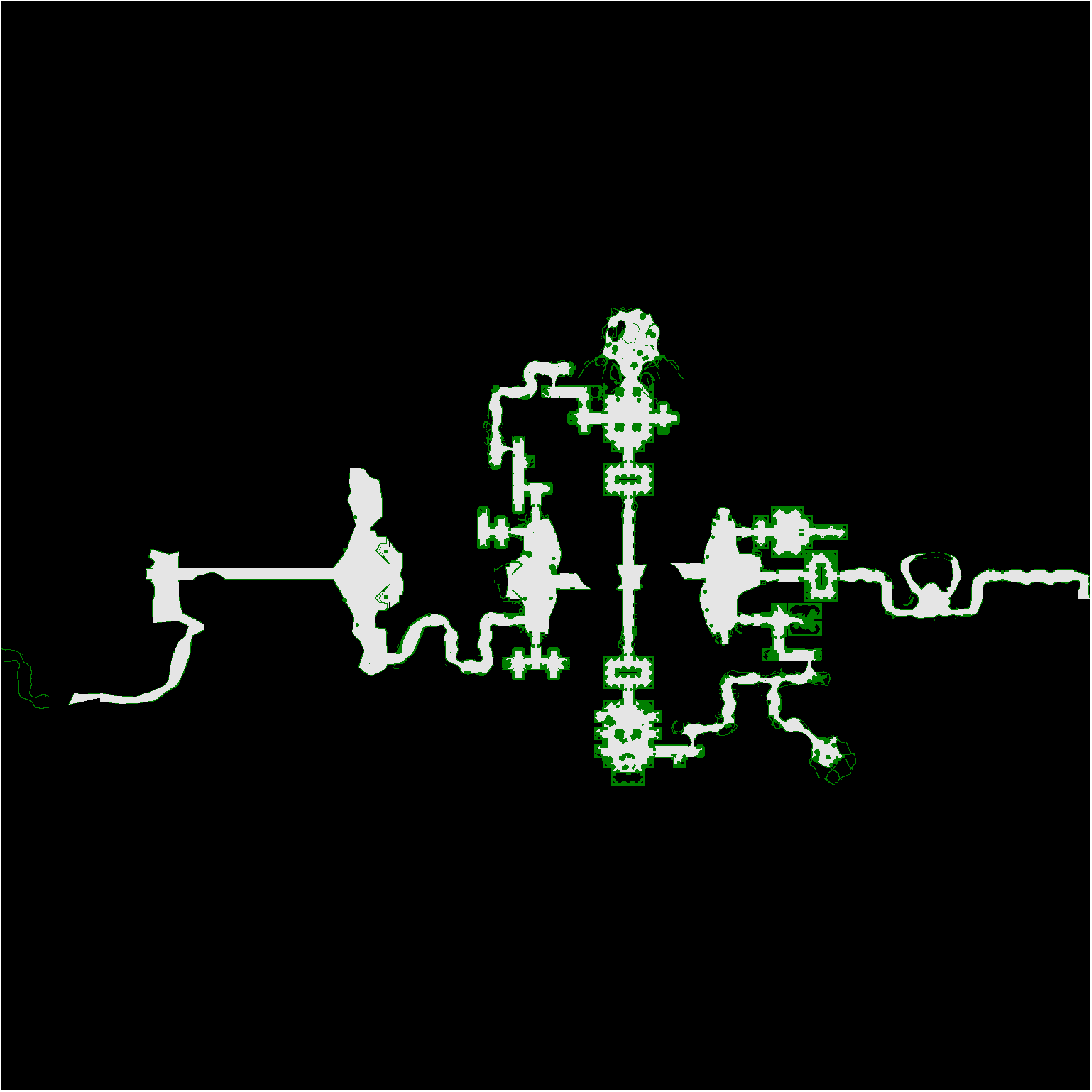}{1,000}{96,603}&
      \figcol{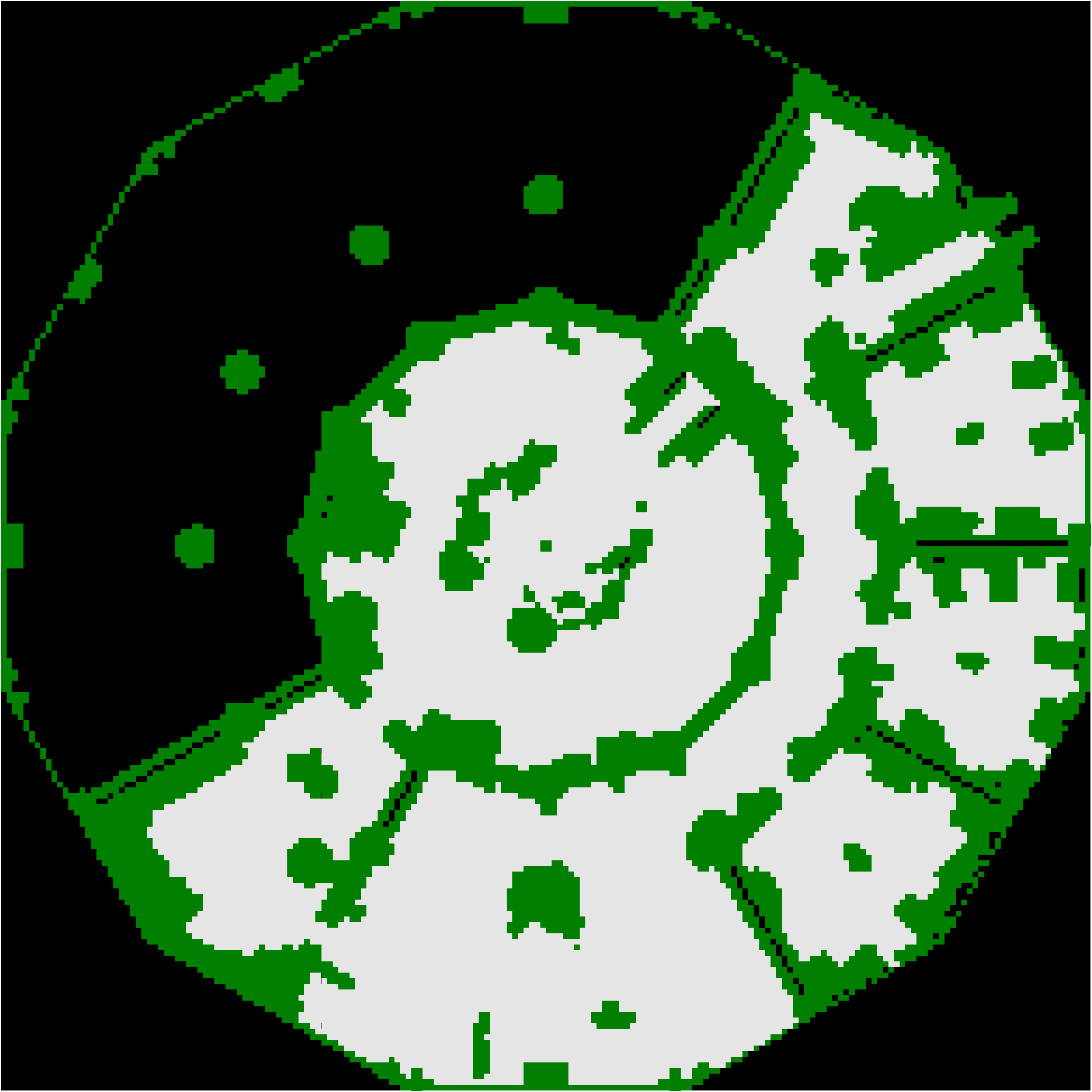}{1,000}{13,214}&
      \figcol{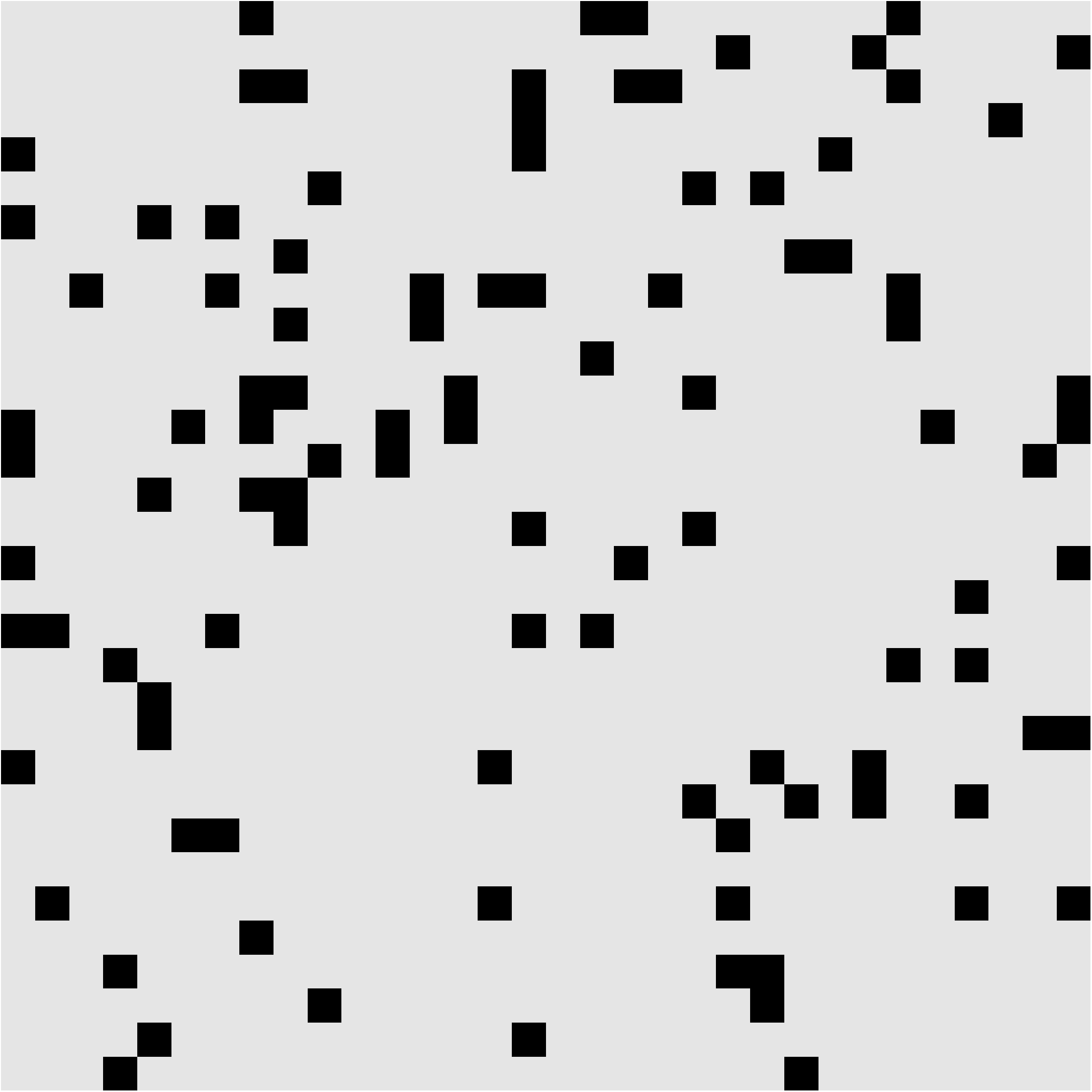}{461}{922}&
      \figcol{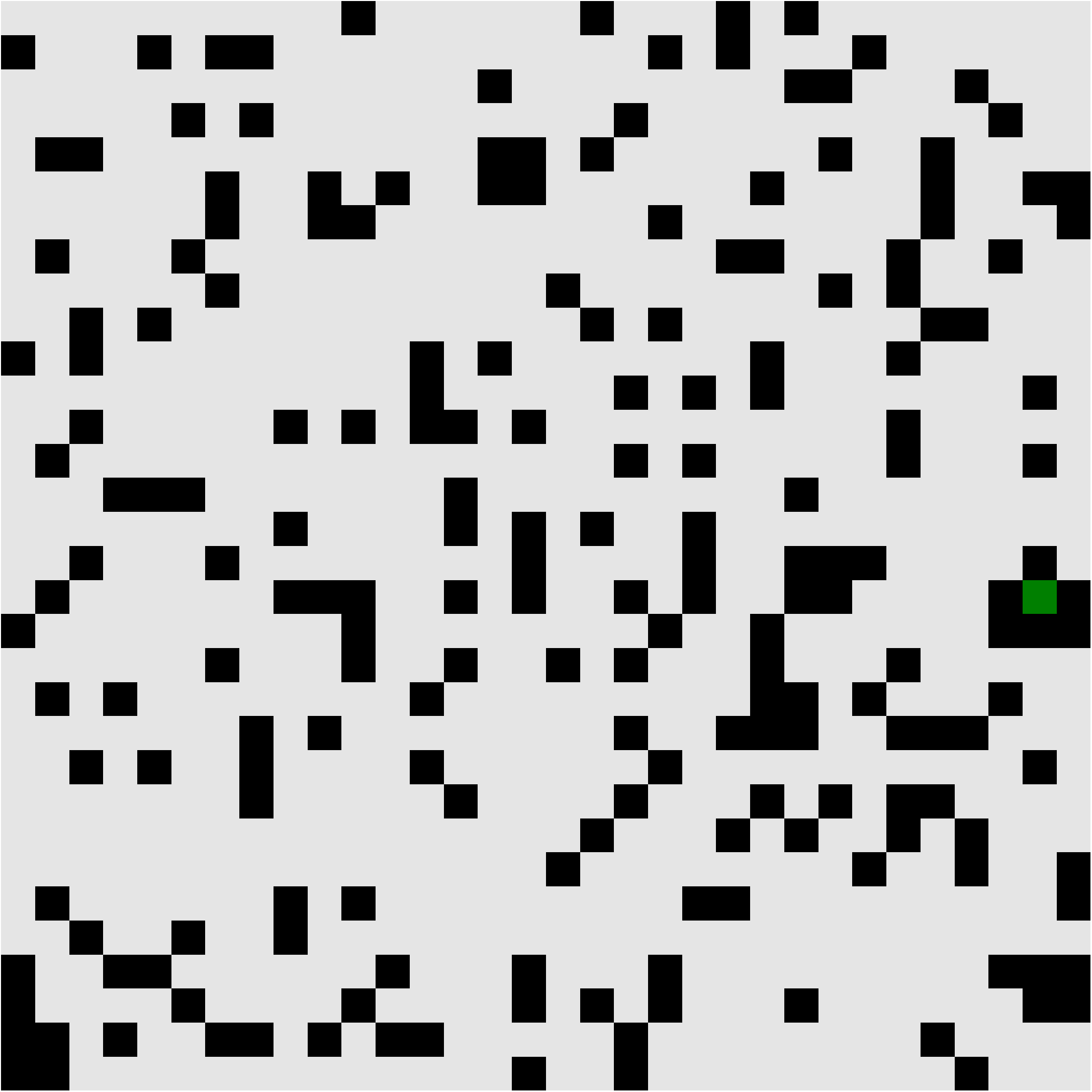}{409}{819}&
      \figcol{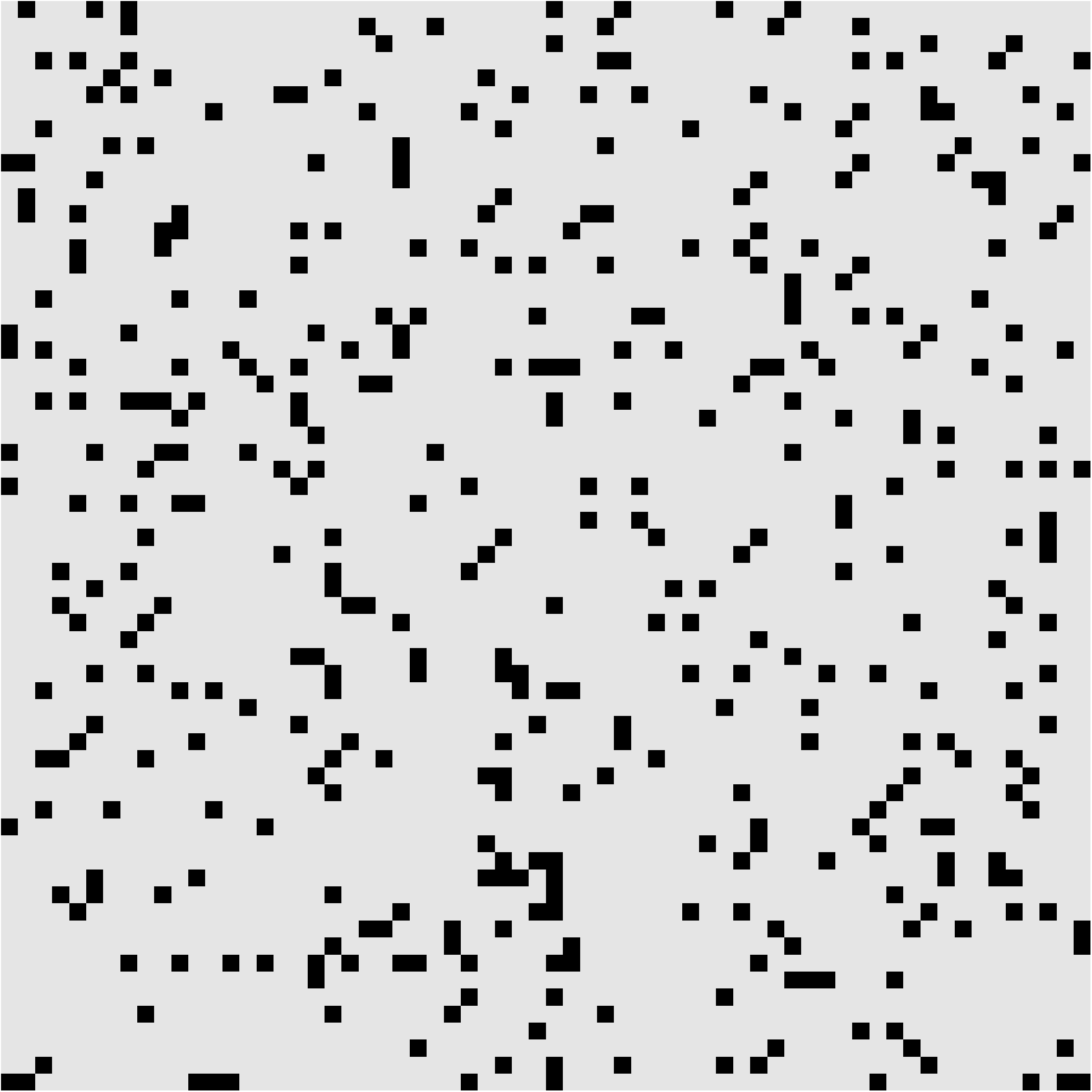}{1,000}{3,687}&
      \figcol{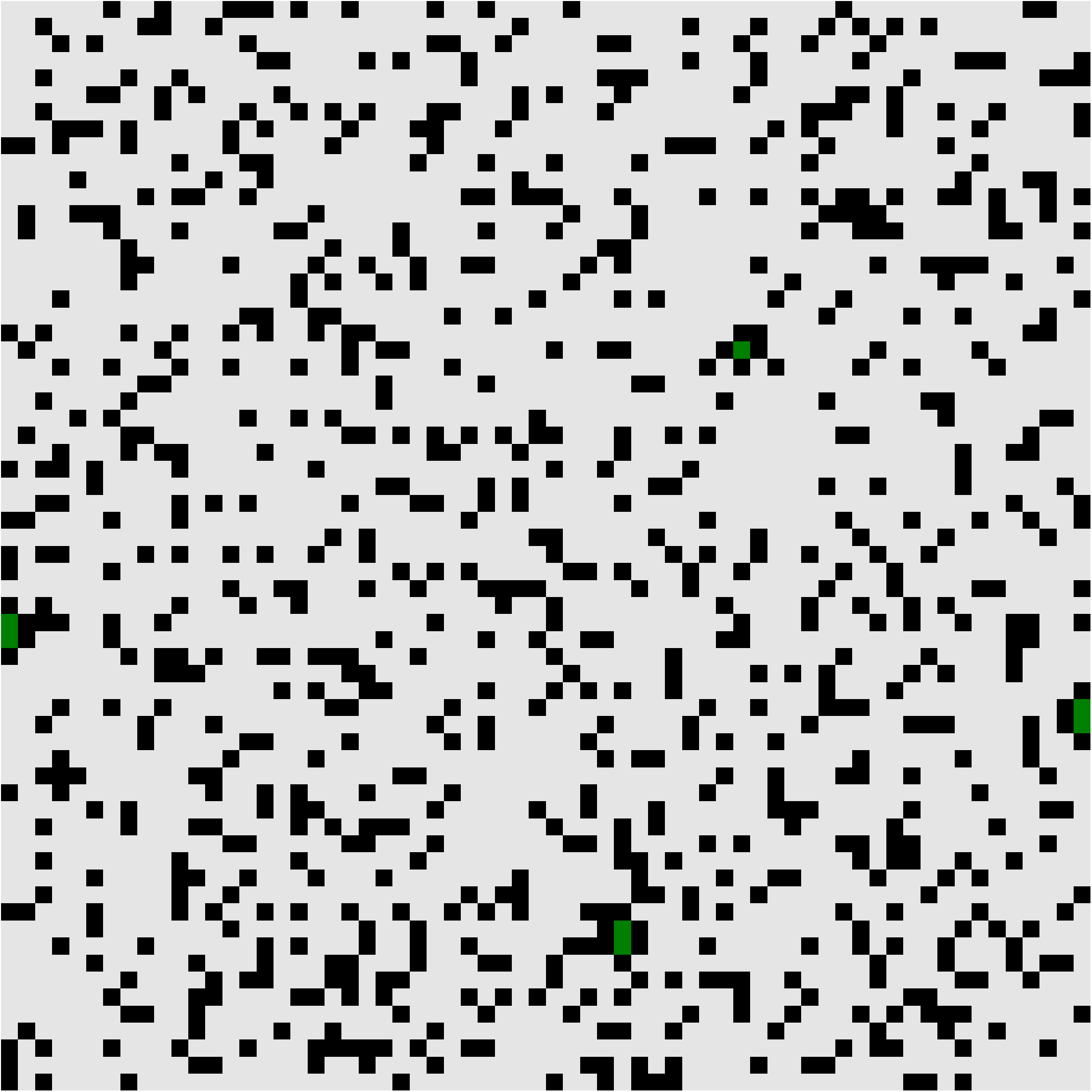}{1,000}{3,270}&
      \figcol{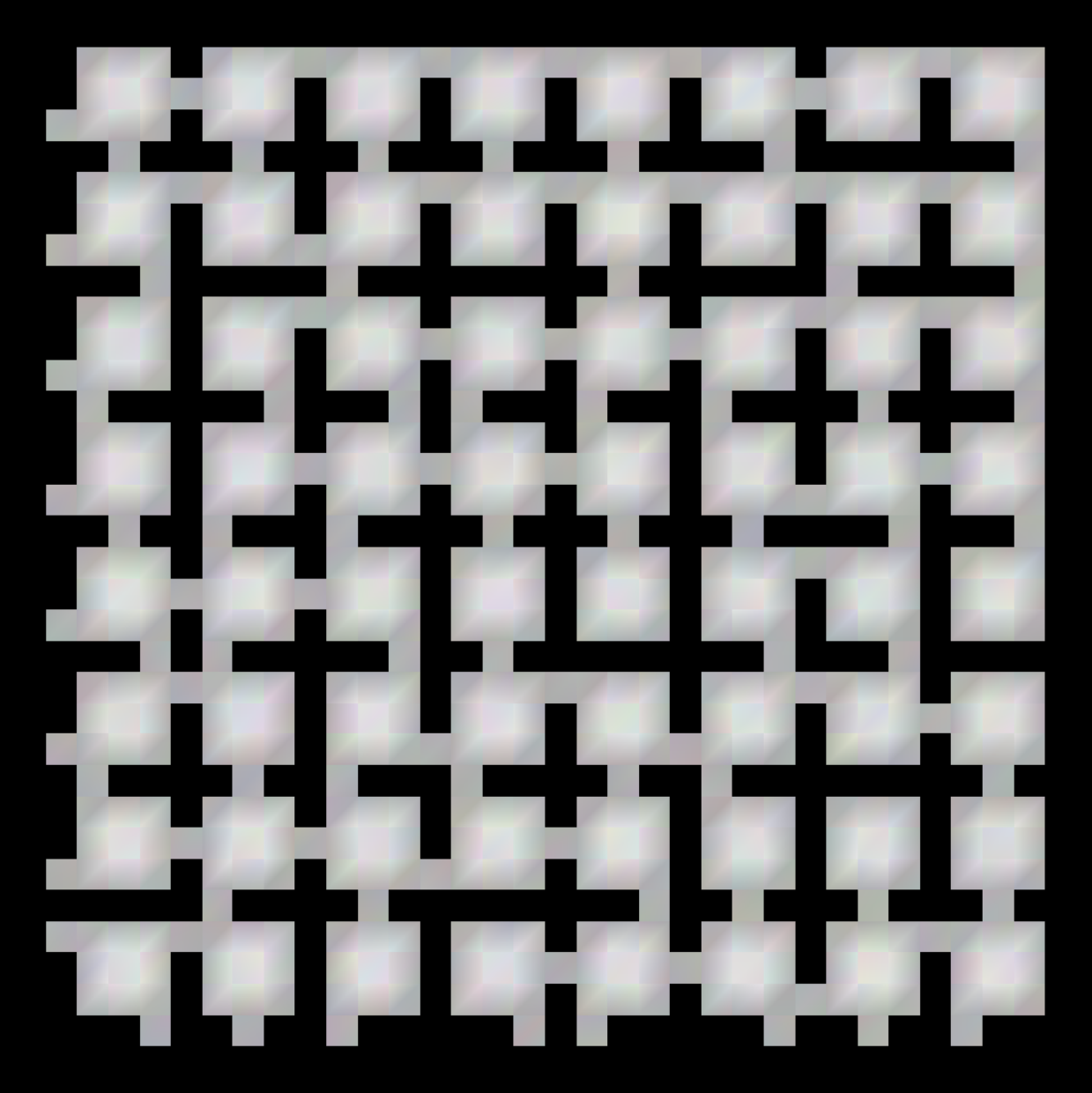}{341}{682}&
      \figcol{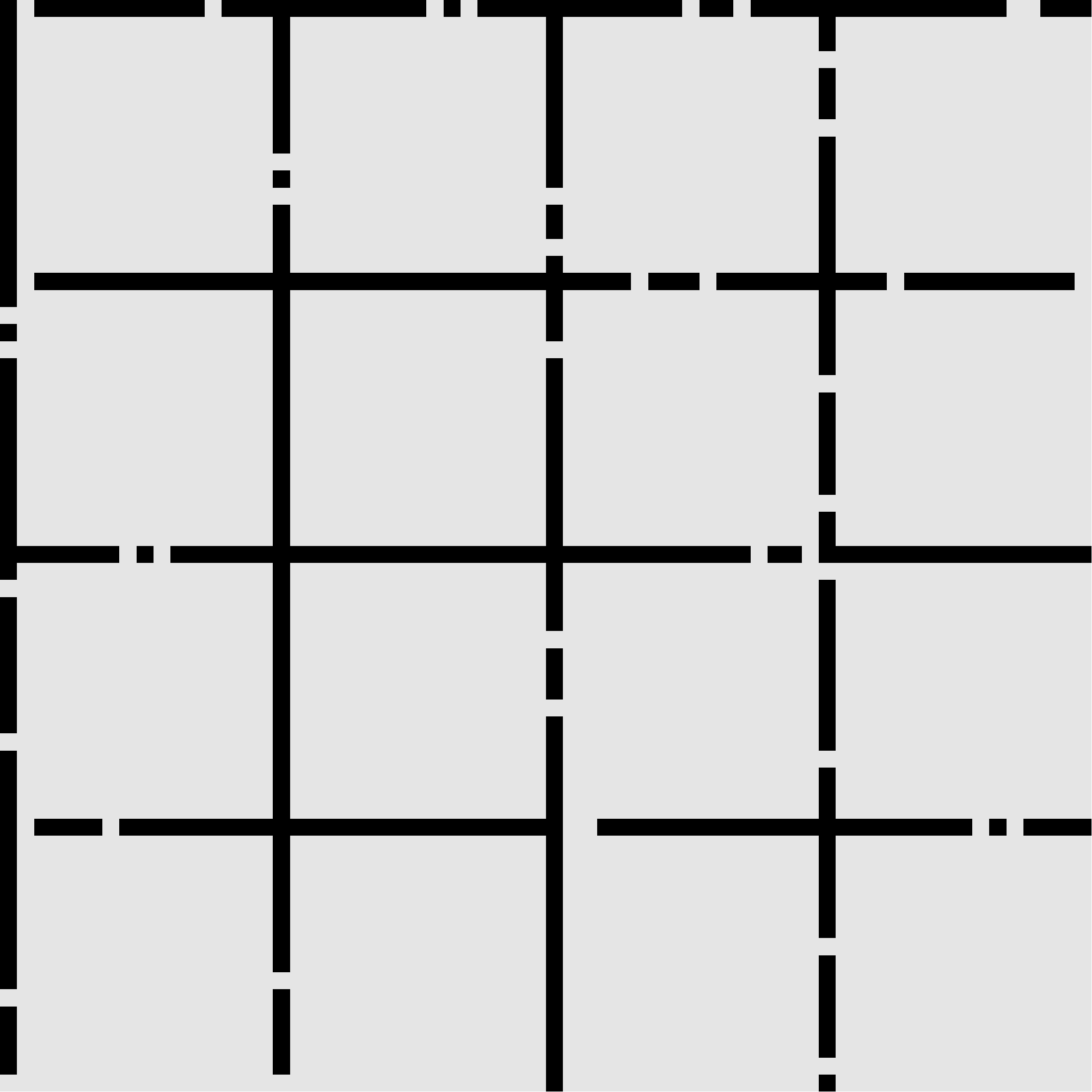}{1,000}{3,646}&
      \figcol{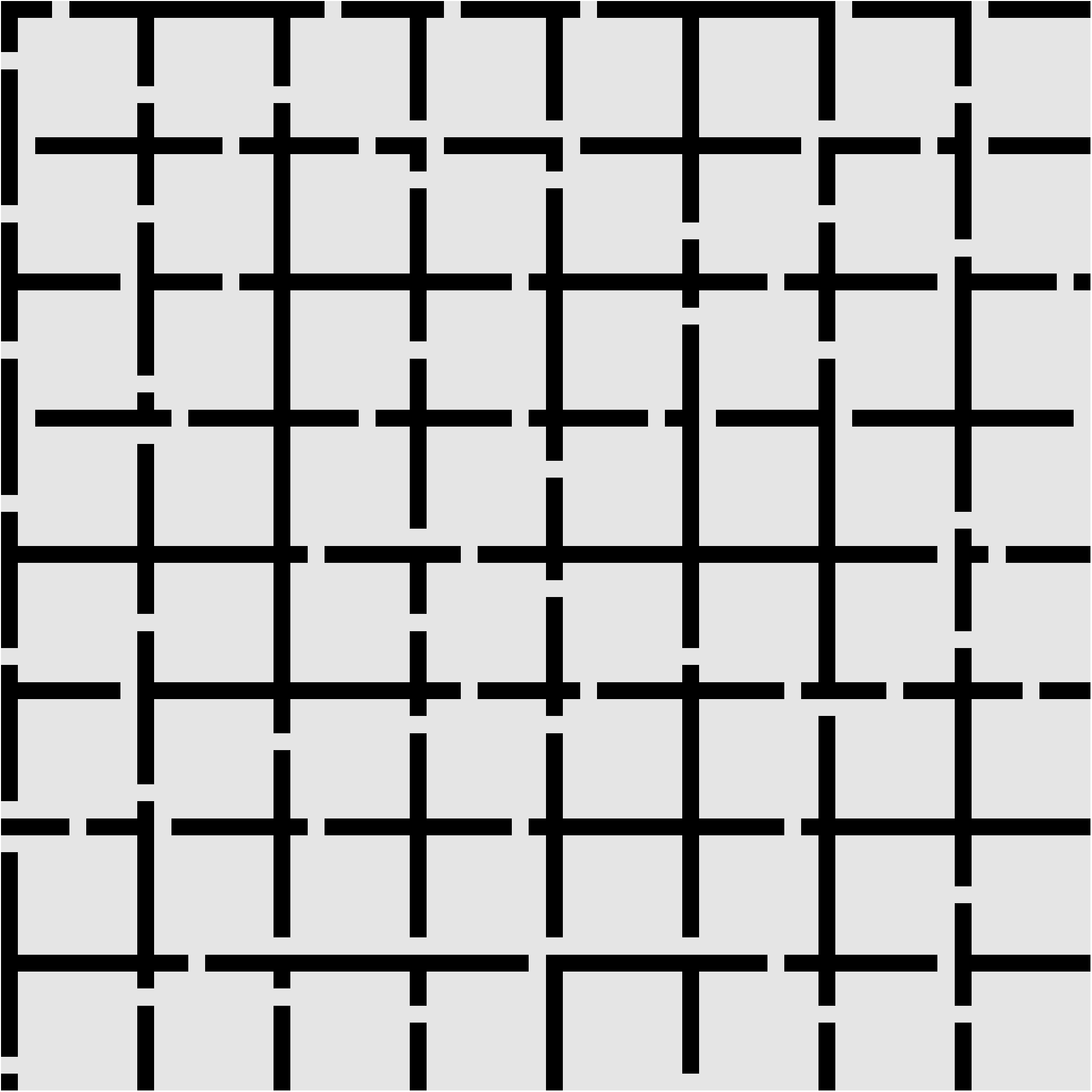}{1,000}{3,232}&
      \figcol{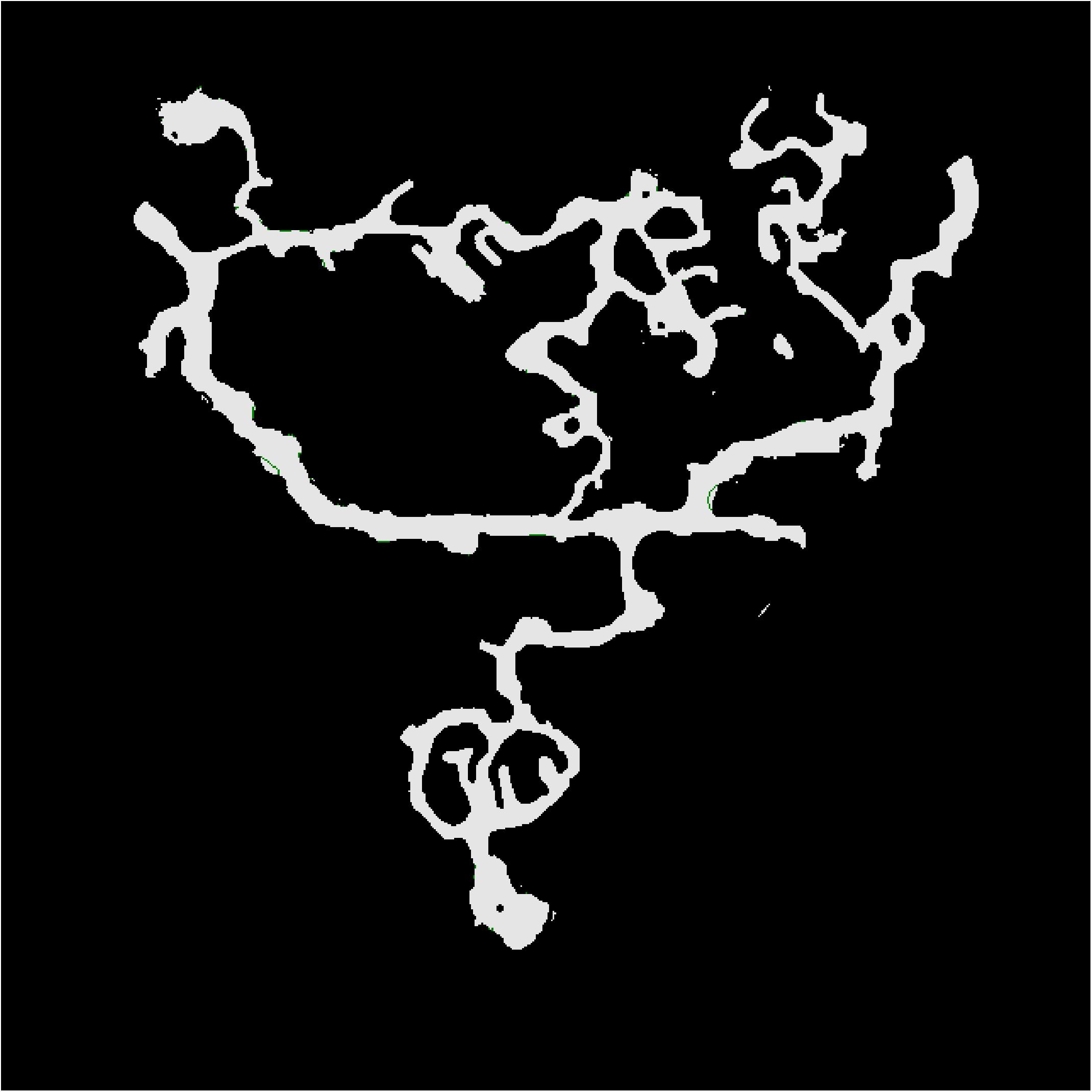}{1,000}{34,020}&
      \figcol{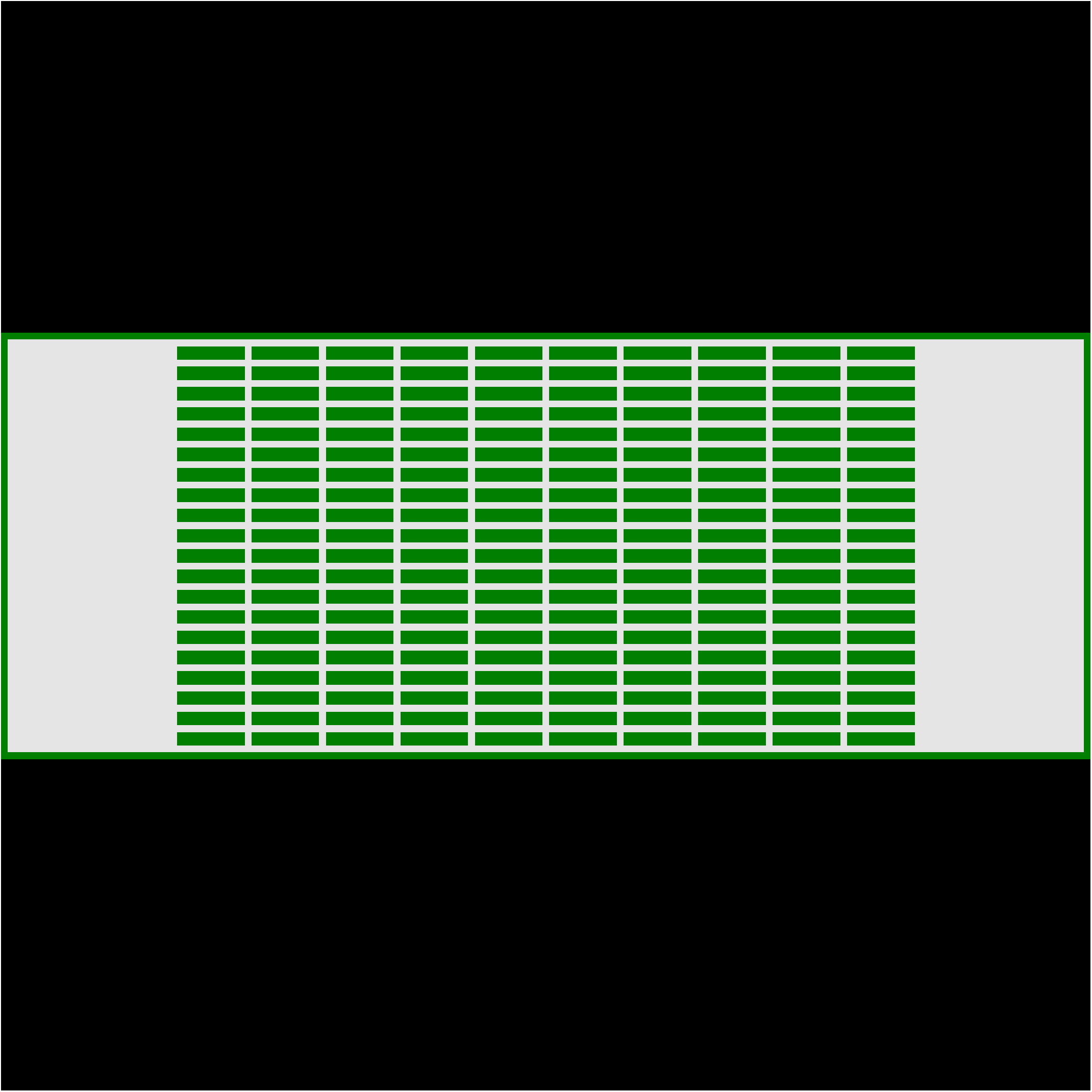}{1,000}{5,699}&
      \figcol{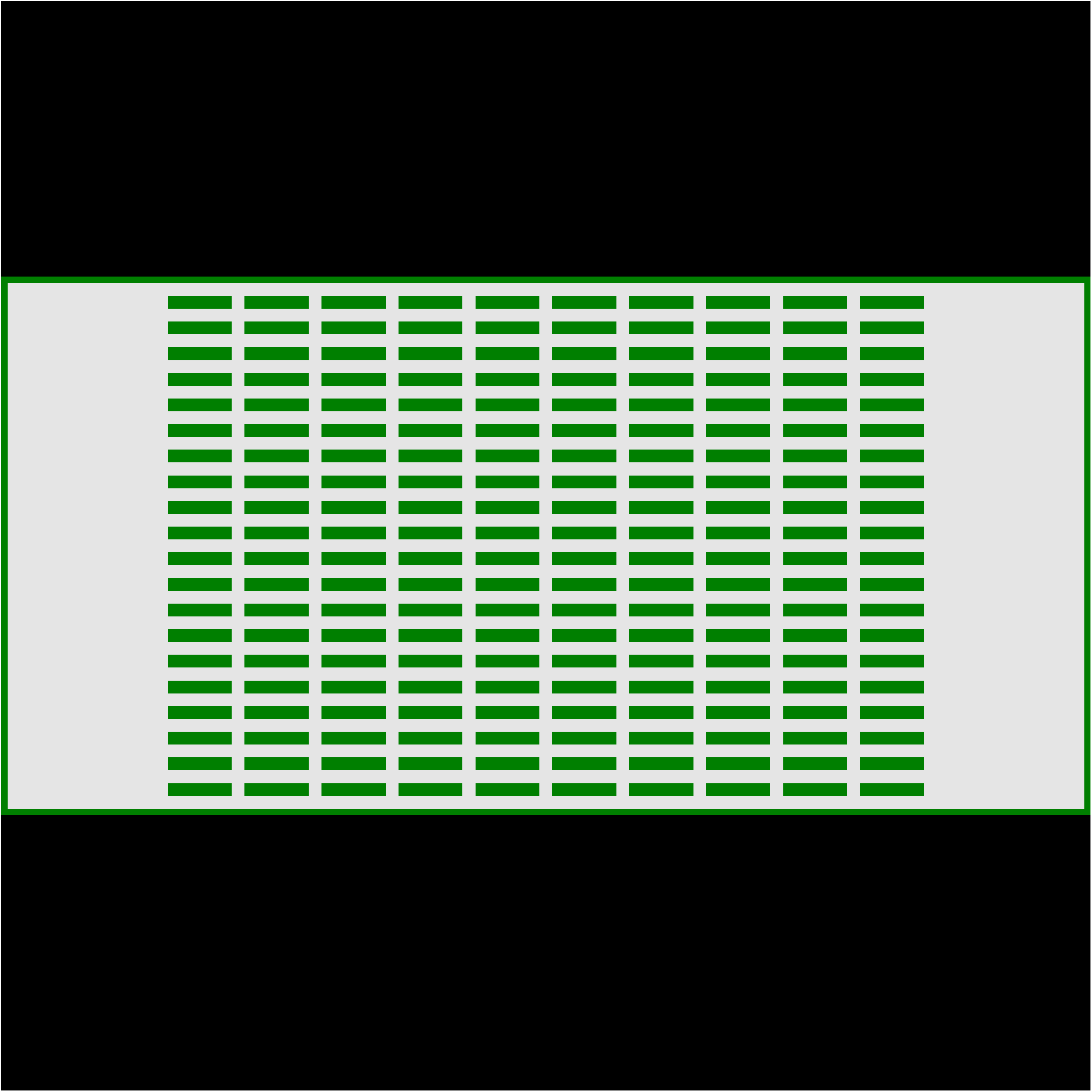}{1,000}{9,776}&
      \figcol{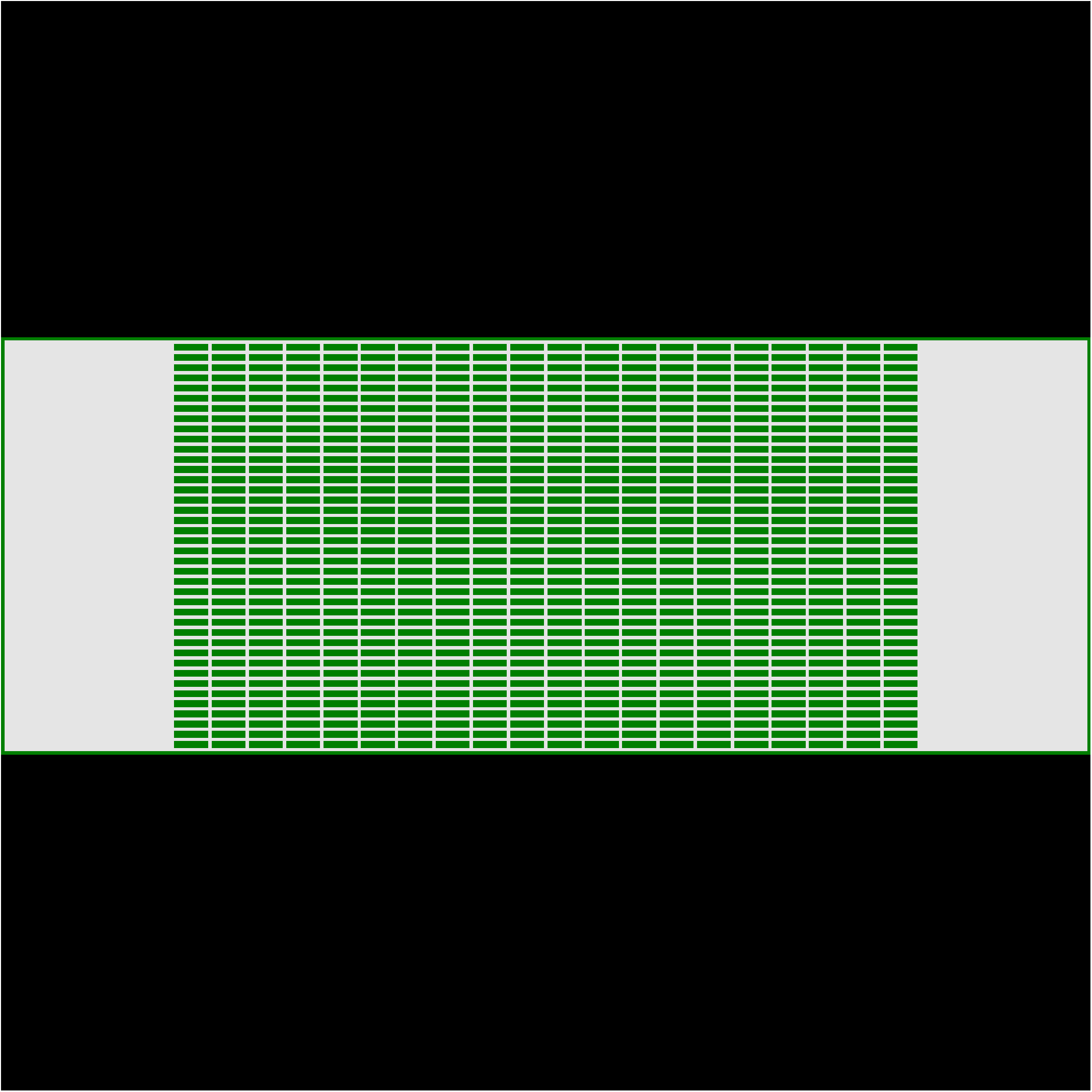}{1,000}{22,599}&
      \figcol{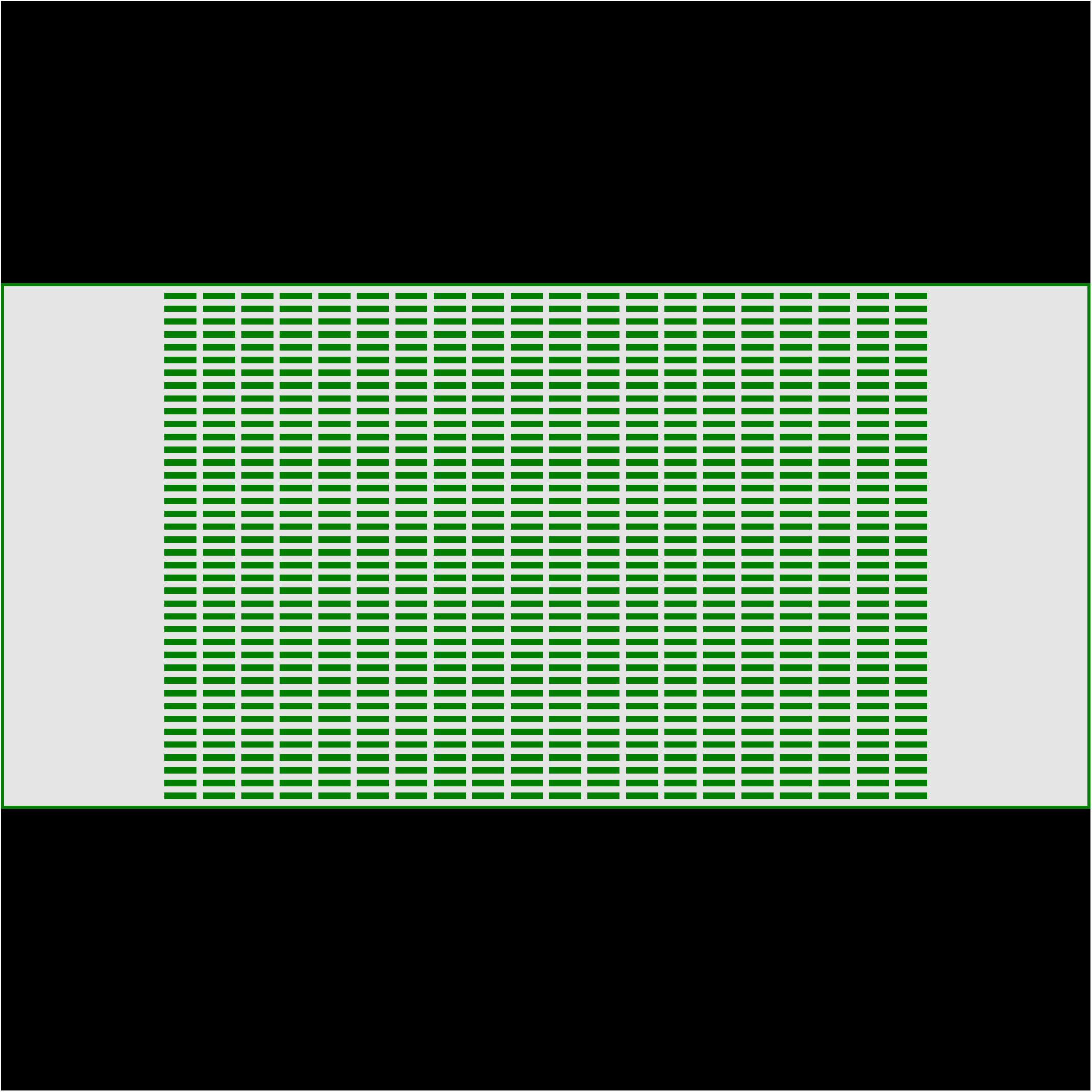}{1,000}{38,756}
      \\
      \multicolumn{18}{c}{instance}
    \end{tabular}
    \includegraphics[width=1.0\linewidth]{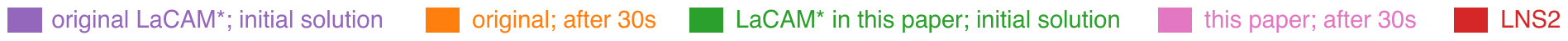}
    \caption{
    Solution quality improvements.
    {\normalfont
    Each chart illustrates the solution quality across 25 instances using a four-connected grid map, retrieved from the MAPF benchmark~\cite{stern2019def}.
    Solution quality is assessed through the \emph{sum-of-loss}, quantifying the total number of actions that agents do not remain at their goals.
    The scores are normalized by division with their corresponding trivial lower bound, i.e., the sum of the shortest path lengths between the agents' start and goal vertices.
    Smaller scores are preferable ($\downarrow$), with the minimum is one.
    For each grid, instances underwent evaluation with the maximum number of agents as specified in the benchmark, mostly a thousand agents.
    The instances are ordered based on the initial solution quality of the original \lacamstar.
    Since \lacamstar is an anytime algorithm, both the initial solution quality and the quality at the \SI{30}{\second} time limit are presented.
    Improved performance over the original \lacamstar is represented by the difference between orange and pink edges.
    For reference, the scores of LNS2~\cite{li2022mapf}, an incomplete suboptimal approach, are included when instances were solved within the time limit (510/800 instances; 64\%).
    The original \lacamstar solved all instances, while the improved version encountered failure in one instance of \mapname{maze-32-32-4}.
    The flowtime results corresponding to this figure are available in \cref{fig:result-soc}.
    }
    }
    \label{fig:result-top}
  \end{figure*}
}

To overcome these limitations, this paper explores various \emph{engineering} techniques for enhancing \lacamstar, drawing inspiration from the literature on MAPF and search methodologies.
An overview of each technique is as follows.
\begin{itemize}
\item \Cref{sec:random-insert} introduces a non-deterministic search node extraction to escape search-stuck situations.
  This accelerates the refinement process.
\item \Cref{sec:scatter} introduces an effective utilization of precomputed paths that are spatially dispersed.
  This aids in discovering better initial solutions.
\item \Cref{sec:mccg} introduces a Monte Carlo-style successor generation method.
  This aids in discovering better initial solutions.
\item \Cref{sec:refiner} introduces the dynamic incorporation of alternative solutions during the search.
  This boosts the refinement.
\item \Cref{sec:recursive} introduces the recursive use of \lacamstar to find alternative solutions.
  Combined with the technique in \cref{sec:refiner}, this accelerates the refinement.
\end{itemize}
While each technique constitutes a modest enhancement, their combined employment significantly boosts the performance of \lacamstar, as evidenced in \cref{fig:result-top,fig:result-soc}.
In such challenging cases, not only do state-of-the-art optimal algorithms~\cite{li2021pairwise,lam2022branch} exhibit complete failure, but even bounded suboptimal approaches~\cite{wagner2015subdimensional,li2021eecbs} and affordable prioritized planning~\cite{silver2005cooperative,okumura2022priority} largely falter.
Meanwhile, the original \lacamstar succeeded in solving \emph{all} of them within the \SI{30}{\second} timeout.
This paper enhances the appeal of \lacamstar by empowering it to generate near-optimal solutions.
For instance, the improved version achieved an average cost reduction of approximately 30\% in the \mapname{random-32-32-20} scenario with 409 agents.
Based on these empirical achievements, we believe that this work opens a new frontier for real-time, large-scale, and near-optimal MAPF, an area that has been difficult to tackle.

Below, the paper provides preliminaries in \cref{sec:prelim}, subsequently presenting each technique in order.
Each technique is evaluated within its dedicated section and collectively evaluated in \cref{sec:everything}.
The supplementary material appears at \url{https://kei18.github.io/lacam3/}.

\section{Preliminaries}
\label{sec:prelim}

We use the following notations.
$S[k]$ denotes the $k$-th element of the collection $S$, where the index starts at one.
For convenience, we use $\bot$ as an ``undefined'' sign.
The \dist function returns the shortest path length between two vertices on a graph.

\subsection{Problem Definition}
An \emph{MAPF instance} is defined by a graph $G = (V, E)$, a set of agents $A = \{1, \ldots, n\}$, a tuple of distinct start vertices $\mathcal{S} = (s_i \in V)^{i \in A}$ and goal vertices $\T = (g_i \in V)^{i \in A}$.
A \emph{configuration} $\Q = (v_1, v_2, \ldots, v_n) \in V^n$ is a tuple of locations for all agents, where $\Q[i] = v_i$ is the location of agent $i \in A$.
For instance, $\mathcal{S}$ and $\T$ represent the start and goal configurations, respectively.
A configuration \Q has a \emph{vertex collision} when there is a pair of agents $i \neq j$ such that $\Q[i] = \Q[j]$.
Two configurations $X$ and $Y$ have an \emph{edge collision} when there is a pair of agents $i \neq j$ such that $X[i] = Y[j] \land Y[i] = X[j]$.
Let $\neigh(v)$ denote a set of vertices adjacent to $v \in V$.
Two configurations $X$ and $Y$ are \emph{connected} when $Y[i] \in \neigh(X[i]) \cup \{X[i]\}$ for all $i \in A$, and there are neither vertex nor edge collisions in $X$ and $Y$.
Given an MAPF instance, a \emph{solution} is a sequence of configurations $\Pi = (\Q_0, \Q_1, \ldots, \Q_k)$, such that $\Q_0 = \mathcal{S}$, $\Q_k = \T$, and any two consecutive configurations in $\Pi$ are connected.
As a solution quality metric, this paper considers minimizing \emph{sum-of-loss}, the number of agent actions that do not stay at goals.
Formally, it is defined by $\sum_{t=0}^{k-1}\cost_e \bigl(\Q_t, \Q_{t+1}\bigr)$, where $\cost_e(X, Y) \defeq |\{ i \in A \mid \lnot(X[i] = Y[i] = g_i)\}|$.

{
  \setlength{\tabcolsep}{1pt}
  \newcommand{\figcol}[3]{
    \begin{minipage}{0.057\linewidth}
      \centering
      \begin{tikzpicture}
      \node[anchor=south east]() at (0, 2.95) {\tiny $#2$};
      \node[anchor=south east]() at (0, 2.78) {\tiny $#3$};
      \node[anchor=south east]() at (0, 2.15) {\includegraphics[width=0.57\linewidth,right]{fig/raw/maps/#1.pdf}};
      \node[anchor=south east]() at (0, 0) {\includegraphics[width=1.0\linewidth,height=2.0\linewidth]{fig/raw/soc_#1.pdf}};
      \node[anchor=south west, rotate=90, text=black]() at (-0.05, 0.15) {\scriptsize \textcolor{black}{\mapname{#1}}};
      \end{tikzpicture}
    \end{minipage}
  }
  \newcommand{\ytitle}{
  \begin{minipage}{0.01\linewidth}
  \rotatebox[origin=c]{90}{\footnotesize $\leftarrow$~~flowtime / LB\hspace{1.2cm}}
  \end{minipage}
  }
  \begin{figure*}[t!]
    \centering
    \begin{tabular}{crrrrrrrrrrrrrrrr}
      \ytitle&
      \figcol{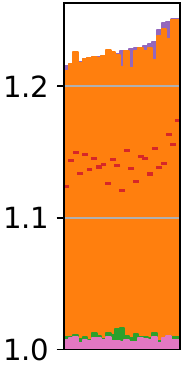}{|A|=1,000}{|V|=47,540}&
      \figcol{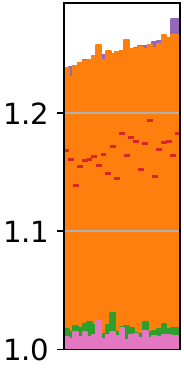}{1,000}{47,768}&
      \figcol{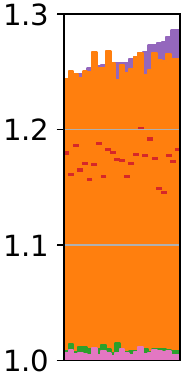}{1,000}{47,240}&
      \figcol{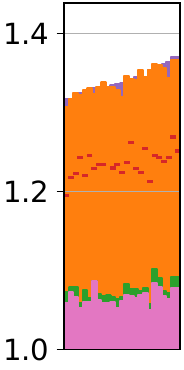}{1,000}{43,151}&
      \figcol{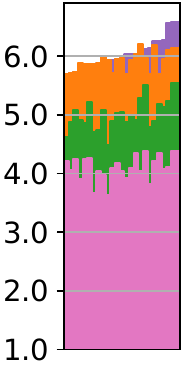}{1,000}{2,445}&
      \figcol{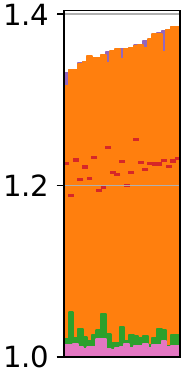}{1,000}{28,178}&
      \figcol{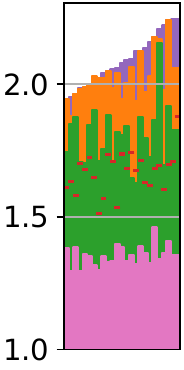}{128}{256}&
      \figcol{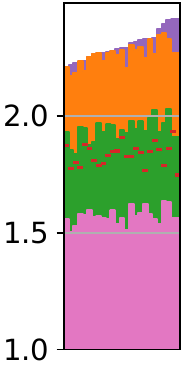}{512}{1,024}&
      \figcol{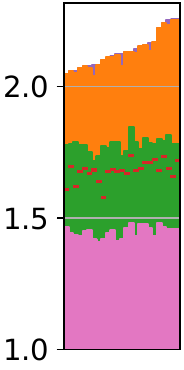}{1000}{2,304}&
      \figcol{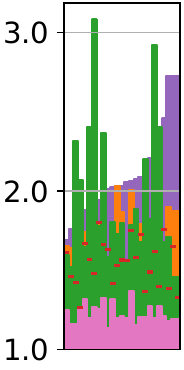}{32}{64}&
      \figcol{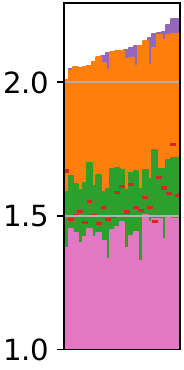}{1,000}{7,461}&
      \figcol{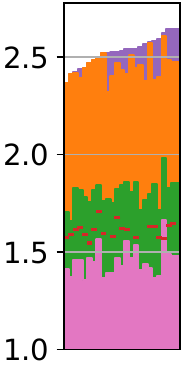}{|A|=1,000}{|V|=8,959}&
      \figcol{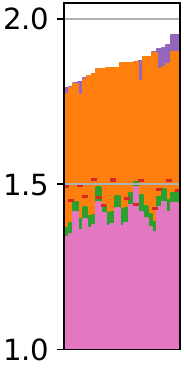}{1,000}{14,784}&
      \figcol{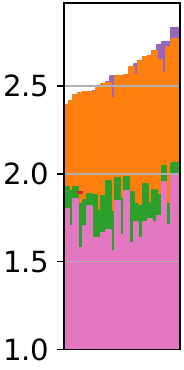}{1,000}{10,021}&
      \figcol{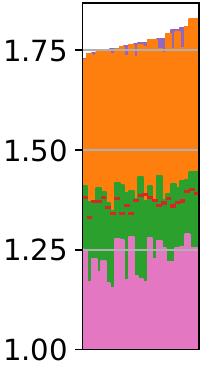}{1,000}{14,818}&
      \figcol{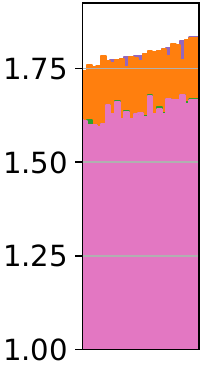}{1,000}{10,858}
      \\
      \ytitle&
      \figcol{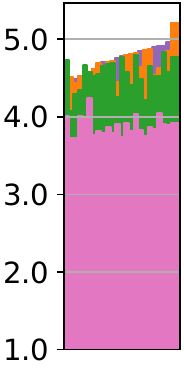}{|A|=333}{|V|=666}&
      \figcol{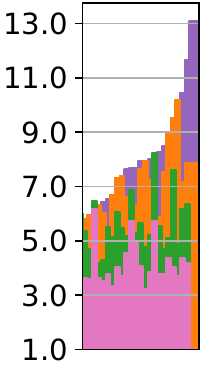}{395}{790}&
      \figcol{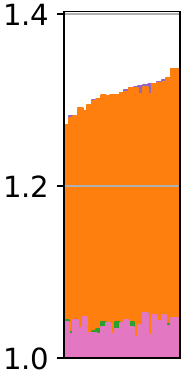}{1,000}{96,603}&
      \figcol{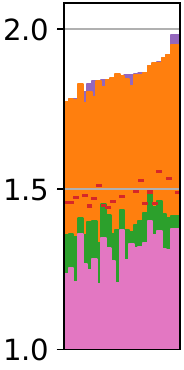}{1,000}{13,214}&
      \figcol{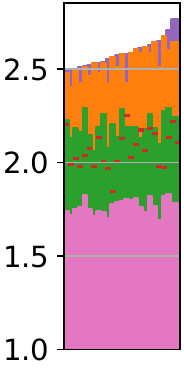}{461}{922}&
      \figcol{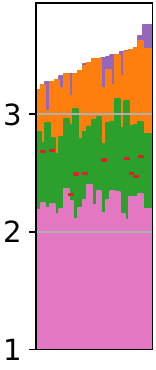}{409}{819}&
      \figcol{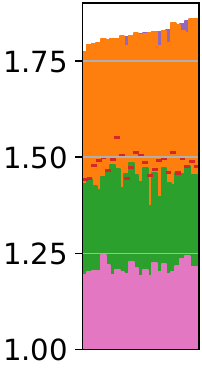}{1,000}{3,687}&
      \figcol{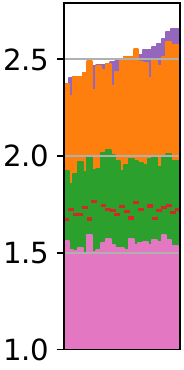}{1,000}{3,270}&
      \figcol{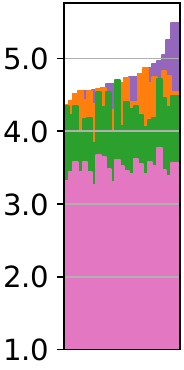}{341}{682}&
      \figcol{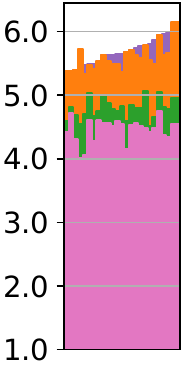}{1,000}{3,646}&
      \figcol{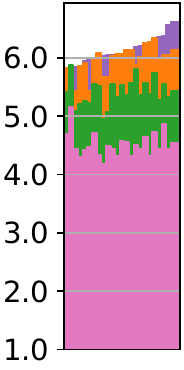}{1,000}{3,232}&
      \figcol{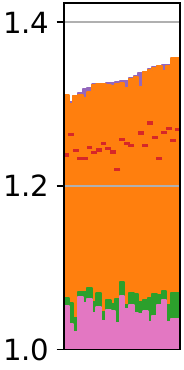}{1,000}{34,020}&
      \figcol{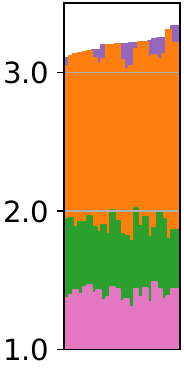}{1,000}{5,699}&
      \figcol{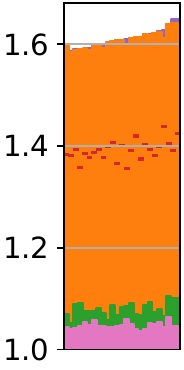}{1,000}{9,776}&
      \figcol{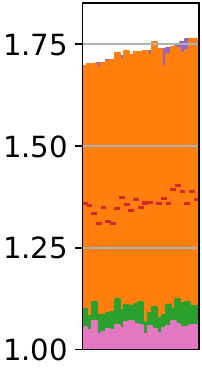}{1,000}{22,599}&
      \figcol{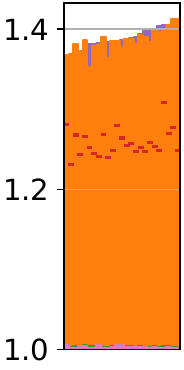}{1,000}{38,756}
      \\
      \multicolumn{17}{c}{instance}
    \end{tabular}
    \includegraphics[width=1.0\linewidth]{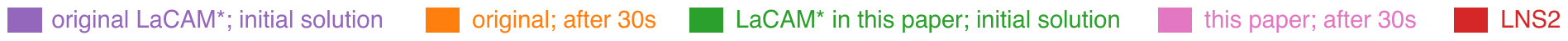}
    \caption{
    Solution quality improvements of flowtime (aka. sum-of-costs).
    {\normalfont
    See also the caption in \cref{fig:result-top}.
    Flowtime is a widely-used metric to evaluate MAPF solutions, computed as the sum of the earliest timesteps that each agent permanently stays at its destination.
    Although \lacamstar primarily focuses on minimizing the sum-of-loss metric, the resultant solutions also excel in terms of flowtime.
    }
    }
    \label{fig:result-soc}
  \end{figure*}
}

\paragraph{Remarks.}
Finding sum-of-loss optimal solutions is NP-hard;
the proof in \cite{yu2013structure} is applicable.
Another popular metric to evaluate solution quality is the \emph{flowtime} (aka. \emph{sum-of-costs}), which accounts for the sum of the timesteps in which each agent stops at its goal and remains there permanently.
This metric is history-dependent for a sequence of configurations, making its representation in terms of cumulative transition costs intricate.
However, due to the similarity in their representation, we empirically observe that the pursuit of near-optimal solutions in terms of sum-of-loss also yield near-optimal outcomes for flowtime, as evidenced in \cref{fig:result-top,fig:result-soc}.

\subsection{LaCAM$^{(\ast)}$}
Given start and goal configurations, \emph{LaCAM}~\cite{okumura2023lacam} is a graph pathfinding algorithm with the search space defined by the configurations and their interconnections.
LaCAM is complete;
it returns a solution for solvable instances within a finite timeframe, otherwise reports the non-existence.
\emph{\lacamstar}~\cite{okumura2023lacam2} is its anytime variant, designed to eventually find the shortest path for cumulative transition costs, even when the transition cost (i.e., $\cost_e$) is not in the sum-of-loss form.
\Cref{algo:lacam} provides an overview of \lacamstar, with the anytime components shaded.
For clarity, two subprocedures have been omitted from the pseudocode and are detailed in the appendix.
This subsection first describes the concept of LaCAM, followed by the extensions used in \lacamstar and the pseudocode.

\subsubsection{Overview}
Similar to the general search scheme such as \astar, LaCAM performs the search by sequentially processing \emph{search nodes} one by one, which are stored within an \open list.
Each node corresponds to a configuration \Q and maintains a pointer to a parent node with another configuration $\Q'$ connected to \Q.
Upon encountering the goal configuration during the search, a solution is derived by backtracking of parent pointers.
The structure of the \open list dictates the search progression.
This study uses the \emph{stack} structure, thereby LaCAM is described similarly to depth-first search (DFS).

Primary deviations from the general search scheme arise from the method of generating successors.
\emph{LaCAM generates a maximum of one successor node upon invoking a node.}
In essence, it produces successors in a \emph{lazy} fashion, and a node may be invoked multiple times during the search.
The details are explained below.

\subsubsection{Lazy Constraints Addition}
In LaCAM, a search node for a configuration \Q also encompasses the management of \emph{constraints}, which delineate the procedure for generating a successor configuration.
A constraint is a pair $i \in A$ and $v \in \neigh(\Q[i]) \cup \{ \Q[i] \}$.
When this constraint is specified, a successor configuration must satisfy the condition that agent-$i$ occupies vertex $v$.
By carefully revising the specified constraints with each invocation of a node, it becomes possible to systematically generate all interconnected configurations stemming from that particular node.

More specifically, a search node of LaCAM preserves a tree structure of constraints, where its root has no constraints.
Other tree nodes correspond to a constraint and encode multiple constraints by tracing a path to the root.
During each invocation of the node, a leaf node $\C$ is chosen from this tree in a breadth-first manner.
The tree will then grow incrementally by creating child tree nodes for $\C$.
These new tree nodes must specify an agent that is different from any ancestor as their constraint, otherwise, the children will not be added.
After updating the tree, \C is used in a successor generation to dictate the relevant constraints.

\subsubsection{Configuration Generation}
The final element of LaCAM is how to generate a successor configuration.
\emph{This is achieved through the implementation of adaptive versions of other MAPF algorithms}, designed to produce a fixed-length sequence of configurations, as exemplified in~\cite{silver2005cooperative,li2021lifelong,okumura2022priority}.
If the configuration generator is promising in producing configurations that are close to the goal configuration, LaCAM can greatly reduce the number of node generations, which is a bottleneck in planning domains with large branching factors such as MAPF.
From an empirical standpoint, the \emph{priority inheritance with backtracking (PIBT)} algorithm~\cite{okumura2022priority}, explained in \cref{sec:pibt}, demonstrated outstanding performance for this sake.

\subsubsection{Attaching Eventual Optimality}
\lacamstar builds upon LaCAM with two extensions:
\emph{(i)}~it continues the search even after encountering the goal configuration, and
\emph{(ii)}~it rewrites parent pointers as required.
A solution is always obtainable by backtracking after encountering the goal configuration.
Furthermore, it is ensured that an optimal solution is constructed through backtracking when the \open list becomes empty.
The objective of the rewriting process is to maintain the optimal path from the start configuration to each configuration in the already explored search space.
This task is efficiently executed through the utilization of an adaptive version of Dijkstra's algorithm with additional search node components, namely, a set of nodes with a connected configuration and cost-to-come (i.e., g-value).

\subsubsection{Pseudocode}
\Cref{algo:lacam} embodies the concept discussed thus far.
The creation of constraints is encapsulated within\\\textsc{LowLevelSearch} (\cref{algo:lacam:lowlevel-search}), and the rewriting procedure is represented by \textsc{DijkstraUpdate} (\cref{algo:lacam:dijkstra}).
Refer to the appendix for further details of these subprocedures.
The configuration generation is denoted as \funcname{configuration\_generator} (\cref{algo:lacam:configuration-generator}).

The algorithm functions as follows.
Following the initialization (Lines~\ref{algo:lacam:init}--\ref{algo:lacam:init-node-insert}), \lacamstar proceeds by sequentially processing a search node (Lines~\ref{algo:lacam:loop-start}--\ref{algo:lacam:loop-end}).
Invoked nodes are not immediately discarded.
Rather, they are set aside after the assessment of potential constraint combinations is complete (\cref{algo:lacam:discard}).
Upon the constraint determination (\cref{algo:lacam:lowlevel-search}), the configuration generation ensues (\cref{algo:lacam:configuration-generator}).
If the configuration is not known for the search, a new node is created (Lines~\ref{algo:lacam:new-configuraiton}--\ref{algo:lacam:update-neighbor});
otherwise, the rewriting process occurs (Lines~\ref{algo:lacam:known-configuraiton}--\ref{algo:lacam:knonw-configuration-end}).
Finally, \lacamstar concludes by reporting a result (Lines~\ref{algo:lacam:solution-found}--\ref{algo:lacam:fin}).

{
  \renewcommand{\S}{\m{\mathcal{S}}}
  \begin{algorithm}[t!]
    \caption{LaCAM$^{\gl{\ast}}$}
    \label{algo:lacam}
    \begin{algorithmic}[1]
    \Input{MAPF instance, \gl{edge cost $\cost_e$, admissible heuristic $\h$}}
    \Output{solution, \nosolution, or \failure}
    \Notation{\gl{$\f(\N) \defeq \N.g + \h(\N.\config)$; $\spadesuit \defeq (\N\goal \neq \bot)$}}
    \State initialize \open (stack), \explored (hash table);~~$\N\goal \leftarrow \bot$
    \label{algo:lacam:init}
    \State
      $\N\init \leftarrow \begin{aligned}[t]
        \bigl\langle
        \config: \S,
        \parent: \bot,
        \tree: \llbracket \bot \rrbracket~\text{(queue)},
        \gl{\neighbors: \emptyset, g: 0}
        \bigr\rangle\end{aligned}$
    \State $\open.\push(\N\init)$;~~$\explored[\S] = \N\init$
    \label{algo:lacam:init-node-insert}
    \While{$\open \neq \emptyset~\land~\lnot\funcname{interrupt}()$}
    \label{algo:lacam:loop-start}
    \State $\N \leftarrow \open.\funcname{top}()$
    \label{algo:lacam:extracting}
    \IfSingle{$\N.\config = \T$}{$\N\goal \leftarrow \N$}
    \IfSingleGl{$\spadesuit \land \f(\N\goal) \leq \f\left(\N\right)$}{$\open.\pop()$;~\Continue}
    \label{algo:lacam:pruning}
    \IfSingle{$\N.\tree = \emptyset$}{$\open.\pop()$;~\Continue}
    \label{algo:lacam:discard}
    \State $\C \leftarrow \Call{LowLevelSearch}{\N}$
    \label{algo:lacam:lowlevel-search}
    \Comment{constraints generation}
    \State $\Q\new \leftarrow \funcname{configuration\_generator}(\N, \C)$
    \label{algo:lacam:configuration-generator}
    \IfSingle{$\Q\new = \bot$}{\Continue}
    \If{$\explored[\Q\new] = \bot$}
    \Comment{new configuration}
    \label{algo:lacam:new-configuraiton}
    \StateGl{$g \leftarrow \N.g + \cost_e(\N.\config, \Q\new)$}
    \label{algo:lacam:compute-g}
    \State $\N\new \leftarrow \begin{aligned}[t]
      \langle
      \config: \Q\new,
      \parent: \N,
      \tree: \llbracket \bot \rrbracket,
      \gl{\neighbors: \emptyset, g: g}
      \bigr\rangle\end{aligned}$
    \State $\open.\push(\N\new)$;\;$\explored[\Q\new] = \N\new$
    \StateGl{$\N.\neighbors.\append(\N\new)$}
    \label{algo:lacam:update-neighbor}
    \Else
    \Comment{known configuration}
    \label{algo:lacam:known-configuraiton}
    \StateGl{$\N.\neighbors.\append(\explored[\Q\new])$}
    \StateGl{$\Call{DijkstraUpdate}{\N}$}
    \label{algo:lacam:dijkstra}
    \State $\open.\push(\explored[\Q\new]~\text{or}~\explored[\S])$
    \label{algo:lacam:reinsert}
    \EndIf
    \label{algo:lacam:knonw-configuration-end}
    \EndWhile
    \label{algo:lacam:loop-end}
    \If{$\spadesuit$}
    \label{algo:lacam:solution-found}
    \Return $\backtrack(\N\goal)$
    \ElsIf{$\open = \emptyset$}
    \Return \nosolution
    \Else~\Return \failure
    \EndIf
    \label{algo:lacam:fin}
    \end{algorithmic}
  \end{algorithm}
}

\subsubsection{Implementation Tips}
The pseudocode also includes several enhancements beyond the minimal implementation.

When a known configuration is encountered, \cref{algo:lacam:reinsert} reinstates the corresponding node to the top of the \open list, a strategy that has been empirically demonstrated to improve solution quality~\cite{okumura2023lacam}.
Furthermore, instead of solely reinserting the rediscovered node, inserting the initial node with a very low probability (e.g., $0.001$) proves advantageous in circumventing search-stuck situations~\cite{okumura2023lacam2}.

\Cref{algo:lacam:pruning} engages in pruning by discarding a node if the sum of its cost-to-come and the estimated cost-to-go surpasses the present best solution cost.
Such a node does not contribute to quality improvement.
The estimation is facilitated by an \emph{admissible} heuristic $\h: V^n \mapsto \mathbb{R}{\geq 0}$, which is admissible if the estimation consistently remains equal to or lower than the true value.
For instance, $\sum_{i \in A}\dist(Q[i], g_i)$ serves as an effective choice.

\subsection{PIBT}
\label{sec:pibt}
As an effective implementation, the original LaCAM$^{(\ast)}$ papers~\cite{okumura2023lacam,okumura2023lacam2} employed PIBT~\cite{okumura2022priority} as their configuration generator.
Following their success, this study also uses PIBT.
The following offers a basic understanding of PIBT, sufficient for subsequent discussions.

Given a configuration \Q, let $\phi_i$ denote an enumeration of the possible subsequent vertices for agent-$i$, i.e., $C_i \defeq \neigh(\Q[i]) \cup \{\Q[i] \}$.
PIBT is conceptualized as a function that takes $\phi_1, \ldots, \phi_n$ as inputs and yields a connected configuration with \Q.
The assignment for agent-$i$ is executed while adhering to the order established by $\phi_i$.
In other words, PIBT allocates the first vertex from $\phi_i$ in the absence of collisions.
The original PIBT arranges vertices in ascending order based on $\dist(u, g_i)$, where $u \in C_i$, using randomization when resolving tie-breaking situations.
Hence, it intends to prioritize the allocation of a vertex from $C_i$ that is closest to the goal $g_i$.

While the original LaCAM~\cite{okumura2023lacam} employs a vanilla PIBT, the subsequent study~\cite{okumura2023lacam2} remarkably improves the planning success rate by devising the construction of $\phi_i$.
Specifically, drawing inspiration from the ``swap'' operation in rule-based MAPF methods~\cite{luna2011push,de2014push}, this adjustment involves reversing the order of $\phi_i$ when two agents require a location exchange within narrow passages.
This can mostly prevent LaCAM from getting stuck in bottleneck situations where the search process repeatedly visits some specific configurations.
The implementation of \lacamstar in this study also uses this trick.

\subsection{Evaluation Environment}
All experiments in this paper were performed on a 28-core desktop PC with Intel Core i9-10940X \SI{3.3}{\giga\hertz} CPU, and \SI{64}{\giga\byte} RAM.
Each MAPF instance was solved sequentially; we never did a concurrent run.
Meanwhile, some techniques in this study assume parallel computation, and they were implemented using multi-threading.

Hereinafter, we assess each technique under three scenarios, all of which employ a four-connected grid map obtained from the MAPF benchmark~\cite{stern2019def}:
\emph{(i)}~\scenname{random}: \mapname{random-32-32-20} with 409 agents, \emph{(ii)}~\scenname{empty}: \mapname{empty-48-48} with 1,000 agents, and \emph{(iii)}~\scenname{chantry}: \mapname{ht_chantry} with 1,000 agents.
Images of these grids are shown in \cref{fig:result-top}.
For each scenario, a total of 25 instances were considered.
Since the implementation of \lacamstar incorporates non-deterministic elements, such as tie-breaking in PIBT, each method was executed for every instance using four distinct random seeds.
Hence, the results below are for 100 trial runs.
Each trial had a runtime limit of \SI{10}{\second}.
The objective is to find the best possible solutions for the sum-of-loss within this time limit.

\section{Non-Deterministic Node Extraction}
\label{sec:random-insert}
We start engineering by a lightweight technique, namely, rethinking the order of node extraction from \open at \cref{algo:lacam:extracting}.
\lacamstar can quickly find an initial solution, facilitated by a DFS-like mechanism utilizing a standard stack as \open.
However, sustaining the same mechanism after the initial solution discovery is not mandatory.
In fact, with the vanilla stack, if the search encounters problematic configurations such that most of their neighbors have worse costs than the present solution cost, the search will stay on the nodes for a very long time due to the pruning at \cref{algo:lacam:pruning}.
Consequently, the solution refinement of \lacamstar stops.

\subsection{Method}
A straightforward approach to tackle this issue is to select a node from \open that differs from the original extraction (i.e., the top entry) with a small probability (e.g., $0.01$), provided the search has already found an initial solution.
Such techniques that rely on non-determinism have been seen in other MAPF methods~\cite{cohen2018rapid,andreychuk2018two}, as well as \lacamstar itself (\cref{algo:lacam:reinsert}).
There are design choices about which node to extract instead.
We empirically tested two candidates: \emph{(i)}~a start node, i.e., $\explored[\mathcal{S}]$, denoted as ``restart,'' partially inspired by~\cite{richter2010joy}, or \emph{(ii)}~a randomly selected node from \open, denoted as ``random.''
Note that these modifications do not break the optimal search structure of \lacamstar.

\subsection{Evaluation}
\Cref{fig:result-insert} shows that both the ``restart'' and ``random'' strategies contribute to steady improvements in the solution quality compared to the original \lacamstar, even though they started from the same initial solutions.
This is because, as anticipated, the non-deterministic node extraction can escape from search-stuck situations.

\section{Space Utilization Optimization}
\label{sec:scatter}

The solution quality of \lacamstar depends heavily on the underlying configuration generator, i.e., PIBT.
Recalling the process, PIBT endeavors to allocate the first vertex in the enumeration $\phi_i$ to agent-$i$ as much as possible, given a configuration $Q$ and $\phi_i$, which lists the potential subsequent positions of agent-$i$.
In terms of implementation, $\phi_i$ is constructed in an ascending order based on the distance to the goal vertex $g_i$.
This design ensures that PIBT aims to steer agents along the shortest paths to their respective goals.
However, the shortest path connecting two vertices is typically not unique;
numerous paths may yield identical costs.
This section delves into the methodology for selecting between these equivalent paths, a decision that can significantly influence solution quality.

\Cref{fig:scatter} provides a concrete motivating example.
Consider agent-1, which has at least two shortest paths originating from its start vertex, designated as path-$a$ and path-$b$.
Employing PIBT to build $\phi_1$ around path-$a$ would impose a higher cost on agent-2 compared to the alternative of selecting path-$b$.
For larger teams of agents, this phenomenon exerts a noteworthy influence on solution quality.

A better construction strategy of $\phi_i$ is imperative to circumvent such issues.
In pursuit of this objective, this section introduces \emph{space utilization optimization (SUO)}~\cite{han2022optimizing} to PIBT, a concept that involves the use of a precomputed collection of less congested paths as an effective aid to MAPF algorithms.%
\footnote{
A recent preprint paper~\cite{chen2023traffic} also employs a similar strategy to optimize $\phi_i$ in PIBT for lifelong MAPF, while this paper provides a simpler yet effective method.
}

\subsection{Method}
\subsubsection{Finding Scattered Paths}
Such scattered paths encourage agents to evenly traverse spatiotemporal locations.
Specifically, we are interested in finding a collection of paths $\pathsuo = [ \pi_1, \ldots, \pi_n ]$, one for each agent, where \emph{(i)}~$\pi_i$ is the shortest path from $s_i$ to $g_i$, while \emph{(ii)}~minimizing the number of collisions in \pathsuo.
For example, in \cref{fig:scatter}, path-$b$ should be selected for agent-$1$, because otherwise a collision with the unique shortest path for agent-$2$ will be included.

Identifying \pathsuo is computationally demanding, given its nature as a combinatorial optimization task, wherein the objective is to find the optimal collection of paths, with each agent potentially possessing numerous shortest paths.
Rather than seeking an exact solution, \cite{han2022optimizing} introduces an approximation that can be computed within a reasonable timeframe.
The pseudocode for this approach is presented in \cref{algo:scatter}.
For the present moment, the hyperparameter $m$ can be regarded as zero.
Then, \cref{algo:scatter} iteratively computes individual shortest paths while diligently evading collisions with other paths, until no further updates are noted.
The single-agent replanting process at \Cref{algo:scatter:replanning} can be executed through \astar.
Note that analogous algorithmic methodologies can be seen in other MAPF methods, such as finding individual paths for the initial solution in conflict-based search~\cite{sharon2015conflict}, or deconfliction of infeasible solutions through large neighborhood search~\cite{li2022mapf}.
Let \pathsuo represent the acquired scattered paths stemming from \cref{algo:scatter}, with $\pathsuo[i]$ signifying the corresponding path attributed to agent-$i$.

{
  \setlength{\tabcolsep}{0.5pt}
  \newcommand{\figcol}[3]{
    \begin{minipage}{0.32\linewidth}
      \centering
      {\small \hspace{0.3cm}\scenname{#3}}
      \includegraphics[width=1.0\linewidth,height=0.7\linewidth]{fig/raw/insert-#1_#2}
    \end{minipage}
  }
  \begin{figure}[t!]
    \centering
    \begin{tabular}{cccc}
      \begin{minipage}{0.03\linewidth}\rotatebox[origin=c]{90}{cost / LB}\end{minipage}&
      \figcol{random-32-32-20}{409}{random}&
      \figcol{empty-48-48}{1000}{empty}&
      \figcol{ht_chantry}{1000}{chantry}
      \\
      &\multicolumn{3}{c}{runtime (sec)}
      \\
      \multicolumn{4}{c}{\includegraphics[height=0.35cm]{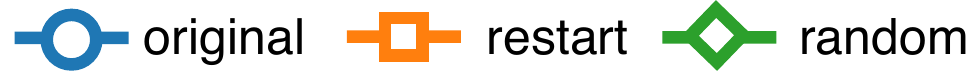}}
    \end{tabular}
    \caption{
    Effect of non-deterministic node extraction after finding initial solutions.
    {\normalfont
    ``original'' refers to the original \lacamstar~\cite{okumura2023lacam2}.
    All methods solved every trial within \SI{10}{\second}.
    For each second, the average sum-of-loss within the solved instances up to that point is depicted.
    The scores are normalized by dividing by their trivial lower bound, $\sum_i\dist(s_i, g_i)$.
    The charts include the average scores of the initial solutions.
    Furthermore, they encompass the minimum and maximum scores achieved within the solved instances up to that specific time, visualized with transparent regions.
    For each method, we tested several probabilities for non-deterministic node extraction, and show those with consistently provided outcomes with smaller costs (``restart'': $0.01$, ``random'': $0.01$).
    }
    }
    \label{fig:result-insert}
  \end{figure}
}

{
  \begin{figure}[t!]
    \centering
    \includegraphics[width=0.9\linewidth]{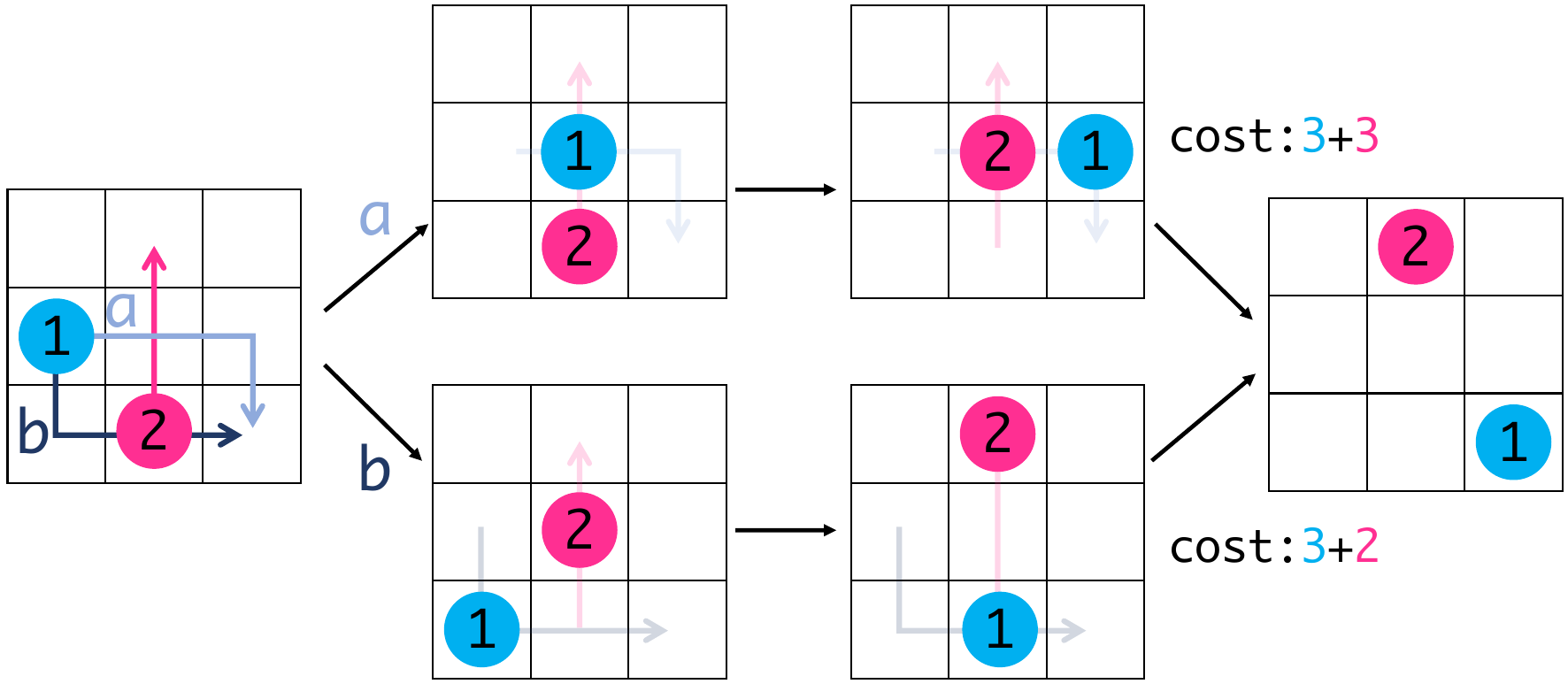}
    \caption{Motivation to compute spatially dispersed paths.}
    \label{fig:scatter}
  \end{figure}
}

{
  \begin{algorithm}[t!]
    \caption{Finding scattered paths}
    \label{algo:scatter}
    \begin{algorithmic}[1]
      \Input{MAPF instance;\;\; \textbf{output:}~\pathsuo (path for each agent)}
      \Params{$m \in \mathbb{N}_{\geq 0}$}
      \State $\Pi \leftarrow [ \bot, \dots, \bot ]$
      \While{$\Pi$ is updated in the last iteration}
      \For{$i \in A$}
      \State $\Pi[i] \leftarrow$
      a path from $s_i$ to $g_i$, such that:
      \begin{itemize}
      \item has a length equal to or less than $\dist(s_i, g_i) + m$
      \item minimizes \#collisions with other paths in $\Pi$
      \end{itemize}
      \label{algo:scatter:replanning}
      \EndFor
      \EndWhile
      \State \Return $\Pi$
    \end{algorithmic}
  \end{algorithm}
}

\subsubsection{Exploitation}
\label{sec:scatter:exploitation}
Upon the creation of \pathsuo, it serves as an effective guide for the PIBT configuration generator, elaborated as follows.
Given a configuration \Q, the construction of $\phi_i$ involves sorting vertices in $\neigh(\Q[i]) \cup \{ \Q[i] \}$ in an ascending order based on $f_i$, which is defined as:
\begin{align*}
  f_i(u) = \begin{cases}
    0 & \text{if an ordered edge}~(\Q[i], u)~\text{exists in}~\pathsuo[i] \\
    \dist(u, g_i) & \text{otherwise}
  \end{cases}
\end{align*}
This allows PIBT to prioritize the use of scattered paths in \pathsuo and enables it to solve an example of \cref{fig:scatter} optimally.

\subsubsection{Allowing Non-Shortest Paths}
The original SUO paper~\cite{han2022optimizing} presupposes the shortest paths at \cref{algo:scatter:replanning}.
However, our empirical observations have indicated that relaxing this condition can yield further improvements to solution quality.
To elaborate, with the introduction of a hyperparameter $m \in \mathbb{N}_{\geq 0}$, we grant \cref{algo:scatter:replanning} the margin to seek out a path whose length extends up to $\dist(s_i, g_i) + m$ for agent-$i$.
The computation of such a path can still be executed by \astar.
Larger $m$ should be avoided due to the increase in search space, but a reasonable $m$ allows the single-agent pathfinding to search for more scattered paths.

\subsection{Evaluation}
\Cref{fig:result-scatter} vividly illustrates the impact of employing spatially scattered paths in improving the quality of the initial solutions.
Employing \pathsuo with the shortest paths already leads to a reduction in solution cost, yet greater improvements are achieved by permitting the utilization of non-shortest paths.
It is worth noting that the computational overhead for computing the scattered paths does exist but remains modest.
In fact, the derivation of initial solutions for scenarios involving 1,000 agents was attained within seconds.
Note that the non-deterministic node extraction detailed in \cref{sec:random-insert} was \emph{not} applied to isolate the influence of SUO.

\section{Monte-Carlo Configuration Generation}
\label{sec:mccg}
To assign a vertex to each agent, PIBT employs an enumeration based on distances to the goal and resolves ties \emph{non-deterministically}.
Even with an identical input configuration, PIBT has the potential to produce distinct configurations.
Among these configurations, certain ones might yield better initial solutions than others.

Motivated by this insight, this section presents the \emph{Monte-Carlo configuration generator}.
This concept draws partial inspiration from the Monte-Carlo tree search~\cite{kocsis2006bandit,swiechowski2023monte}, and its applicability extends to non-PIBT configuration generators.

\subsection{Method}
Transitioning from the original \lacamstar is straightforward;
substitute the configuration generator at \cref{algo:lacam:configuration-generator} in \cref{algo:lacam} with \cref{algo:mccg}.
The procedure entails gathering $k$ configurations through $k$ iterations of the generator.
Subsequently, the best configuration is chosen based on the sum of the cost-to-come and the estimated cost-to-go.

\paragraph{Parallel Computation}
\Cref{algo:mccg:parallel} can take advantage of \emph{parallel computation}, allowing for the simultaneous generation of numerous configurations.
This can improve the quality of initial solutions without incurring a significant slowdown compared to the original \lacamstar.
Given the prevalence of multi-core machines, leveraging the capabilities of parallel computing presents a logical strategy for the advancement of powerful MAPF methodologies.

\subsection{Evaluation}
\Cref{fig:result-parallel} shows the effect of the Monte-Carlo configuration generator with multi-threading.
As before, the techniques introduced in previous sections were \emph{not} applied here.
Even with just ten Monte-Carlo runs, \lacamstar exhibits an enhanced capability to refine the quality of initial solutions.
As the number of samples $k$ is increased, a continued enhancement in solution quality is observed.
However, there have been instances of failure when $k=100$.
This occurrence can be attributed to the behavior of the configuration generator, which becomes more deterministic with an increased number of Monte-Carlo samples.
Then \lacamstar may find it difficult to escape from the wrong search direction to reach the goal.
This phenomenon also accounts for the delay in acquiring initial solutions in the \scenname{random} scenario under the $k=100$ condition.
The empirical results of speedup by multi-threading is available in the appendix.

{
  \setlength{\tabcolsep}{0.5pt}
  \newcommand{\figcol}[3]{
    \begin{minipage}{0.32\linewidth}
      \centering
      {\small \hspace{0.3cm}\scenname{#3}}
      \includegraphics[width=1.0\linewidth,height=0.7\linewidth]{fig/raw/scatter-#1_#2}
    \end{minipage}
  }
  \begin{figure}[t!]
    \centering
    \begin{tabular}{cccc}
      \begin{minipage}{0.03\linewidth}\rotatebox[origin=c]{90}{cost / LB}\end{minipage}&
      \figcol{random-32-32-20}{409}{random}&
      \figcol{empty-48-48}{1000}{empty}&
      \figcol{ht_chantry}{1000}{chantry}
      \\
      &\multicolumn{3}{c}{runtime (sec)}
      \\
      \multicolumn{4}{c}{\includegraphics[height=0.35cm]{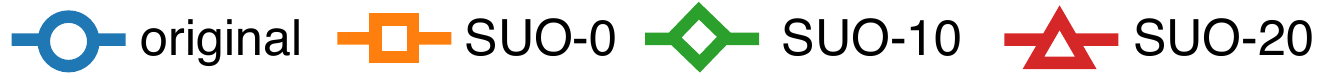}}
    \end{tabular}
    \caption{
    Effect of space utilization optimization.
    {\normalfont
    See also the caption of \cref{fig:result-insert}.
    The runtime includes computing \pathsuo.
    ``SUO-$x$'' means to set $m$ in \cref{algo:scatter} to $x$.
    All methods solved all trials.
    }
    }
    \label{fig:result-scatter}
  \end{figure}
}

{
  \renewcommand{\P}{\m{\mathcal{P}}}
  \begin{algorithm}[t!]
    \caption{\funcname{monte\_carlo\_configuration\_generator}}
    \label{algo:mccg}
    \begin{algorithmic}[1]
      \Input{search node $\N$, constraints $\C$;\;\;\textbf{output:}~configuration}
      \Params{$k \in \mathbb{N}_{>0}$ (number of samples)}
      \State $\P \leftarrow \bigl(\funcname{configuration\_generator}(\N, \C)\bigr)^k$
      \label{algo:mccg:parallel}
      \Comment{possibly parallel}
      \State \Return $\mathit{argmin}_{\Q \in \P} \cost_e(\N.\config, \Q) + \h(\Q)$
    \end{algorithmic}
  \end{algorithm}
}

{
  \setlength{\tabcolsep}{0.5pt}
  \newcommand{\figcol}[3]{
    \begin{minipage}{0.32\linewidth}
      \centering
      {\small \hspace{0.3cm}\scenname{#3}}
      \includegraphics[width=1.0\linewidth,height=0.7\linewidth]{fig/raw/parallel-#1_#2}
    \end{minipage}
  }
  \begin{figure}[t!]
    \centering
    \begin{tabular}{cccc}
      \begin{minipage}{0.03\linewidth}\rotatebox[origin=c]{90}{cost / LB}\end{minipage}&
      \figcol{random-32-32-20}{409}{random}&
      \figcol{empty-48-48}{1000}{empty}&
      \figcol{ht_chantry}{1000}{chantry}
      \\
      &\multicolumn{3}{c}{runtime (sec)}
      \\
      \multicolumn{4}{c}{\includegraphics[height=0.35cm]{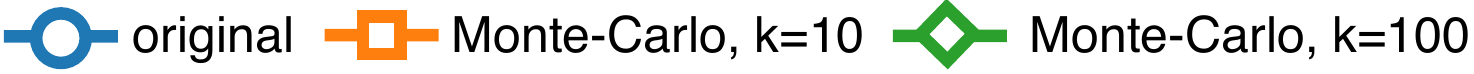}}
    \end{tabular}
    \caption{Effect of Monte-Carlo configuration generation.
    {\normalfont
     5/100 attempts in \scenname{random} failed with $k=100$;
     all others succeeded.
    }}
    \label{fig:result-parallel}
  \end{figure}
}

\section{Dynamic Incorporation of Alternative Solutions}
\label{sec:refiner}
In optimization problems, once a feasible solution is found, improving its quality using known solutions is generally achievable.
This procedure is commonly referred to as \emph{local search}.
For instance, a traveling salesman problem (TSP) has many algorithms available to improve solution quality~\cite{rego2011traveling}.
Analogous to TSP, multiple methods exist for refining solution quality in MAPF~\cite{surynek2013redundancy,de2014push,okumura2021iterative,li2021anytime}.
While these approaches may not ensure convergence to optimal solutions, when combined with \lacamstar, they can offer a theoretical assurance that optimal solutions will eventually be obtained.

\subsection{Method}
During the search of \lacamstar, suppose an alternate solution $\Pi$ is discovered by some means.
Such scenarios can occur when a different MAPF algorithm, operating concurrently with \lacamstar, finds a solution.
Another practical illustration, particularly assumed in this study, involves initiating a local search from the present \lacamstar solution using parallel computation.
To integrate a new solution $\Pi$ into the search process, \lacamstar can employ \cref{algo:incoporation} and then proceed with its search.
In essence, this entails sequentially incorporating configurations from $\Pi$ to create a new search node or rewrite the search tree structure.

Each time a new solution is fed in, \cref{algo:incoporation} should be called.
This operation preserves the integrity of \lacamstar's complete and optimal search structure.
Furthermore, it efficiently propels the search forward when the provided solution surpass the present one.

This technique is more powerful than it seems because it offers theoretical support for various suboptimal MAPF algorithms;
i.e., when coupled with \lacamstar, \emph{it is theoretically possible to eventually find optimal solutions from arbitrary suboptimal solutions}, provided that the solution cost accounts for cumulative transition costs.

\subsection{Evaluation}

\subsubsection{Implementation}
The experiments used an iterative refinement framework for arbitrary MAPF solutions~\cite{okumura2021iterative,li2021anytime}.
This method initially selects a subset of agents and exclusively applies MAPF algorithms to them, yielding refined solutions, while treating unselected agents as dynamic obstacles.
If the refined solution proves superior, it supplants the original one.
At the implementation level, a random subset of $1-30$ agents was chosen for each refinement iteration.
The refinement process employed prioritized planning~\cite{erdmann1987multiple,silver2005cooperative} combined with SIPP~\cite{phillips2011sipp} for single-agent pathfinding, as seen in~\cite{li2021scalable}.
Once an initial solution is found by \lacamstar, the refinement procedures, called \emph{refiners}, operated concurrently with \lacamstar through multi-threading, using the initial solution as input.
Upon the completion of a refiner, \cref{algo:incoporation} was invoked to incorporate alternative solutions, followed by the execution of another refiner utilizing the best solution at that time.

\subsubsection{Results}
The effect of this technique is depicted in \cref{fig:result-refiner} with a single refiner or four refiners.
With the inclusion of a refiner, \lacamstar effectively improves solutions in the short term.
Introducing additional refiners can further improve solution quality under specific circumstances.
The refinement in the \scenname{chantry} scenario was not as fast as the other two scenarios because the subproblems for the refiners were challenging due to the complicated grid structure.

{
  \renewcommand{\S}{\m{\mathcal{S}}}
  \begin{algorithm}[t!]
    \caption{Incorporating new solution}
    \label{algo:incoporation}
    \begin{algorithmic}[1]
      \Procedure{IncoporateNewSolution}{$\Pi \defeq (\Q_0, \Q_1, \ldots, \Q_k)$}
      \State $\N\from \leftarrow \explored[\Q_0]$;~$\N\to \leftarrow \bot$
      \For{\Q in $\Q_1, \Q_2, \ldots, \Q_k$}
      \If{$\explored[\Q] = \bot$}
      \State create $\N\new$ using Lines~\ref{algo:lacam:compute-g}--\ref{algo:lacam:update-neighbor} in \cref{algo:lacam}
      \State $\N\from \leftarrow \N\new$
      \Else
      \State $\N\to \leftarrow \explored[\Q]$;\;$\N\from.\neighbors.\append(\N\to)$
      \State $\Call{DijkstraUpdate}{\N\from}$;\;$\N\from \leftarrow \N\to$
      \EndIf
      \EndFor
      \EndProcedure{}
    \end{algorithmic}
  \end{algorithm}
}

{
  \setlength{\tabcolsep}{0.5pt}
  \newcommand{\figcol}[3]{
    \begin{minipage}{0.32\linewidth}
      \centering
      {\small \hspace{0.3cm}\scenname{#3}}
      \includegraphics[width=1.0\linewidth,height=0.7\linewidth]{fig/raw/refiner-#1_#2}
    \end{minipage}
  }
  \begin{figure}[t!]
    \centering
    \begin{tabular}{cccc}
      \begin{minipage}{0.03\linewidth}\rotatebox[origin=c]{90}{cost / LB}\end{minipage}&
      \figcol{random-32-32-20}{409}{random}&
      \figcol{empty-48-48}{1000}{empty}&
      \figcol{ht_chantry}{1000}{chantry}
      \\
      &\multicolumn{3}{c}{runtime (sec)}
      \\
      \multicolumn{4}{c}{\includegraphics[height=0.4cm]{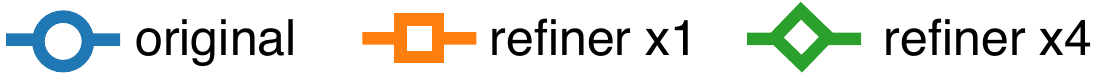}}
    \end{tabular}
    \caption{Effect of incorporating alternative solutions.}
    \label{fig:result-refiner}
  \end{figure}
}

\section{Recursive Call of \lacamstar}
\label{sec:recursive}
Through the techniques introduced thus far, \lacamstar can rapidly find plausible solutions.
As a result, \lacamstar itself can be harnessed to discover alternate solutions for the technique in \cref{sec:refiner}.
Such a recursive utilization of algorithms has already surfaced in several MAPF studies~\cite{wagner2015subdimensional,li2019improved}.

This concept is embodied as follows.
We can generate another MAPF instance based on the present solution by selecting a configuration \Q randomly from it and treating \Q as a new start configuration.
This new instance, where the goal configuration is unchanged from the original, is expected to discover a better-quality solution more readily than the original, owing to that \Q is closer to the goal configuration compared to the start configuration of the original.
Subsequently, another \lacamstar is invoked concurrently to address this new instance.
Upon the completion of this recursive call, the invocation of \cref{algo:incoporation} facilitates the incorporation of its outcomes into the ongoing search process.

\paragraph{Implementation}
Each recursive call was assigned a timeout of \SI{1}{\second}.
The recursive call did not call another \lacamstar.
Our informal observations indicated that a single level of recursion was generally adequate.
The recursive call was implemented as one of the refiners.
More specifically, the recursive call was conducted instead of the standard refiner with a certain probability ($0.2$).
The empirical results of this technique are detailed in \cref{sec:everything}.

\section{Putting Everything Together}
\label{sec:everything}

\subsection{Effect of Combined Use}
This section jointly evaluates the introduced techniques.
\Cref{fig:result-all} shows the results corresponding to the previous evaluations.
The combined use substantially improves solution quality, with the recursive call further amplifying the improvement.
Notably, a nearly 30\% reduction in solution cost was realized in the \scenname{random} scenario.

\Cref{fig:result-ablation} shows the ablation results for the combined use of the techniques in this paper, indicating that all techniques contribute to improving solution quality.
Specifically, SUO and the dynamic incorporation of alternative solutions have a significant impact.

\subsection{Evaluation on Large-Scale MAPF Instances}
We applied the combined use, including the recursive call, to various challenging MAPF instances retrieved from the MAPF benchmark~\cite{stern2019def}.
The benchmark encompasses 33 grid maps, each housing 25 ``random'' scenario files, each of which outlines a list of start-goal pairs.
We procured 800 instances from the benchmark by using 32 grid maps,%
\footnote{
The \mapname{maze-128-128-1} map was excluded due to the original \lacamstar's inability to solve instances featuring the maximum number of agents from this map~\cite{okumura2023lacam2}.
}
each with 25 instances, while adhering to the specified maximum number of agents outlined in the scenario files.
The allotted time limit stood at \SI{30}{\second}, in alignment with~\cite{stern2019def}.
A single random seed was employed during testing.
We also tested LNS2~\cite{li2022mapf}, an incomplete suboptimal MAPF algorithm, as reference records.
LNS2 was chosen because it was the only algorithm we were aware of, other than PIBT and LaCAM, that could handle such difficult cases.
Note that LNS2 still failed to solve 290 instances (36\%).

The results are summarized in \cref{fig:result-top} and illustrate the significant improvement in solution quality achieved by the proposed techniques.
Furthermore, we present flowtime, a commonly used metric for assessing MAPF solutions, in \cref{fig:result-soc}.
Although \lacamstar is designed to minimize the sum-of-loss, these techniques also result in a dramatic reduction in the flowtime metric.

Additional observations include the following.
In certain instances where agents are not densely placed, e.g., \mapname{Berline_1_256}, the enhanced version generates solutions close to optimal, despite the significant disparities present in the original \lacamstar.
Moreover, the quality of the initial solution is already comparable to or better than that of LNS2 in most cases.
Meanwhile, the enhancements do not invariably guarantee cost reduction, especially about the initial solution quality, e.g., \mapname{empty-8-8}.
Some cases are still difficult to refine; see \mapname{maze-128-128-2}.
Note that the enhanced \lacamstar encountered \emph{one} scenario it could not solve in the \mapname{maze-32-32-4} instance for the same reason stated in \cref{sec:mccg}.

\subsection{Evaluation with Ten Thousand Agents}
We finally ventured into an extreme scenario -- MAPF instances involving 10,000 agents.
For this test, we prepared 25 instances using the \mapname{warehouse-20-40-10-2-2} map.
This time, the time limit was set to \SI{5}{\minute}.
Such instances were unsolvable in a realistic timeframe for search-based MAPF algorithms prior to the advent of LaCAM~\cite{okumura2023lacam}.
Given the enormity of the problem instances, we implemented a timeout mechanism for the SUO technique, restricting its execution to half of the time limit;
i.e., \cref{algo:scatter} compelled the premature generation of scattered paths, even before the iterative process stabilized.
Furthermore, a timeout of \SI{10}{\second} was instituted for the recursive call.

Displayed in \cref{fig:result-10k}, the outcomes highlight the superiority of the enhanced version of this paper over the original.
We further assessed multiple versions with specific features omitted to examine the individual contributions to the enhancements achieved.
Among them, the SUO technique demonstrated a substantial impact, despite the initial computational overhead involved.
While the other techniques exhibited discernible performance enhancements for \lacamstar in moderately large instances as presented so far, their effects were less pronounced in the context of these immense instances (hence, omitted from the figure).
For such huge instances, it remains challenging to refine solution quality effectively.

{
  \setlength{\tabcolsep}{1pt}
  \newcommand{\colwidth}{0.315\linewidth}
  \newcommand{\imgwidth}{0.75\linewidth}
  \newcommand{\imgheight}{0.86\linewidth}
  \newcommand{\xcoord}{2.8}
  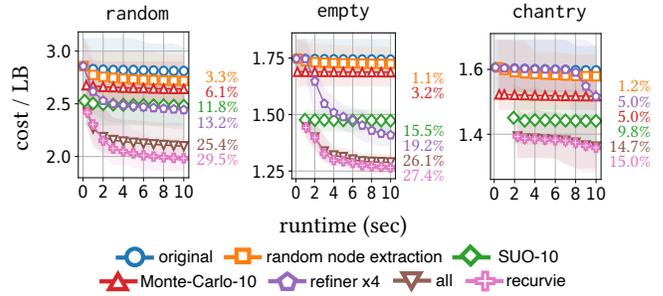
\begin{figure}[t!]
    \centering
    \begin{tabular}{cccc}
      \begin{minipage}{0.03\linewidth}\rotatebox[origin=c]{90}{cost / LB}\end{minipage}&
      \begin{minipage}{\colwidth}
        \centering
        {\small \scenname{random}}
        \begin{tikzpicture}
        \node[anchor=south west]() at (0, 0) {\includegraphics[width=\imgwidth,height=\imgheight]{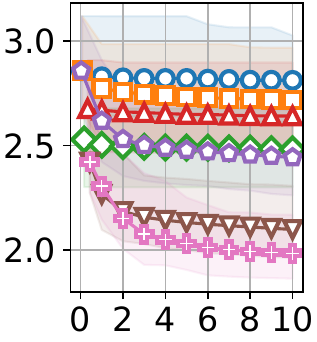}};
        \node[anchor=south east]() at (\xcoord, 1.6) {\footnotesize \textcolor[HTML]{ff7f0e}{3.3\%}};
        \node[anchor=south east]() at (\xcoord, 1.2) {\footnotesize \textcolor[HTML]{2ca02c}{11.8\%}};
        \node[anchor=south east]() at (\xcoord, 1.4) {\footnotesize \textcolor[HTML]{d62728}{6.1\%}};
        \node[anchor=south east]() at (\xcoord, 1.0) {\footnotesize \textcolor[HTML]{9467bd}{13.2\%}};
        \node[anchor=south east]() at (\xcoord, 0.7) {\footnotesize \textcolor[HTML]{8c564b}{25.4\%}};
        \node[anchor=south east]() at (\xcoord, 0.5) {\footnotesize \textcolor[HTML]{e377c2}{29.5\%}};
        \end{tikzpicture}
      \end{minipage}
      &
      \begin{minipage}{\colwidth}
        \centering
        {\small \scenname{empty}}
        \begin{tikzpicture}
        \node[anchor=south west]() at (0, 0) {\includegraphics[width=\imgwidth,height=\imgheight]{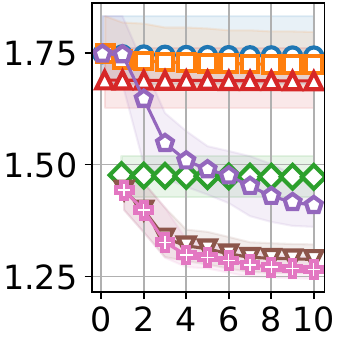}};
        \node[anchor=south east]() at (\xcoord, 1.6) {\footnotesize \textcolor[HTML]{ff7f0e}{1.1\%}};
        \node[anchor=south east]() at (\xcoord, 0.9) {\footnotesize \textcolor[HTML]{2ca02c}{15.5\%}};
        \node[anchor=south east]() at (\xcoord, 1.4) {\footnotesize \textcolor[HTML]{d62728}{3.2\%}};
        \node[anchor=south east]() at (\xcoord, 0.7) {\footnotesize \textcolor[HTML]{9467bd}{19.2\%}};
        \node[anchor=south east]() at (\xcoord, 0.5) {\footnotesize \textcolor[HTML]{8c564b}{26.1\%}};
        \node[anchor=south east]() at (\xcoord, 0.3) {\footnotesize \textcolor[HTML]{e377c2}{27.4\%}};
        \end{tikzpicture}
      \end{minipage}
      &
      \begin{minipage}{\colwidth}
        \centering
        {\small \scenname{chantry}}
        \begin{tikzpicture}
        \node[anchor=south west]() at (0, 0) {\includegraphics[width=\imgwidth,height=\imgheight]{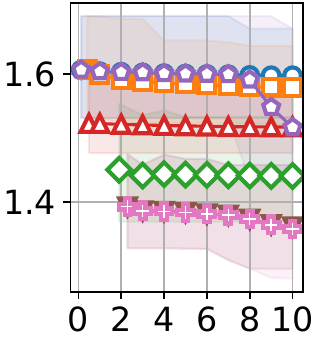}};
        \node[anchor=south east]() at (\xcoord, 1.5) {\footnotesize \textcolor[HTML]{ff7f0e}{1.2\%}};
        \node[anchor=south east]() at (\xcoord, 0.9) {\footnotesize \textcolor[HTML]{2ca02c}{9.8\%}};
        \node[anchor=south east]() at (\xcoord, 1.1) {\footnotesize \textcolor[HTML]{d62728}{5.0\%}};
        \node[anchor=south east]() at (\xcoord, 1.3) {\footnotesize \textcolor[HTML]{9467bd}{5.0\%}};
        \node[anchor=south east]() at (\xcoord, 0.7) {\footnotesize \textcolor[HTML]{8c564b}{14.7\%}};
        \node[anchor=south east]() at (\xcoord, 0.5) {\footnotesize \textcolor[HTML]{e377c2}{15.0\%}};
        \end{tikzpicture}
      \end{minipage}
      \\
      &\multicolumn{3}{c}{runtime (sec)}
      \\
      \multicolumn{4}{c}{\includegraphics[height=0.7cm]{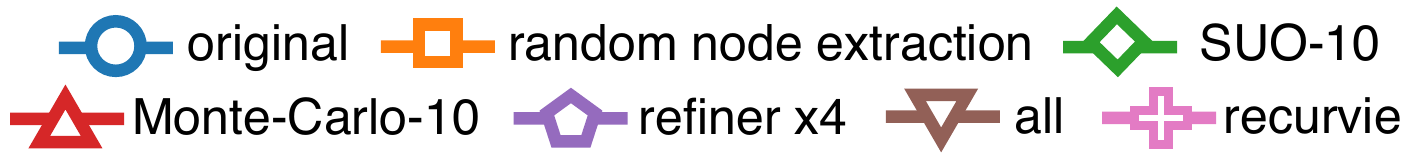}}
    \end{tabular}
    \caption{Effect of combining techniques.
    {\normalfont
    ``all'' denotes a combined use of the techniques in \cref{sec:random-insert}--\ref{sec:refiner}.
    ``recursive'' uses the recursive call of \lacamstar described in \cref{sec:recursive}, in addition to the techniques in ``all.''
    The graphs also show the results of each technique for comparison.
    Their hyperparameters are shown on the label.
    ``all'' and ``recursive'' are also based on these parameters.
    The improvement rates are shown on the right side of each chart, where the baseline is the original \lacamstar (blue circles).
    }}
    \label{fig:result-all}
  \end{figure}
}

{
  \setlength{\tabcolsep}{1pt}
  \newcommand{\colwidth}{0.315\linewidth}
  \newcommand{\imgwidth}{0.75\linewidth}
  \newcommand{\imgheight}{0.86\linewidth}
  \newcommand{\xcoord}{2.8}
  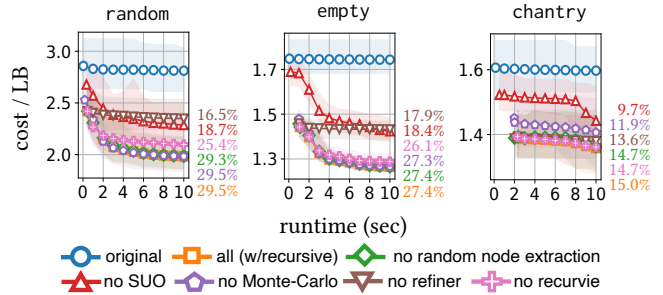
\begin{figure}[t!]
    \centering
    \begin{tabular}{cccc}
      \begin{minipage}{0.03\linewidth}\rotatebox[origin=c]{90}{cost / LB}\end{minipage}&
      \begin{minipage}{\colwidth}
        \centering
        {\small \scenname{random}}
        \begin{tikzpicture}
        \node[anchor=south west]() at (0, 0) {\includegraphics[width=\imgwidth,height=\imgheight]{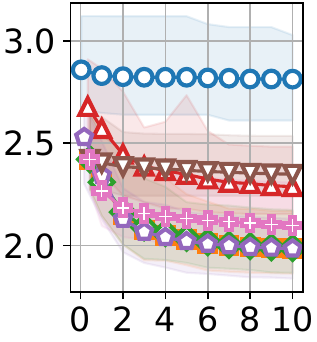}};
        \node[anchor=south east]() at (\xcoord, 0.1) {\footnotesize \textcolor[HTML]{ff7f0e}{29.5\%}};
        \node[anchor=south east]() at (\xcoord, 0.5) {\footnotesize \textcolor[HTML]{2ca02c}{29.3\%}};
        \node[anchor=south east]() at (\xcoord, 0.9) {\footnotesize \textcolor[HTML]{d62728}{18.7\%}};
        \node[anchor=south east]() at (\xcoord, 0.3) {\footnotesize \textcolor[HTML]{9467bd}{29.5\%}};
        \node[anchor=south east]() at (\xcoord, 1.1) {\footnotesize \textcolor[HTML]{8c564b}{16.5\%}};
        \node[anchor=south east]() at (\xcoord, 0.7) {\footnotesize \textcolor[HTML]{e377c2}{25.4\%}};
        \end{tikzpicture}
      \end{minipage}
      &
      \begin{minipage}{\colwidth}
        \centering
        {\small \scenname{empty}}
        \begin{tikzpicture}
        \node[anchor=south west]() at (0, 0) {\includegraphics[width=\imgwidth,height=\imgheight]{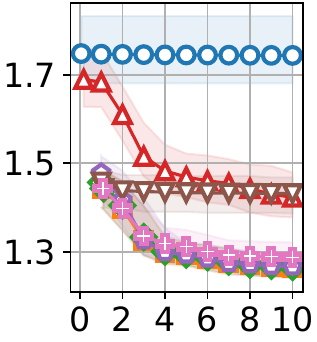}};
        \node[anchor=south east]() at (\xcoord, 0.1) {\footnotesize \textcolor[HTML]{ff7f0e}{27.4\%}};
        \node[anchor=south east]() at (\xcoord, 0.3) {\footnotesize \textcolor[HTML]{2ca02c}{27.4\%}};
        \node[anchor=south east]() at (\xcoord, 0.9) {\footnotesize \textcolor[HTML]{d62728}{18.4\%}};
        \node[anchor=south east]() at (\xcoord, 0.5) {\footnotesize \textcolor[HTML]{9467bd}{27.3\%}};
        \node[anchor=south east]() at (\xcoord, 1.1) {\footnotesize \textcolor[HTML]{8c564b}{17.9\%}};
        \node[anchor=south east]() at (\xcoord, 0.7) {\footnotesize \textcolor[HTML]{e377c2}{26.1\%}};
        \end{tikzpicture}
      \end{minipage}
      &
      \begin{minipage}{\colwidth}
        \centering
        {\small \scenname{chantry}}
        \begin{tikzpicture}
        \node[anchor=south west]() at (0, 0) {\includegraphics[width=\imgwidth,height=\imgheight]{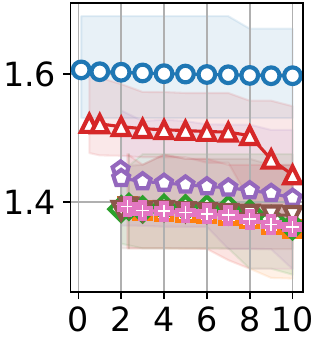}};
        \node[anchor=south east]() at (\xcoord, 0.2) {\footnotesize \textcolor[HTML]{ff7f0e}{15.0\%}};
        \node[anchor=south east]() at (\xcoord, 0.6) {\footnotesize \textcolor[HTML]{2ca02c}{14.7\%}};
        \node[anchor=south east]() at (\xcoord, 1.2) {\footnotesize \textcolor[HTML]{d62728}{9.7\%}};
        \node[anchor=south east]() at (\xcoord, 1.0) {\footnotesize \textcolor[HTML]{9467bd}{11.9\%}};
        \node[anchor=south east]() at (\xcoord, 0.8) {\footnotesize \textcolor[HTML]{8c564b}{13.6\%}};
        \node[anchor=south east]() at (\xcoord, 0.4) {\footnotesize \textcolor[HTML]{e377c2}{14.7\%}};
        \end{tikzpicture}
      \end{minipage}
      \\
      &\multicolumn{3}{c}{runtime (sec)}
      \\
      \multicolumn{4}{c}{\includegraphics[height=0.7cm]{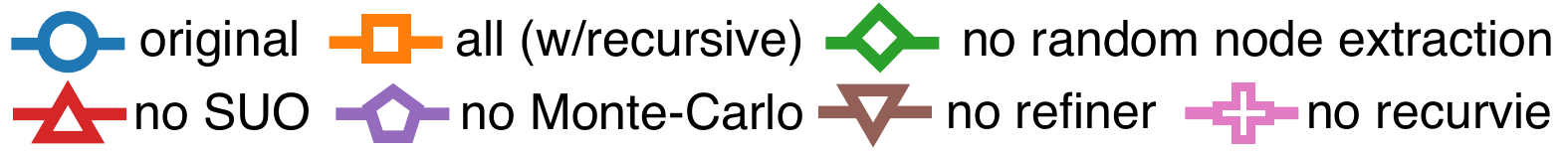}}
    \end{tabular}
    \caption{Ablation study.
    \normalfont{
    ``no SUO'' drops the SUO feature from ``all''; same for others.
      The same parameters were used with \cref{fig:result-all}.
      }
    }
    \label{fig:result-ablation}
  \end{figure}
}

{
  \begin{figure}[t!]
    \centering
    \begin{tikzpicture}
      \node[anchor=south west]() at (0, 0)
           {\includegraphics[width=0.8\linewidth]{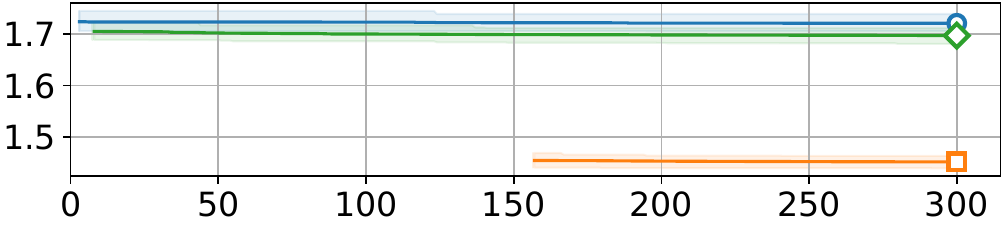}};
      \node[anchor=center]() at (2.8, 1.0)
           {\includegraphics[width=0.45\linewidth]{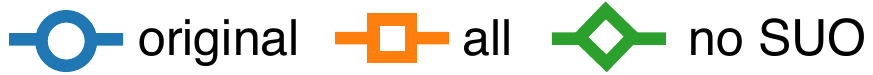}};
       \node[]() at (3.9, -0.1) {runtime (sec)};
       \node[]() at (-0.3, 1.0) {\rotatebox{90}{cost / LB}};
    \end{tikzpicture}
    \caption{
      Results for 10,000 agents in \mapname{warehouse-20-40-10-2-2}.
      \normalfont{
        ``all'' includes all techniques including the recursive call.
      }
    }
    \label{fig:result-10k}
  \end{figure}
}

\section{Conclusion}
\label{sec:conclusion}
This study introduced various techniques to improve the solution quality of \lacamstar.
While each technique itself is an adjustment influenced by established methods from search or MAPF literature, their amalgamated utilization yielded substantial improvements over the original \lacamstar.
With these results, we believe that this study continues to push the boundaries of MAPF, signifying a stride towards real-time, large-scale, and near-optimal MAPF.

\paragraph{Acknowledgments}
I thank the anonymous reviewers for their constructive comments.
This work was partly supported by JST ACT-X Grant Number JPMJAX22A1.

\balance
\bibliographystyle{sty/ACM-Reference-Format}
\bibliography{ref-macro,ref}


\begin{thebibliography}{37}


\ifx \showCODEN    \undefined \def \showCODEN     #1{\unskip}     \fi
\ifx \showDOI      \undefined \def \showDOI       #1{#1}\fi
\ifx \showISBNx    \undefined \def \showISBNx     #1{\unskip}     \fi
\ifx \showISBNxiii \undefined \def \showISBNxiii  #1{\unskip}     \fi
\ifx \showISSN     \undefined \def \showISSN      #1{\unskip}     \fi
\ifx \showLCCN     \undefined \def \showLCCN      #1{\unskip}     \fi
\ifx \shownote     \undefined \def \shownote      #1{#1}          \fi
\ifx \showarticletitle \undefined \def \showarticletitle #1{#1}   \fi
\ifx \showURL      \undefined \def \showURL       {\relax}        \fi
\providecommand\bibfield[2]{#2}
\providecommand\bibinfo[2]{#2}
\providecommand\natexlab[1]{#1}
\providecommand\showeprint[2][]{arXiv:#2}

\bibitem[\protect\citeauthoryear{Andreychuk and Yakovlev}{Andreychuk and
  Yakovlev}{2018}]%
        {andreychuk2018two}
\bibfield{author}{\bibinfo{person}{Anton Andreychuk} {and}
  \bibinfo{person}{Konstantin Yakovlev}.} \bibinfo{year}{2018}\natexlab{}.
\newblock \showarticletitle{Two techniques that enhance the performance of
  multi-robot prioritized path planning}. In
  \bibinfo{booktitle}{\emph{Proceedings of International Joint Conference on
  Autonomous Agents \& Multiagent Systems (AAMAS)}}.
\newblock


\bibitem[\protect\citeauthoryear{Barer, Sharon, Stern, and Felner}{Barer
  et~al\mbox{.}}{2014}]%
        {barer2014suboptimal}
\bibfield{author}{\bibinfo{person}{Max Barer}, \bibinfo{person}{Guni Sharon},
  \bibinfo{person}{Roni Stern}, {and} \bibinfo{person}{Ariel Felner}.}
  \bibinfo{year}{2014}\natexlab{}.
\newblock \showarticletitle{Suboptimal variants of the conflict-based search
  algorithm for the multi-agent pathfinding problem}. In
  \bibinfo{booktitle}{\emph{Proceedings of Annual Symposium on Combinatorial
  Search (SOCS)}}.
\newblock


\bibitem[\protect\citeauthoryear{Chen, Harabor, Li, and Stuckey}{Chen
  et~al\mbox{.}}{2023}]%
        {chen2023traffic}
\bibfield{author}{\bibinfo{person}{Zhe Chen}, \bibinfo{person}{Daniel Harabor},
  \bibinfo{person}{Jioyang Li}, {and} \bibinfo{person}{Peter~J Stuckey}.}
  \bibinfo{year}{2023}\natexlab{}.
\newblock \showarticletitle{Traffic Flow Optimisation for Lifelong Multi-Agent
  Path Finding}.
\newblock \bibinfo{journal}{\emph{arXiv preprint arXiv:2308.11234}}
  (\bibinfo{year}{2023}).
\newblock


\bibitem[\protect\citeauthoryear{Cohen, Greco, Ma, Hern{\'a}ndez, Felner,
  Kumar, and Koenig}{Cohen et~al\mbox{.}}{2018a}]%
        {cohen2018anytime}
\bibfield{author}{\bibinfo{person}{Liron Cohen}, \bibinfo{person}{Matias
  Greco}, \bibinfo{person}{Hang Ma}, \bibinfo{person}{Carlos Hern{\'a}ndez},
  \bibinfo{person}{Ariel Felner}, \bibinfo{person}{TK~Satish Kumar}, {and}
  \bibinfo{person}{Sven Koenig}.} \bibinfo{year}{2018}\natexlab{a}.
\newblock \showarticletitle{Anytime Focal Search with Applications.}. In
  \bibinfo{booktitle}{\emph{Proceedings of International Joint Conference on
  Artificial Intelligence (IJCAI)}}.
\newblock


\bibitem[\protect\citeauthoryear{Cohen, Wagner, Chan, Choset, Sturtevant,
  Koenig, and Kumar}{Cohen et~al\mbox{.}}{2018b}]%
        {cohen2018rapid}
\bibfield{author}{\bibinfo{person}{Liron Cohen}, \bibinfo{person}{Glenn
  Wagner}, \bibinfo{person}{David Chan}, \bibinfo{person}{Howie Choset},
  \bibinfo{person}{Nathan Sturtevant}, \bibinfo{person}{Sven Koenig}, {and}
  \bibinfo{person}{TK~Satish Kumar}.} \bibinfo{year}{2018}\natexlab{b}.
\newblock \showarticletitle{Rapid randomized restarts for multi-agent path
  finding solvers}. In \bibinfo{booktitle}{\emph{Proceedings of Annual
  Symposium on Combinatorial Search (SOCS)}}.
\newblock


\bibitem[\protect\citeauthoryear{De~Wilde, Ter~Mors, and Witteveen}{De~Wilde
  et~al\mbox{.}}{2014}]%
        {de2014push}
\bibfield{author}{\bibinfo{person}{Boris De~Wilde}, \bibinfo{person}{Adriaan~W
  Ter~Mors}, {and} \bibinfo{person}{Cees Witteveen}.}
  \bibinfo{year}{2014}\natexlab{}.
\newblock \showarticletitle{Push and rotate: a complete multi-agent pathfinding
  algorithm}.
\newblock \bibinfo{journal}{\emph{Journal of Artificial Intelligence Research
  (JAIR)}} (\bibinfo{year}{2014}).
\newblock


\bibitem[\protect\citeauthoryear{Erdmann and Lozano-Perez}{Erdmann and
  Lozano-Perez}{1987}]%
        {erdmann1987multiple}
\bibfield{author}{\bibinfo{person}{Michael Erdmann} {and}
  \bibinfo{person}{Tomas Lozano-Perez}.} \bibinfo{year}{1987}\natexlab{}.
\newblock \showarticletitle{On multiple moving objects}.
\newblock \bibinfo{journal}{\emph{Algorithmica}} (\bibinfo{year}{1987}).
\newblock


\bibitem[\protect\citeauthoryear{Felner, Li, Boyarski, Ma, Cohen, Kumar, and
  Koenig}{Felner et~al\mbox{.}}{2018}]%
        {felner2018adding}
\bibfield{author}{\bibinfo{person}{Ariel Felner}, \bibinfo{person}{Jiaoyang
  Li}, \bibinfo{person}{Eli Boyarski}, \bibinfo{person}{Hang Ma},
  \bibinfo{person}{Liron Cohen}, \bibinfo{person}{TK~Satish Kumar}, {and}
  \bibinfo{person}{Sven Koenig}.} \bibinfo{year}{2018}\natexlab{}.
\newblock \showarticletitle{Adding heuristics to conflict-based search for
  multi-agent path finding}. In \bibinfo{booktitle}{\emph{Proceedings of
  International Conference on Automated Planning and Scheduling (ICAPS)}}.
\newblock


\bibitem[\protect\citeauthoryear{Han and Yu}{Han and Yu}{2022}]%
        {han2022optimizing}
\bibfield{author}{\bibinfo{person}{Shuai~D Han} {and} \bibinfo{person}{Jingjin
  Yu}.} \bibinfo{year}{2022}\natexlab{}.
\newblock \showarticletitle{Optimizing space utilization for more effective
  multi-robot path planning}. In \bibinfo{booktitle}{\emph{Proceedings of IEEE
  International Conference on Robotics and Automation (ICRA)}}.
\newblock


\bibitem[\protect\citeauthoryear{Hart, Nilsson, and Raphael}{Hart
  et~al\mbox{.}}{1968}]%
        {hart1968formal}
\bibfield{author}{\bibinfo{person}{Peter~E Hart}, \bibinfo{person}{Nils~J
  Nilsson}, {and} \bibinfo{person}{Bertram Raphael}.}
  \bibinfo{year}{1968}\natexlab{}.
\newblock \showarticletitle{A formal basis for the heuristic determination of
  minimum cost paths}.
\newblock \bibinfo{journal}{\emph{IEEE transactions on Systems Science and
  Cybernetics}} (\bibinfo{year}{1968}).
\newblock


\bibitem[\protect\citeauthoryear{Kocsis and Szepesv{\'a}ri}{Kocsis and
  Szepesv{\'a}ri}{2006}]%
        {kocsis2006bandit}
\bibfield{author}{\bibinfo{person}{Levente Kocsis} {and} \bibinfo{person}{Csaba
  Szepesv{\'a}ri}.} \bibinfo{year}{2006}\natexlab{}.
\newblock \showarticletitle{Bandit based monte-carlo planning}. In
  \bibinfo{booktitle}{\emph{European Conference on Machine Learning (ECML)}}.
\newblock


\bibitem[\protect\citeauthoryear{Lam, Le~Bodic, Harabor, and Stuckey}{Lam
  et~al\mbox{.}}{2022}]%
        {lam2022branch}
\bibfield{author}{\bibinfo{person}{Edward Lam}, \bibinfo{person}{Pierre
  Le~Bodic}, \bibinfo{person}{Daniel Harabor}, {and} \bibinfo{person}{Peter~J
  Stuckey}.} \bibinfo{year}{2022}\natexlab{}.
\newblock \showarticletitle{Branch-and-cut-and-price for multi-agent path
  finding}.
\newblock \bibinfo{journal}{\emph{Computers \& Operations Research (COR)}}
  (\bibinfo{year}{2022}).
\newblock


\bibitem[\protect\citeauthoryear{Li, Chen, Harabor, Stuckey, and Koenig}{Li
  et~al\mbox{.}}{2021a}]%
        {li2021anytime}
\bibfield{author}{\bibinfo{person}{Jiaoyang Li}, \bibinfo{person}{Zhe Chen},
  \bibinfo{person}{Daniel Harabor}, \bibinfo{person}{P Stuckey}, {and}
  \bibinfo{person}{Sven Koenig}.} \bibinfo{year}{2021}\natexlab{a}.
\newblock \showarticletitle{Anytime multi-agent path finding via large
  neighborhood search}. In \bibinfo{booktitle}{\emph{Proceedings of
  International Joint Conference on Artificial Intelligence (IJCAI)}}.
\newblock


\bibitem[\protect\citeauthoryear{Li, Chen, Harabor, Stuckey, and Koenig}{Li
  et~al\mbox{.}}{2022}]%
        {li2022mapf}
\bibfield{author}{\bibinfo{person}{Jiaoyang Li}, \bibinfo{person}{Zhe Chen},
  \bibinfo{person}{Daniel Harabor}, \bibinfo{person}{Peter~J Stuckey}, {and}
  \bibinfo{person}{Sven Koenig}.} \bibinfo{year}{2022}\natexlab{}.
\newblock \showarticletitle{MAPF-LNS2: Fast Repairing for Multi-Agent Path
  Finding via Large Neighborhood Search}. In
  \bibinfo{booktitle}{\emph{Proceedings of AAAI Conference on Artificial
  Intelligence (AAAI)}}.
\newblock


\bibitem[\protect\citeauthoryear{Li, Chen, Zheng, Chan, Harabor, Stuckey, Ma,
  and Koenig}{Li et~al\mbox{.}}{2021b}]%
        {li2021scalable}
\bibfield{author}{\bibinfo{person}{Jiaoyang Li}, \bibinfo{person}{Zhe Chen},
  \bibinfo{person}{Yi Zheng}, \bibinfo{person}{Shao-Hung Chan},
  \bibinfo{person}{Daniel Harabor}, \bibinfo{person}{Peter~J Stuckey},
  \bibinfo{person}{Hang Ma}, {and} \bibinfo{person}{Sven Koenig}.}
  \bibinfo{year}{2021}\natexlab{b}.
\newblock \showarticletitle{Scalable rail planning and replanning: Winning the
  2020 flatland challenge}. In \bibinfo{booktitle}{\emph{Proceedings of
  International Conference on Automated Planning and Scheduling (ICAPS)}}.
\newblock


\bibitem[\protect\citeauthoryear{Li, Felner, Boyarski, Ma, and Koenig}{Li
  et~al\mbox{.}}{2019}]%
        {li2019improved}
\bibfield{author}{\bibinfo{person}{Jiaoyang Li}, \bibinfo{person}{Ariel
  Felner}, \bibinfo{person}{Eli Boyarski}, \bibinfo{person}{Hang Ma}, {and}
  \bibinfo{person}{Sven Koenig}.} \bibinfo{year}{2019}\natexlab{}.
\newblock \showarticletitle{Improved Heuristics for Multi-Agent Path Finding
  with Conflict-Based Search.}. In \bibinfo{booktitle}{\emph{Proceedings of
  International Joint Conference on Artificial Intelligence (IJCAI)}}.
\newblock


\bibitem[\protect\citeauthoryear{Li, Harabor, Stuckey, and Koenig}{Li
  et~al\mbox{.}}{2021c}]%
        {li2021pairwise}
\bibfield{author}{\bibinfo{person}{Jiaoyang Li}, \bibinfo{person}{Daniel
  Harabor}, \bibinfo{person}{Peter~J Stuckey}, {and} \bibinfo{person}{Sven
  Koenig}.} \bibinfo{year}{2021}\natexlab{c}.
\newblock \showarticletitle{Pairwise Symmetry Reasoning for Multi-Agent Path
  Finding Search}.
\newblock \bibinfo{journal}{\emph{Artificial Intelligence (AIJ)}}
  (\bibinfo{year}{2021}).
\newblock


\bibitem[\protect\citeauthoryear{Li, Ruml, and Koenig}{Li
  et~al\mbox{.}}{2021d}]%
        {li2021eecbs}
\bibfield{author}{\bibinfo{person}{Jiaoyang Li}, \bibinfo{person}{Wheeler
  Ruml}, {and} \bibinfo{person}{Sven Koenig}.}
  \bibinfo{year}{2021}\natexlab{d}.
\newblock \showarticletitle{EECBS: A Bounded-Suboptimal Search for Multi-Agent
  Path Finding}. In \bibinfo{booktitle}{\emph{Proceedings of AAAI Conference on
  Artificial Intelligence (AAAI)}}.
\newblock


\bibitem[\protect\citeauthoryear{Li, Tinka, Kiesel, Durham, Kumar, and
  Koenig}{Li et~al\mbox{.}}{2021e}]%
        {li2021lifelong}
\bibfield{author}{\bibinfo{person}{Jiaoyang Li}, \bibinfo{person}{Andrew
  Tinka}, \bibinfo{person}{Scott Kiesel}, \bibinfo{person}{Joseph~W Durham},
  \bibinfo{person}{TK~Satish Kumar}, {and} \bibinfo{person}{Sven Koenig}.}
  \bibinfo{year}{2021}\natexlab{e}.
\newblock \showarticletitle{Lifelong multi-agent path finding in large-scale
  warehouses}. In \bibinfo{booktitle}{\emph{Proceedings of AAAI Conference on
  Artificial Intelligence (AAAI)}}.
\newblock


\bibitem[\protect\citeauthoryear{Luna and Bekris}{Luna and Bekris}{2011}]%
        {luna2011push}
\bibfield{author}{\bibinfo{person}{Ryan Luna} {and} \bibinfo{person}{Kostas~E
  Bekris}.} \bibinfo{year}{2011}\natexlab{}.
\newblock \showarticletitle{Push and swap: Fast cooperative path-finding with
  completeness guarantees}. In \bibinfo{booktitle}{\emph{Proceedings of
  International Joint Conference on Artificial Intelligence (IJCAI)}}.
\newblock


\bibitem[\protect\citeauthoryear{Okumura}{Okumura}{2023a}]%
        {okumura2023lacam2}
\bibfield{author}{\bibinfo{person}{Keisuke Okumura}.}
  \bibinfo{year}{2023}\natexlab{a}.
\newblock \showarticletitle{Improving LaCAM for Scalable Eventually Optimal
  Multi-Agent Pathfinding}. In \bibinfo{booktitle}{\emph{Proceedings of
  International Joint Conference on Artificial Intelligence (IJCAI)}}.
\newblock


\bibitem[\protect\citeauthoryear{Okumura}{Okumura}{2023b}]%
        {okumura2023lacam}
\bibfield{author}{\bibinfo{person}{Keisuke Okumura}.}
  \bibinfo{year}{2023}\natexlab{b}.
\newblock \showarticletitle{LaCAM: Search-Based Algorithm for Quick Multi-Agent
  Pathfinding}. In \bibinfo{booktitle}{\emph{Proceedings of AAAI Conference on
  Artificial Intelligence (AAAI)}}.
\newblock


\bibitem[\protect\citeauthoryear{Okumura, Machida, Défago, and Tamura}{Okumura
  et~al\mbox{.}}{2022}]%
        {okumura2022priority}
\bibfield{author}{\bibinfo{person}{Keisuke Okumura}, \bibinfo{person}{Manao
  Machida}, \bibinfo{person}{Xavier Défago}, {and} \bibinfo{person}{Yasumasa
  Tamura}.} \bibinfo{year}{2022}\natexlab{}.
\newblock \showarticletitle{Priority Inheritance with Backtracking for
  Iterative Multi-agent Path Finding}.
\newblock \bibinfo{journal}{\emph{Artificial Intelligence (AIJ)}}
  (\bibinfo{year}{2022}).
\newblock


\bibitem[\protect\citeauthoryear{Okumura, Tamura, and D\'{e}fago}{Okumura
  et~al\mbox{.}}{2021}]%
        {okumura2021iterative}
\bibfield{author}{\bibinfo{person}{Keisuke Okumura}, \bibinfo{person}{Yasumasa
  Tamura}, {and} \bibinfo{person}{Xavier D\'{e}fago}.}
  \bibinfo{year}{2021}\natexlab{}.
\newblock \showarticletitle{Iterative Refinement for Real-Time Multi-Robot Path
  Planning}. In \bibinfo{booktitle}{\emph{Proceedings of IEEE/RSJ International
  Conference on Intelligent Robots and Systems (IROS)}}.
\newblock


\bibitem[\protect\citeauthoryear{Pearl and Kim}{Pearl and Kim}{1982}]%
        {pearl1982studies}
\bibfield{author}{\bibinfo{person}{Judea Pearl} {and} \bibinfo{person}{Jin~H
  Kim}.} \bibinfo{year}{1982}\natexlab{}.
\newblock \showarticletitle{Studies in semi-admissible heuristics}.
\newblock \bibinfo{journal}{\emph{IEEE transactions on pattern analysis and
  machine intelligence}} (\bibinfo{year}{1982}).
\newblock


\bibitem[\protect\citeauthoryear{Phillips and Likhachev}{Phillips and
  Likhachev}{2011}]%
        {phillips2011sipp}
\bibfield{author}{\bibinfo{person}{Mike Phillips} {and} \bibinfo{person}{Maxim
  Likhachev}.} \bibinfo{year}{2011}\natexlab{}.
\newblock \showarticletitle{Sipp: Safe interval path planning for dynamic
  environments}. In \bibinfo{booktitle}{\emph{Proceedings of IEEE International
  Conference on Robotics and Automation (ICRA)}}.
\newblock


\bibitem[\protect\citeauthoryear{Rego, Gamboa, Glover, and Osterman}{Rego
  et~al\mbox{.}}{2011}]%
        {rego2011traveling}
\bibfield{author}{\bibinfo{person}{C{\'e}sar Rego}, \bibinfo{person}{Dorabela
  Gamboa}, \bibinfo{person}{Fred Glover}, {and} \bibinfo{person}{Colin
  Osterman}.} \bibinfo{year}{2011}\natexlab{}.
\newblock \showarticletitle{Traveling salesman problem heuristics: Leading
  methods, implementations and latest advances}.
\newblock \bibinfo{journal}{\emph{European Journal of Operational Research
  (EJOR)}} (\bibinfo{year}{2011}).
\newblock


\bibitem[\protect\citeauthoryear{Richter, Thayer, and Ruml}{Richter
  et~al\mbox{.}}{2010}]%
        {richter2010joy}
\bibfield{author}{\bibinfo{person}{Silvia Richter}, \bibinfo{person}{Jordan
  Thayer}, {and} \bibinfo{person}{Wheeler Ruml}.}
  \bibinfo{year}{2010}\natexlab{}.
\newblock \showarticletitle{The joy of forgetting: Faster anytime search via
  restarting}. In \bibinfo{booktitle}{\emph{Proceedings of International
  Conference on Automated Planning and Scheduling (ICAPS)}}.
\newblock


\bibitem[\protect\citeauthoryear{Sharon, Stern, Felner, and Sturtevant}{Sharon
  et~al\mbox{.}}{2015}]%
        {sharon2015conflict}
\bibfield{author}{\bibinfo{person}{Guni Sharon}, \bibinfo{person}{Roni Stern},
  \bibinfo{person}{Ariel Felner}, {and} \bibinfo{person}{Nathan~R Sturtevant}.}
  \bibinfo{year}{2015}\natexlab{}.
\newblock \showarticletitle{Conflict-based search for optimal multi-agent
  pathfinding}.
\newblock \bibinfo{journal}{\emph{Artificial Intelligence (AIJ)}}
  (\bibinfo{year}{2015}).
\newblock


\bibitem[\protect\citeauthoryear{Shen, Chen, Cheema, Harabor, and Stuckey}{Shen
  et~al\mbox{.}}{2023}]%
        {shen2023tracking}
\bibfield{author}{\bibinfo{person}{Bojie Shen}, \bibinfo{person}{Zhe Chen},
  \bibinfo{person}{Muhammad~Aamir Cheema}, \bibinfo{person}{Daniel~D Harabor},
  {and} \bibinfo{person}{Peter~J Stuckey}.} \bibinfo{year}{2023}\natexlab{}.
\newblock \showarticletitle{Tracking Progress in Multi-Agent Path Finding}.
\newblock \bibinfo{journal}{\emph{arXiv preprint}} (\bibinfo{year}{2023}).
\newblock


\bibitem[\protect\citeauthoryear{Silver}{Silver}{2005}]%
        {silver2005cooperative}
\bibfield{author}{\bibinfo{person}{David Silver}.}
  \bibinfo{year}{2005}\natexlab{}.
\newblock \showarticletitle{Cooperative Pathfinding}. In
  \bibinfo{booktitle}{\emph{Proceedings of AAAI Conference on Artificial
  Intelligence and Interactive Digital Entertainment (AIIDE)}}.
\newblock


\bibitem[\protect\citeauthoryear{Stern, Sturtevant, Felner, Koenig, Ma, Walker,
  Li, Atzmon, Cohen, Kumar, et~al\mbox{.}}{Stern et~al\mbox{.}}{2019}]%
        {stern2019def}
\bibfield{author}{\bibinfo{person}{Roni Stern}, \bibinfo{person}{Nathan
  Sturtevant}, \bibinfo{person}{Ariel Felner}, \bibinfo{person}{Sven Koenig},
  \bibinfo{person}{Hang Ma}, \bibinfo{person}{Thayne Walker},
  \bibinfo{person}{Jiaoyang Li}, \bibinfo{person}{Dor Atzmon},
  \bibinfo{person}{Liron Cohen}, \bibinfo{person}{TK Kumar}, {et~al\mbox{.}}}
  \bibinfo{year}{2019}\natexlab{}.
\newblock \showarticletitle{Multi-Agent Pathfinding: Definitions, Variants, and
  Benchmarks}. In \bibinfo{booktitle}{\emph{Proceedings of Annual Symposium on
  Combinatorial Search (SOCS)}}.
\newblock


\bibitem[\protect\citeauthoryear{Surynek}{Surynek}{2013}]%
        {surynek2013redundancy}
\bibfield{author}{\bibinfo{person}{Pavel Surynek}.}
  \bibinfo{year}{2013}\natexlab{}.
\newblock \showarticletitle{Redundancy elimination in highly parallel solutions
  of motion coordination problems}.
\newblock \bibinfo{journal}{\emph{International Journal on Artificial
  Intelligence Tools (IJAIT)}} (\bibinfo{year}{2013}).
\newblock


\bibitem[\protect\citeauthoryear{{\'S}wiechowski, Godlewski, Sawicki, and
  Ma{\'n}dziuk}{{\'S}wiechowski et~al\mbox{.}}{2023}]%
        {swiechowski2023monte}
\bibfield{author}{\bibinfo{person}{Maciej {\'S}wiechowski},
  \bibinfo{person}{Konrad Godlewski}, \bibinfo{person}{Bartosz Sawicki}, {and}
  \bibinfo{person}{Jacek Ma{\'n}dziuk}.} \bibinfo{year}{2023}\natexlab{}.
\newblock \showarticletitle{Monte Carlo tree search: A review of recent
  modifications and applications}.
\newblock \bibinfo{journal}{\emph{Artificial Intelligence Review}}
  (\bibinfo{year}{2023}).
\newblock


\bibitem[\protect\citeauthoryear{Wagner and Choset}{Wagner and Choset}{2015}]%
        {wagner2015subdimensional}
\bibfield{author}{\bibinfo{person}{Glenn Wagner} {and} \bibinfo{person}{Howie
  Choset}.} \bibinfo{year}{2015}\natexlab{}.
\newblock \showarticletitle{Subdimensional expansion for multirobot path
  planning}.
\newblock \bibinfo{journal}{\emph{Artificial Intelligence (AIJ)}}
  (\bibinfo{year}{2015}).
\newblock


\bibitem[\protect\citeauthoryear{Wurman, D'Andrea, and Mountz}{Wurman
  et~al\mbox{.}}{2008}]%
        {wurman2008coordinating}
\bibfield{author}{\bibinfo{person}{Peter~R Wurman}, \bibinfo{person}{Raffaello
  D'Andrea}, {and} \bibinfo{person}{Mick Mountz}.}
  \bibinfo{year}{2008}\natexlab{}.
\newblock \showarticletitle{Coordinating hundreds of cooperative, autonomous
  vehicles in warehouses}.
\newblock \bibinfo{journal}{\emph{AI magazine}} (\bibinfo{year}{2008}).
\newblock


\bibitem[\protect\citeauthoryear{Yu and LaValle}{Yu and LaValle}{2013}]%
        {yu2013structure}
\bibfield{author}{\bibinfo{person}{Jingjin Yu} {and} \bibinfo{person}{Steven~M
  LaValle}.} \bibinfo{year}{2013}\natexlab{}.
\newblock \showarticletitle{Structure and Intractability of Optimal Multi-Robot
  Path Planning on Graphs.}. In \bibinfo{booktitle}{\emph{Proceedings of AAAI
  Conference on Artificial Intelligence (AAAI)}}.
\newblock


\end{thebibliography}
\appendix
\section*{Appendix}

\section{Subprocedures of \lacamstar}

\Cref{algo:subprocedures} presents the subprocedures of \lacamstar.
It uses the same notation as \cref{algo:lacam}.
Most parts are explicitly explained in the paper, while \cref{algo:subprocedures:revive} is not touched and hence complemented as follows.

Recall that \lacamstar employs pruning at \cref{algo:lacam:pruning} based on the $f$-value, which is the sum of cost-to-come ($g$-value) and the estimated cost-to-go ($h$-value).
As the search tree is refined during the search, previously pruned nodes might have lower $f$-values than the goal configuration.
\cref{algo:subprocedures:revive} reintroduces such nodes to \open to maintain an optimal search strategy.

{
  \newcommand{\dopen}{\m{\mathcal{D}}}
  \begin{algorithm}[ht!]
    \caption{subprocedures used in \lacamstar (\cref{algo:lacam})}
    \label{algo:subprocedures}
    \begin{algorithmic}[1]
      \small
      \Procedure{LowLevelSearch}{\N: search node}
      \State $\C \leftarrow \N.\tree.\pop()$
      \If{$\depth(\C) \leq |A|$}
      \State $i \leftarrow $ an agent that does not appear in the ancestors of $\C$
      \For{$u \in \neigh(\N.\config[i]) \cup \left\{ \N.\config[i] \right\}$}
      \State $\C\new \leftarrow \langle~\parent: \C, \who: i, \where: u~\rangle$
      \State $\N.\tree.\push\left(\C\new\right)$
      \EndFor
      \EndIf
      \State \Return \C
      \EndProcedure
      \medskip
      \Procedure{DijkstraUpdate}{\N: search node}
      \State $\dopen \leftarrow \llbracket \N \rrbracket$
      \While{$\dopen \neq \emptyset$}
      \State $\N\from \leftarrow \dopen.\pop()$
      \For{$\N\to \in \N\from.\neighbors$}
      \State $g \leftarrow \N\from.g + \cost_e(\N\from.\config, \N\to.\config)$
      \If{$g < \N\to.g$}
      \State $\N\to.g \leftarrow g$;~$\N\to.\parent \leftarrow \N\from$;~$\dopen.\push(\N\to)$
      \IfSingle{$\spadesuit \land \f(\N\to) < \f(\N\goal)$}{$\open.\push\left(\N\to\right)$}
      \label{algo:subprocedures:revive}
      \EndIf
      \EndFor
      \EndWhile
      \EndProcedure
    \end{algorithmic}
  \end{algorithm}
}

\section{Effect of parallel computation in Monte-Carlo Configuration Generation}
The employment of Monte-Carlo configuration generation presented in \cref{sec:mccg} assumes parallel computation to generate multiple configurations rapidly.
The speedup achieved through multi-threading is presented in \cref{table:result-parallel-seq}, where it compares with a sequential version of the Monte-Carlo generator without multi-threading.
As anticipated, the speedup becomes particularly pronounced for larger values of $k$.
Even for smaller $k$, an observable improvement is evident, although it is constrained by the associated overhead of multi-threading.

{
  \begin{table}[ht!]
    \centering
    \begin{tabular}{rrrrr}
      \toprule
      & $k=1$ & $k=10$ & $k=100$ \\\midrule
      sequential (ms) & 59.3 & 198.6 & 1575.8 \\
      parallel (ms) & - & 160.8 & 377.4 \\\midrule
      improvement (\%) & - & 19.1  &  76.1 \\
      \bottomrule
    \end{tabular}
    \medskip
    \caption{
      Average runtime for finding initial solutions.
      {\normalfont
        The data was from the \scenname{empty} scenario.
      }
    }
    \label{table:result-parallel-seq}
  \end{table}
}

\section{Yet Being Unsuccessful Ideas}
Up until this point, the paper has showcased effective engineering techniques for \lacamstar.
In contrast, this section briefly presents tried but yet unsuccessful attempts as a future guide.

\subsection{Improving Admissible Heuristics}
\paragraph{Motivation}
\lacamstar is based on the general search scheme, sequentially processing search nodes, each containing information about the cost-to-come ($g$-value) and the anticipated cost-to-go ($h$-value), which is determined by heuristics.
In general, possessing an accurate cost-to-go estimate can significantly alleviate search efforts.
Indeed, for conflict-based search (CBS)~\cite{sharon2015conflict}, the use of a more precise $h$-value has been empirically demonstrated to enhance scalability~\cite{felner2018adding,li2019improved}.
Within \lacamstar, a heuristic is designed for a configuration \Q, e.g., $\sum_{i \in A}(\Q[i], g_i)$.
Then this heuristic aids in pruning superfluous nodes at \cref{algo:lacam:pruning}.
The use of an advanced heuristic, closely aligned with the actual cost, is anticipated to expedite the convergence to optimal solutions by eliminating large sections of nodes extraneous.

\paragraph{Method}
We examined this direction by introducing supplementary costs to the initial heuristic whenever pairwise symmetries~\cite{li2021pairwise} within \Q are identified.
In brief, when dealing with two agents, if one of them must take additional steps to circumvent collisions and deviate from the shortest paths, that incremental distance can be integrated into the cost-to-go estimate.
Such scenarios can be promptly recognized in cases featuring specific symmetrical arrangements (e.g., rectangular symmetry) within a constant timeframe.

\paragraph{Observations}
For extremely small instances, the enhanced heuristic contributed to the discovery of optimal solutions.
Nonetheless, this benefit did not manifest for larger instances.
Instead, the computation of the heuristic itself emerged as a runtime bottleneck for \lacamstar.
This outcome arose from the heuristic estimate still being considerably distant from the actual cost, especially with many agents, thereby hindering effective pruning.
The potential resolution for this quandary might entail devising a more precise heuristic that can be computed rapidly.

\subsection{Bounded Suboptimal \lacamstar}
\paragraph{Motivation}
Bounded suboptimal algorithms guarantee quality gaps of the obtained solutions from their optimal counterparts in a given parameter.
These algorithms hold appeal for MAPF, given that while finding optimal solutions can be computationally demanding, these algorithms afford the potential to identify near-optimal solutions within a reasonable timeframe.
In fact, several bounded suboptimal algorithms have been proposed for MAPF, yielding favorable outcomes~\cite{barer2014suboptimal,wagner2015subdimensional,cohen2018anytime,li2021eecbs}.
Since \lacamstar follows the general search scheme, the creation of its bounded suboptimal version is feasible.
Through this adaptation, we hypothesized that refining the initial solution's quality could be achieved through pruning using the upper bound.

\paragraph{Method}
Drawing inspiration from focal search~\cite{pearl1982studies}, we formulated the bounded suboptimal version of \lacamstar as follows.
The modified version incorporates the tracking of a minimal $f$-value (the sum of $g$- and $h$-values) among nodes in $\open$.
Let this value $f\sub{min}$.
Given a suboptimality threshold $w \in \mathbb{R}_{\geq 1}$, when a node is extracted from \open at \cref{algo:lacam:extracting} with an $f$-value surpassing $w\cdot f\sub{min}$, the node is promptly popped and then reinserted into \open at a randomly chosen position.
No further operations are executed for this node in this search iteration.
This alteration guarantees that the initial solution quality of \lacamstar is confined above by a factor of $w$ compared to the optimal solution.

\paragraph{Observation}
We were interested in a modest $w$ value (e.g., $1.2$), as opting for an excessively large $w$ (e.g., 5) holds limited rationale;
the original \lacamstar already satisfies such thresholds.
Nevertheless, adopting a small $w$ led to the search process becoming stagnant.
This issue arises due to infrequent updates to $f\sub{min}$ in \lacamstar, primarily because the search nodes are not discarded upon extraction from \open.
The effective implementation of bounded suboptimal \lacamstar remains an open question.

\end{document}